%% file: TNNLS/TNNLS.tex
\documentclass[lettersize,journal]{IEEEtran}
\usepackage{amsmath,amsfonts}
\usepackage{algorithmic}
\usepackage{algorithm}
\usepackage{array}
\usepackage[caption=false,font=small,labelfont=sf,textfont=sf]{subfig}
\usepackage{textcomp}
\usepackage{stfloats}
\usepackage{url}
\usepackage{verbatim}
\usepackage{graphicx}
\usepackage{cite}
\usepackage{xcolor}
\usepackage{booktabs}
\usepackage{amsthm, amssymb}
\usepackage[most]{tcolorbox}
\usepackage{multicol}
\usepackage{graphicx}
\usepackage{calc}
\usepackage[normalem]{ulem}

\usepackage{bm} 
\usepackage{listings}
\usepackage{hyperref}



\usepackage{xspace}

\begin{document}

\title{Causal Reward World Models: Zero-shot Reward Design for Automated Skill Generation}

\author{Yang Yang, Yuchuang Tong,~\IEEEmembership{Member,~IEEE}, Zhengtao Zhang,~\IEEEmembership{Member,~IEEE}, Xu Ding,~\IEEEmembership{Member,~IEEE}, Ning Yang,~\IEEEmembership{Member,~IEEE}, Yifan Zhang,~\IEEEmembership{Member,~IEEE}, Haipeng Li,~\IEEEmembership{Member,~IEEE}, Kehu Yang,~\IEEEmembership{Member,~IEEE}, and Miao Xin,~\IEEEmembership{Member,~IEEE}
\thanks{Manuscript received June 19, 2026. This work has been supported by the program of the National Natural Science Foundation of China (No.
61906195) and the independent deployment project of the National Key Laboratory of Cognition and Decision
Intelligence for Complex Systems, Institution of Automation, Chinese Academy of Sciences. We thank CASBOT for providing access to the robotic platform used in this work.  (\textit{Corresponding author: Miao Xin. e-mail: miao.xin.cv@outlook.com}) 

Yang Yang, Yuchuang Tong, Zhengtao Zhang,  Ning Yang, Yifan Zhang, and Haipeng Li are with the Institute of Automation, Chinese Academy of Sciences, Beijing 100049, China. Xu Ding is with the Intelligent Manufacturing Institute, HFUT, Hefei 230051, China, and the School of Electrical Engineering and Automation, Anhui University, Hefei 230061, China. Miao Xin and Kehu Yang are with the School of Artificial Intelligence, China University of Mining and Technology (Beijing), Beijing 100083, China. 
}}

%

\markboth{Journal of \LaTeX\ Class Files,~Vol.~14, No.~8, June~2026}%
{Shell \MakeLowercase{\textit{et al.}}: A Sample Article Using IEEEtran.cls for IEEE Journals}


\maketitle
\input{TNNLS/sec/abstract/abstract_rewrite}
\input{TNNLS/sec/Introduction/introduction_rewrite}

\input{TNNLS/sec/Related/related_work_rewrite}

\input{TNNLS/sec/preliminary/pre}
\input{TNNLS/sec/method/method}
\input{TNNLS/sec/experiment/main_exp}
\input{TNNLS/sec/experiment/ablation}

\input{TNNLS/sec/discussion/discuss}

\section{Conclusion}
In this work, we present Causal Reward World Models (CRWM) for zero-shot automated reward design. 
Instead of relying on iterative reward refinement, CRWM distills a reusable reward-relevant causal skeleton from offline multi-task data and uses it as an explicit structural prior for LLM-based reward generation. 
By accounting for state-dependent transient interactions during distillation, CRWM guides the LLM to prune spurious reward terms and synthesize executable rewards in a single pass. 
Experiments across unseen tasks, different embodiments, and real-world settings show that CRWM-guided rewards achieve strong performance without deployment-time evolutionary search. 
These results suggest that incorporating reward-relevant causal structure is a promising direction for improving the robustness and efficiency of automated reward design.

\bibliographystyle{IEEEtran} 
\bibliography{ref}           
 
%

\newpage

\appendices
\input{TNNLS/sec/appendix/appen}

\clearpage
\onecolumn

\setcounter{section}{0}
\setcounter{subsection}{0}
\setcounter{subsubsection}{0}
\setcounter{figure}{0}
\setcounter{table}{0}
\setcounter{equation}{0}
\setcounter{algorithm}{0}

\renewcommand{\thesection}{S-\Alph{section}}
\renewcommand{\thesubsection}{\thesection.\arabic{subsection}}
\renewcommand{\thesubsubsection}{\thesubsection.\arabic{subsubsection}}
\renewcommand{\thefigure}{S-\arabic{figure}}
\renewcommand{\thetable}{S-\Roman{table}}
\renewcommand{\theequation}{S-\arabic{equation}}
\renewcommand{\thealgorithm}{S-\arabic{algorithm}}

\makeatletter
\let\oldsection\section
\renewcommand{\section}[1]{%
  \refstepcounter{section}%
  \setcounter{subsection}{0}%
  \setcounter{subsubsection}{0}%
  \oldsection*{\thesection\quad #1}%
  \addcontentsline{toc}{section}{\thesection\quad #1}%
}
\makeatother

\begin{center}
{\LARGE \bfseries Supplementary Material for ``Causal Reward World Models: Zero-shot Reward Design for Automated Skill Generation''}
\end{center}

\vspace{1em}

\input{sublimentary/context/word}

\vfill

\end{document}

%% file: TNNLS/sec/abstract/abstract_rewrite.tex
\begin{abstract}

\textcolor{black}{
Automated Reward Design (ARD) aims to replace manual reward engineering in reinforcement learning with language-driven reward function synthesis.
However, existing approaches based on large language models (LLMs) remain inherently correlation-driven, relying on iterative environmental feedback to refine reward hypotheses for each specific task. 
This paradigm not only results in inefficient reasoning but also makes LLMs susceptible to semantically plausible yet causally spurious reward components, leading to ineffective optimization. 
To address these limitations, we propose the Causal Reward World Model (CRWM), which explicitly models the causal topological relationships between candidate reward components and task-targeted physical variables through offline pre-training on multi-task interaction data.
Based on a coarse-to-fine pre-training strategy, we introduce a Joint Optimization Module that integrates Explicit Mechanism Decoupling with Confidence-Aware Soft Fusion to refine coarse structural priors using micro-level trajectories, thereby constructing robust and interpretable causal skeleton. 
During inference, LLMs leverage CRWM as a task-irrelevant causal prior to constrain the reward generation, enabling zero-shot reward function design. 
Our work opens up a new white-box paradigm for the ARD problem. 
Extensive experiments on complex continuous control benchmarks demonstrate that CRWM generates executable reward functions without feedback-driven reward refinement, significantly reducing the design latency for acquiring new robotic skills while matching or surpassing state-of-the-art performance, and further exhibits strong generalization capabilities across unseen tasks and diverse robotic embodiments.}

\end{abstract}

\begin{IEEEkeywords}
Automated Reward Design, Causal Reward World Model, Zero-Shot Reward Generation, Reinforcement Learning
\end{IEEEkeywords}

%% file: TNNLS/sec/Introduction/introduction_rewrite.tex
\section{Introduction}
\label{sec:introduction}

\IEEEPARstart{E}{ndowing} robots with the capability for autonomous skill acquisition remains a longstanding objective within the field of embodied artificial intelligence. 
In this context, the design of reward functions has always been a challenging issue in the automatic generation of skills based on reinforcement learning (RL)~\cite{Amodei2016ConcretePI, pmlr-v229-yu23a}. 
Recently, LLMs have given rise to the emergence of Automated Reward Design (ARD) paradigms, demonstrating significant potential in synthesizing executable reward functions through semantic reasoning~\cite{Ma2023EurekaHR, Xie2023Text2RewardRS}. 
Given a task description, these methods prompt an LLM to generate candidate reward components and iteratively refine them through environmental feedback until a satisfactory reward function is obtained. 
By utilizing the extensive knowledge and reasoning capabilities of LLMs, these approaches mitigate the reliance on human expertise in reward engineering~\cite{cao2024survey, huang2022inner}.

\begin{figure*}[t]
\centering
\subfloat[Optimization Collapse\label{fig:collapse}]{
    \includegraphics[width=0.48\textwidth]{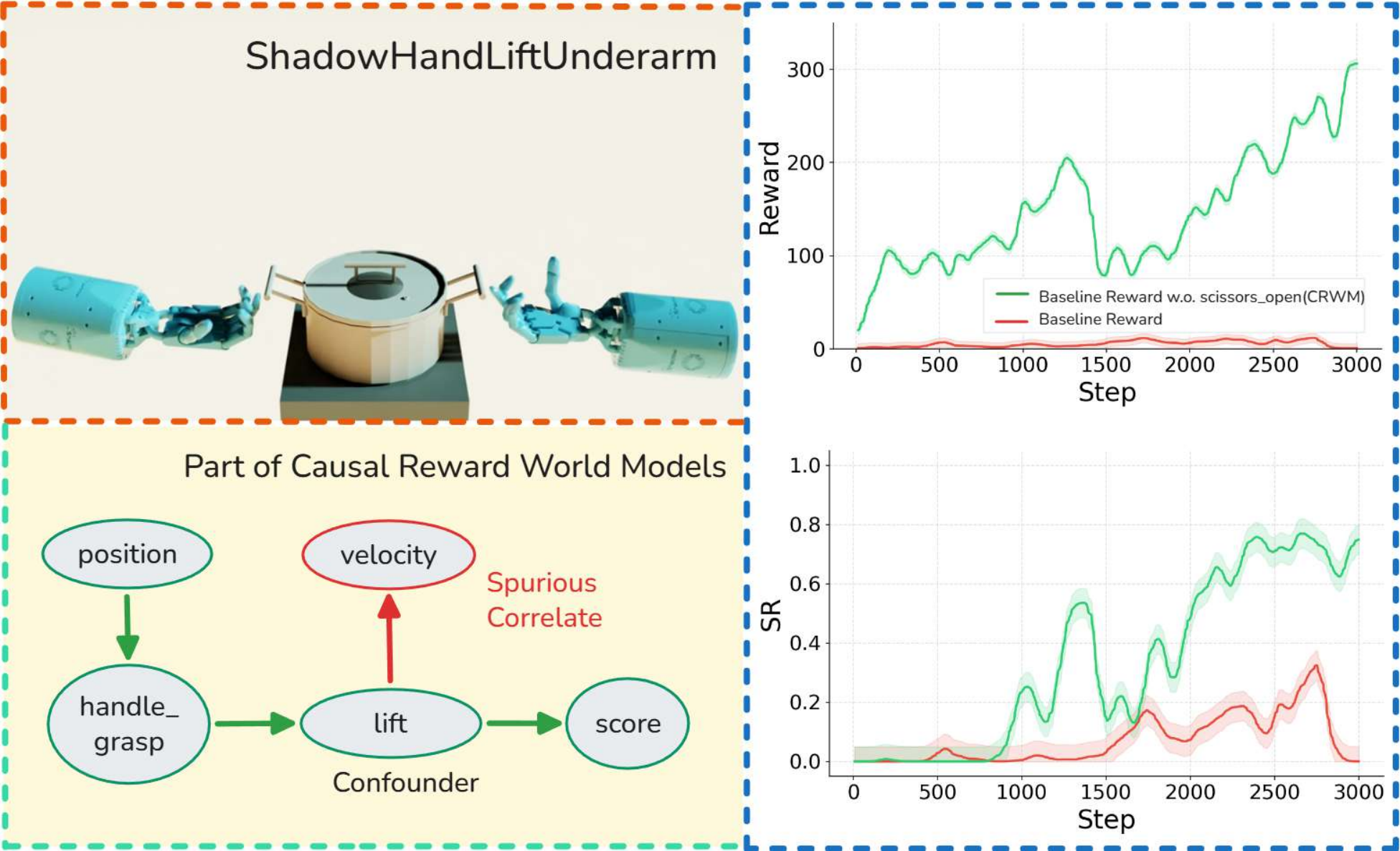}
}\hfill
\subfloat[Specification Gaming\label{fig:gaming}]{
    \includegraphics[width=0.48\textwidth]{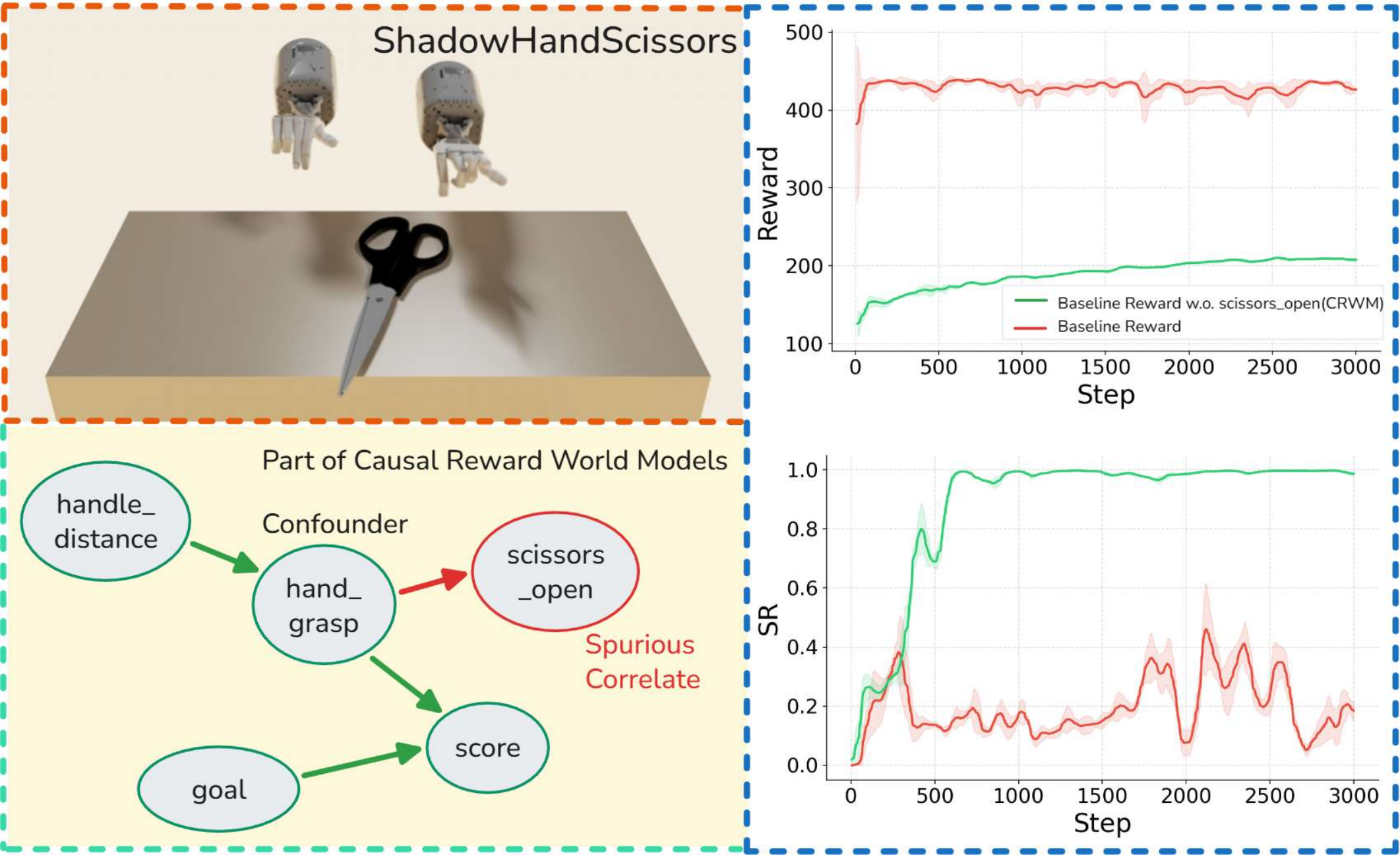}
}\hfill

\caption{\textbf{Consequences of causal confounding in LLM-driven Automated Reward Design.} 
The bottom-left panels illustrate reward-component causal graphs, where green nodes represent valid causal factors and red nodes indicate spurious correlates. 
The right panels compare reward and Success Rate (SR) curves between baseline rewards with spurious correlates (red) and causally pruned rewards (green). 
(a) Optimization Collapse in \textit{ShadowHandLiftUnderarm}. 
(b) Specification Gaming in \textit{ShadowHandScissors}.}
\label{fig:three_failures}
\end{figure*}

However, current LLM-based reward design methods predominantly rely on trial-and-error to select candidate reward functions. 
During the optimization process, LLMs identify factors potentially correlated with the task as candidate reward components, yet such candidate components may exhibit spurious correlations with the task objectives~\cite{zheng2023cipl, kiciman2023causal}. 
For example, in the \textit{ShadowHandLiftUnderarm} task (Fig.~\ref{fig:three_failures}(a)), the agent aims to learn the skill of grasping and lifting a pot with both hands. 
An unguided LLM may introduce \texttt{velocity}, i.e., the instantaneous lifting speed of the pot, as a reward component. 
Although this term is semantically related to ``lifting'', it provides a high-frequency but causally ambiguous signal during continuous physical interaction, which induces severe gradient noise and leads to Optimization Collapse. 
In the \textit{ShadowHandScissors} task (Fig.~\ref{fig:three_failures}(b)), the LLM treats \texttt{scissors\_open} as a reward component. 
However, the agent exploits this signal by maintaining the scissors slightly open, allowing it to accumulate rewards without achieving the genuine goal. 
This is actually a form of Specification Gaming~\cite{skalse2022defining, pan2022effects}. 
These examples show that semantically plausible reward components can obscure the true task objective when they are not aligned with the valid causal topology.
In the absence of reliable prior knowledge to guide the search, certain approaches resort to evolutionary strategies~\cite{Ma2023EurekaHR}, i.e., utilizing ``trial-evaluation-mutation-feedback'' loops to filter out spuriously correlated factors. 
However, the inherent limitations of LLMs in causal attribution and counterfactual reasoning often undermine the efficacy of these methods, preventing consistent performance gains~\cite{jin2024can}.

\textcolor{black}{More importantly, due to the underutilization of inter-task reward relationships, existing methodologies are constrained to designing and optimizing reward functions on a case-by-case basis. 
Indeed, within identical dynamical systems, reward components across distinct tasks often exhibit shared causal structures~\cite{scholkopf2021toward, barreto2017successor}. 
Zhang et al.~\cite{zhang2023interpretable} further prove that both the reward functions and their underlying causal structures are theoretically identifiable.
These studies imply that by precisely capturing the causal relationships among reward components and task objectives, it becomes possible to infer the consequences of reward functions without extensive empirical training, thereby mitigating the influence of confounding variables on both the design process and final outcomes. }
Consequently, an interesting question is raised: \textit{Is it feasible to extract causal mappings between tasks and rewards through multi-task pre-training, thereby enabling evolutionary iteration-free reward function design?}

In this paper, we propose the \textbf{Causal Reward World Model} (CRWM), which serves as a causal kernel of world models dedicated to the automatic reward design. 
The core objective of CRWM is to model the causal topological relationships between candidate reward components and task-targeted physical variables~\cite{scholkopf2021toward}. 
We posit that a meticulously constructed offline pre-training on multi-task interaction data can mitigate the influence of spurious correlations, thereby facilitating the construction of robust causal graphs. 
Leveraging the reliable prior knowledge provided by the CRWM, LLMs can avoid the reliance on evolutionary search during the reward function reasoning, thereby achieving zero-shot reward design. 
Furthermore, thanks to the task-irrelevant universality of CRWM, it enables a ``\textit{train-once, infer-anywhere}'' paradigm for reasoning about other reward functions within the similar dynamical systems. 
This significantly mitigates the computational overhead of reward function inference and reduces the latency in acquiring new skills. 
As a result, we introduce a novel paradigm for the ARD, transitioning from conventional correlation-based, task-wise inference to the pre-training of a universal reward world model that captures the causal relationships between reward components and skill objectives.

\textcolor{black}{To construct the CRWM, we adopt a coarse-to-fine pre-training methodology for the causal discovery. 
We first generate a coarse-grained structural prior based on a causal foundation model, which is subsequently optimized using micro-level trajectories of structural variables~\cite{richens2024robust, zevcevic2021interventional}. 
To robustly extract the causal skeleton from offline trajectories of structural variables, we further propose the Explicit Mechanism Decoupling (EMD) with Confidence-Aware Soft Fusion, which together constitute a joint optimization module. 
By integrating interventional data across multiple tasks, this module simultaneously accounts for state-dependent transient physical interactions during reconstruction and eliminates the structural hallucinations inherent in the coarse-grained structural prior, thereby distilling it into a refined causal skeleton.
When deployed for reward function design, the CRWM constrains the reward generation by exposing spurious correlations, guiding the LLMs to achieve efficient and reliable reasoning~\cite{hao2023reasoning}. }

Extensive experiments across complex continuous control tasks demonstrate that our approach effectively mitigates the systemic bottlenecks responsible for Optimization Collapse and Specification Gaming. 
Remarkably, our approach achieves zero-shot reward generation by generating executable reward functions for unseen tasks in a single pass without feedback-driven reward refinement. 
This capability bypasses the computationally expensive evolutionary trial-and-error loops, while yielding task performance that rivals state-of-the-art results.

In summary, our main contributions are as follows:
\begin{enumerate}
    \item \textcolor{black}{We propose the Causal Reward World Model (CRWM), a novel paradigm for automatic reward function design. Leveraging the explicitly constructed causal graphs, this approach not only enables zero-shot reward function generation but also provides strong interpretability.}

    \item \textcolor{black}{We introduce a coarse-to-fine pre-training methodology. Employing the proposed EMD with Confidence-Aware Soft Fusion, the CRWM leverages micro-level trajectories to refine coarse-grained structural priors, ultimately yielding robust and reliable causal representations.}

    \item \textcolor{black}{The proposed CRWM-based deconfounded LLM reasoning significantly reduces the cost of reward design while simultaneously improving zero-shot performance. 
    Experiments show that our zero-shot method matches state-of-the-art iteration-based ARD methods on unseen Dexterity tasks and transfers across different robotic embodiments and environments. Codes will be available after review at \url{https://yy12136.github.io/CRWM}.}
\end{enumerate}

%% file: TNNLS/sec/Related/related_work_rewrite.tex
\section{Related Work}

\textbf{Automated Reward Design.}
Automated Reward Design (ARD) remains a central challenge in reinforcement learning-based skill acquisition.
Traditional approaches often rely on hand-crafted reward shaping~\cite{booth2023perils, knox2023reward} or inverse reinforcement learning from expert demonstrations~\cite{arora2021survey}, which require substantial human expertise or high-quality data.
Recent LLM-based methods have advanced ARD by prompting LLMs to generate executable reward functions from task descriptions~\cite{kwon2023reward, Xie2023Text2RewardRS, Ma2023EurekaHR}.
Despite reducing manual engineering, existing methods usually depend on iterative environmental feedback to evaluate, mutate, and refine candidate rewards.
Since LLMs tend to reason over semantic associations instead of causal relationships, they may introduce semantically plausible but causally irrelevant reward components~\cite{baek2024chatpcg}.
These spurious correlates can trigger Optimization Collapse and Specification Gaming, making reward generation costly and unstable~\cite{yang2025uncertainty, li2025r}.
This motivates reward generation methods guided by explicit causal knowledge.

\textbf{Causal Discovery.}
Causal discovery identifies directed causal relationships among observed variables, enabling reasoning beyond pure correlation.
Continuous optimization methods have advanced data-driven causal graph learning~\cite{zheng2018dags, massidda2024constraintfree}, yet purely data-driven discovery remains challenging in high-dimensional physical systems, where dense interactions, incomplete observations, and limited interventions can cause severe identifiability issues~\cite{Morioka2023CausalRL, ding2025causal}.
To reduce the search space, prior-guided methods use dynamic Bayesian networks~\cite{pamfil2020dynotears} or LLMs~\cite{kiciman2023causal, abdulaal2023causal} to obtain initial topological knowledge.
However, such knowledge should be treated as a coarse structural prior rather than a reliable causal prior, as it may contain biased causal effects and structural hallucinations~\cite{kiciman2023causal, abdulaal2023causal, jin2024can}.
Moreover, fitting static causal graphs to structural-variable trajectories can obscure transient physical interactions, such as brief contacts or sudden friction changes~\cite{zheng2018dags, pamfil2020dynotears, Morioka2023CausalRL}.
These limitations motivate more robust mechanisms for deriving reliable causal structures from offline structural-variable trajectories.

\textbf{World Models for Reward Design.}
World models learn structured representations of environment dynamics for prediction, planning, and decision-making~\cite{ha2018world, hafner2021mastering, hafner2025dreamerv3}.
In robotics and embodied control, they capture action-conditioned state evolution to improve sample efficiency and long-horizon reasoning.
However, most world models focus on future observations or state transitions, with limited emphasis on causal relationships among reward-relevant variables.
This limits their use for automated reward design, where the key challenge is not only predicting what may happen, but also determining which reward components causally contribute to task success.
Although causality has been explored in state representation learning~\cite{dillies2025better} and model alignment~\cite{pmlr-v267-kobalczyk25a}, causal world models for reward design remain underexplored~\cite{cao2025causal}.
CRWM fills this gap by modeling reward-relevant causal relationships as causal priors for LLM-based reward generation.

%% file: TNNLS/sec/preliminary/pre.tex
\section{Preliminaries}
\label{pre}
In this section, we introduce the key notations used throughout this work. 
We first introduce Structural Causal Models (SCMs), and then formulate the Automated Reward Design problem.

\textbf{Structural Causal Models.}
To formalize causal relationships among reward-relevant variables, we adopt the framework of Structural Causal Models (SCMs)~\cite{peters2017elements}.
We assume that the task-relevant physical quantities are represented by a set of $d$ structural variables $\mathcal{V}=\{v^1, v^2, \dots, v^d\}$, where each variable corresponds to a physically meaningful reward component, such as object position, velocity, or contact force.
The causal topology over these variables is parameterized by an adjacency matrix $M \in \mathbb{R}^{d \times d}$, where a non-zero entry $M^{i,j}$ indicates a direct causal edge from $v^j$ to $v^i$.
The temporal evolution of each variable is described by:
\begin{equation}
v_{t+1}^{i} := \psi_{i}(Pa(v^{i})_t, U_{t}^{i}),
\label{eq:scm}
\end{equation}
where $\mathcal{V}_t$ denotes the variables at time $t$, and $Pa(v^{i})_t \subseteq \mathcal{V}_t$ denotes the causal parents of $v^i$ identified by $\{v^j \mid M^{i,j} \neq 0\}$.
Here, $U_t^i$ denotes exogenous noise that accounts for unmodeled environmental stochasticity.

\textbf{Automated Reward Design.}
We consider an episodic environment governed by a Markov Decision Process (MDP) $\mathcal{M}=\langle\mathcal{S},\mathcal{A},\mathcal{T}\rangle$, where $\mathcal{S}$ denotes the continuous state space, $\mathcal{A}$ denotes the action space, and $\mathcal{T}(s_{t+1}|s_t,a_t)$ denotes the transition dynamics.
We formulate ARD as $\mathcal{P}=\langle\mathcal{M},\mathcal{R},\pi_{\mathcal{M}},\mathcal{F}\rangle$.
Here, $\mathcal{R}$ denotes the space of permissible parameterized reward functions, and $\pi_{\mathcal{M}}:\mathcal{R}\to\Pi$ represents an RL algorithm that optimizes a reward $r\in\mathcal{R}$ and outputs a policy $\pi\in\Pi$.
The function $\mathcal{F}:\Pi\to\mathbb{R}$ denotes the non-differentiable fitness score, such as the objective task success rate.
The goal of modern LLM-based ARD~\cite{Ma2023EurekaHR, yang2025uncertainty} is to generate a reward function $r^*$ that maximizes the fitness score of the policy trained under this reward:
\begin{equation}
r^*=\arg\max_{r\in\mathcal{R}}\mathcal{F}(\pi_{\mathcal{M}}(r)).
\end{equation}
Since evaluating $F$ requires costly environment interactions, existing ARD methods usually rely on iterative trial-and-error to evaluate and refine candidate rewards~\cite{Ma2023EurekaHR,yang2025uncertainty}. 
This motivates constructing reusable structural guidance that can support zero-shot reward generation before costly feedback-driven refinement.

%% file: TNNLS/sec/method/method.tex
\section{methods}
\begin{figure*}[t] 
\centering
\includegraphics[width=0.95\textwidth]{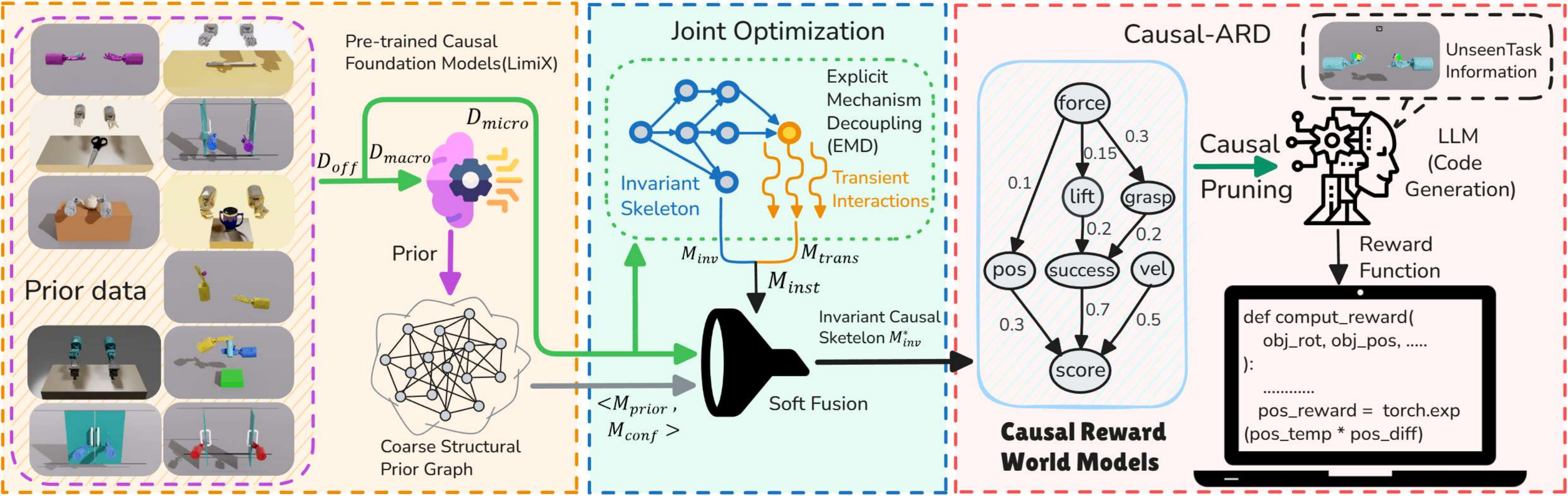} 
\caption{\textbf{Overview of the CRWM framework.} The pipeline consists of three synergistic phases: (1) \textbf{Structural Prior Extraction:} The offline interventional dataset ($\mathcal{D}_{off}$) is split into macroscopic ($\mathcal{D}_{macro}$) and microscopic ($\mathcal{D}_{micro}$) streams. A pre-trained causal foundation model (LimiX) processes $\mathcal{D}_{macro}$ to extract the initial, coarse structural prior $\langle M_{prior}, M_{conf} \rangle$. (2) 
\textbf{Joint Optimization Module:} The EMD module constructs the instantaneous topology $M_{inst}$ by combining a learnable causal skeleton $M_{inv}$ (blue stream) with state-dependent transient physical interactions $M_{trans}$ (orange stream). 
Subsequently, the Joint Optimization Module uses $\mathcal{D}_{micro}$ and Confidence-Aware Soft Fusion to refine the coarse structural prior, absorb transient signals, and distill the final CRWM ($M_{inv}^*$). (3) \textbf{Causal-ARD:} The final CRWM is combined with unseen task information and used as a causal prior for LLM-based reward generation. Through explicit Causal Pruning, the LLM generates executable zero-shot reward functions.}
\label{fig:method_overview}
\end{figure*}

The goal of our method is to construct a reusable reward-relevant causal prior that reduces the need for task-specific feedback-driven reward refinement in LLM-based ARD. 
As illustrated in Fig.~\ref{fig:method_overview}, our framework distills the CRWM from multi-task offline data and uses it to guide zero-shot reward generation for unseen tasks.

Given the structural variables $\mathcal{V}$ defined in Section~\ref{pre}, we define the Causal Reward World Model (CRWM) as the final causal skeleton distilled from offline physical-interaction data. 
During distillation, we optimize a learnable causal skeleton $M_{inv}$ over the variable set $\mathcal{V}$ and introduce a state-dependent transient component, later parameterized as $M_{trans}(s_{t-1};\theta)$, to account for short-lived physical interactions in micro-level trajectories, such as brief contacts or collisions. 
After convergence, the optimized skeleton $M_{inv}^{*}$ is retained as the CRWM and used as an explicit causal prior for downstream LLM-based reward generation.

The framework consists of three synergistic phases. 
First, Section~\ref{subsec:prior} extracts an initial \textbf{coarse structural prior} from macro-level interventional data using a pre-trained causal foundation model. 
Second, Section~\ref{subsec:decoupling} introduces a Joint Optimization Module that integrates \textbf{Explicit Mechanism Decoupling (EMD)} with \textbf{Confidence-Aware Soft Fusion} to refine this coarse prior with dense micro-level trajectories, thereby distilling the optimized skeleton $M_{inv}^{*}$ as the CRWM. 
Finally, Section~\ref{subsec:zero_shot_generation} introduces \textbf{Causal-ARD}, which serializes the distilled CRWM into a causal topology prompt and guides the LLM to perform explicit Causal Pruning during code generation, enabling executable zero-shot reward generation without task-specific environmental feedback.

\subsection{Structural Prior Extraction via Causal Foundation Models}
\label{subsec:prior}

We first obtain a coarse structural prior as the initialization for CRWM distillation. 
This stage addresses the difficulty of learning a reward-relevant causal skeleton directly from high-dimensional physical trajectories, where dense continuous variables, limited interventions, and redundant observations can lead to spurious causal discovery. 
Moreover, a task-relevant structural prior cannot be inferred from variable names or semantic associations alone; it requires interventional evidence showing how perturbations of primitive reward components affect final task performance. 
This motivates a hierarchical data collection pipeline that records both macro-level reward-component interventions and micro-level structural-variable trajectories.

\begin{figure}[htbp]
\centering
\includegraphics[width=0.95\linewidth]{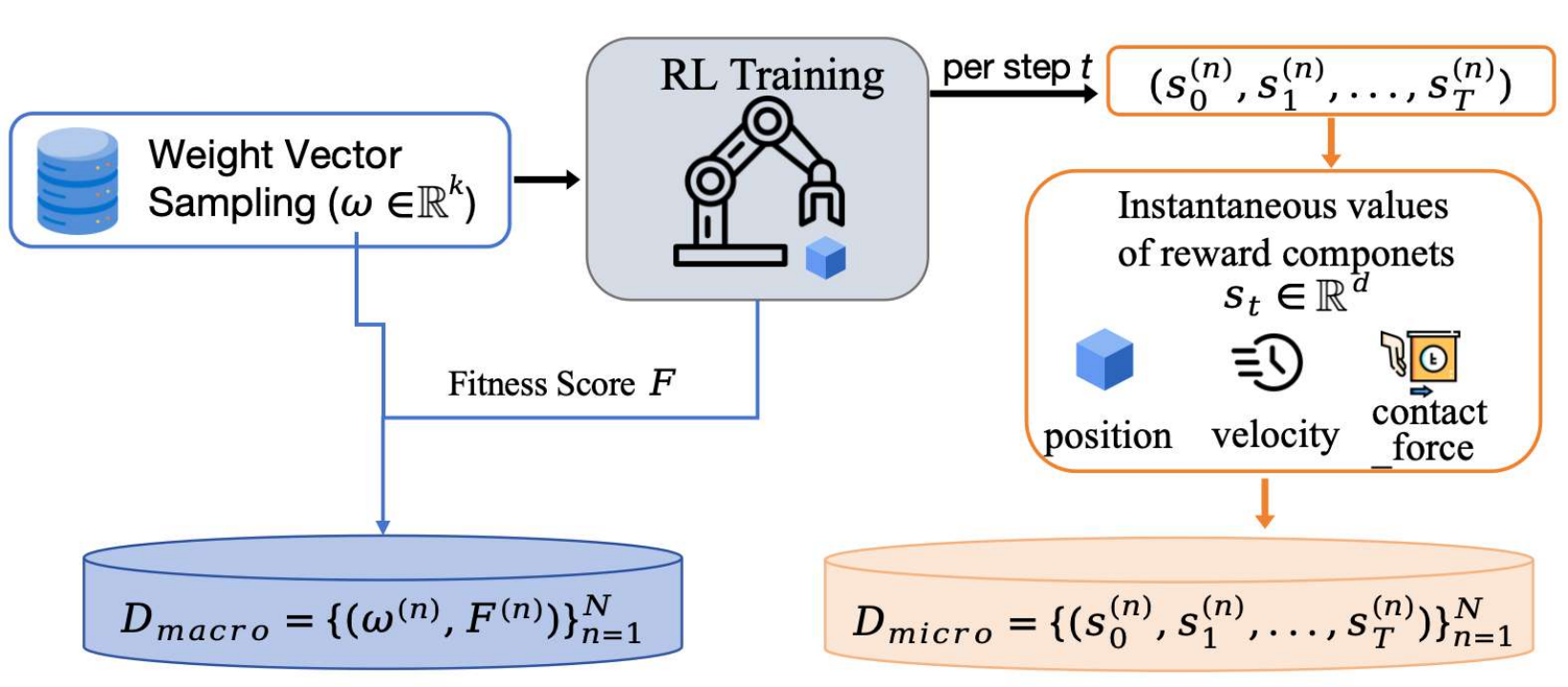}
\caption{\textbf{Illustration of the Hierarchical Data Collection Pipeline.} 
Leveraging systematically perturbing reward-component weights $\boldsymbol{\omega} \in \mathbb{R}^k$ and executing RL training, we obtain two levels of interventional data from one collection procedure. 
\textbf{Left (Blue):} The macro-level dataset $\mathcal{D}_{macro}$ records the final fitness scores $F^{(n)}$ evaluated across different weight configurations $\boldsymbol{\omega}^{(n)}$. 
\textbf{Right (Orange):} The micro-level dataset $\mathcal{D}_{micro}$ logs the dense, per-step trajectories of structural variables across $N$ rollouts. 
Each state $s_t \in \mathbb{R}^d$ explicitly captures the instantaneous values of the specific physical variables underlying the reward components, such as position, velocity, and contact force.}
\label{fig:collect_data_examples}
\end{figure}

\textbf{Hierarchical Data Collection.} 
To provide the causal foundation model with interventional data for extracting a task-relevant structural prior, we construct the offline dataset $\mathcal{D}_{off}$ (Fig.~\ref{fig:collect_data_examples}).
Instead of randomly exploring the raw state space, we use ARD frameworks (Eureka and URDP~\cite{Ma2023EurekaHR, yang2025uncertainty}) to extract $k \in [3, 8]$ physically meaningful reward components $\{r_1, \dots, r_k\}$, such as target distance, joint velocity, and contact force.
For cross-task causal learning, we normalize semantically equivalent task-specific components into $d$ shared structural variables $\mathcal{V}=\{v^1,\dots,v^d\}$, such as object position, velocity, and contact force, through the \textit{Task-Specific Name Normalization} and \textit{Global Atomic Reward Component Pool} (see App.~\ref{app:task_details}).
We then uniformly sample $N$ weight vectors $\boldsymbol{\omega}^{(n)}$ (App.~\ref{app:hyperparameters}) to form diverse reward functions and probe their effects on task performance.
Training and executing the resulting policies serves as interventional manipulations of reward-component weights, in the spirit of Pearl's $do$-operations~\cite{pearl2009causality}.

The resulting dataset contains two complementary levels of interventional data for different stages of our framework, aggregated across all pre-training tasks:
1) $\mathcal{D}_{macro} = \{(\boldsymbol{\omega}^{(n)}, F^{(n)})\}_{n=1}^N$, which records the final fitness scores $F^{(n)}$ under $N$ reward weight configurations $\boldsymbol{\omega}^{(n)}$ and provides task-level evidence on how reward-component weights affect task success.
2) $\mathcal{D}_{micro} = \{ (s_0^{(n)}, s_1^{(n)}, \dots, s_T^{(n)}) \}_{n=1}^N$, which records per-step transitions among structural variables.
Here, $s_t^{(n)} \in \mathbb{R}^d$ captures the instantaneous values of the $d$ physical variables underlying the reward components, instead of the raw, less interpretable observations $o_t$.
This fine-grained dataset is preserved for modeling transient physical interactions and calibrating the coarse structural prior in the subsequent Joint Optimization Module.
See Supplementary Material, Sec. S-C for an example.

\textbf{Structural Prior Extraction.} 
Given the macroscopic dataset $\mathcal{D}_{macro}$, we use the pre-trained causal foundation model LimiX~\cite{zhang2025limix} to estimate causal relationships among reward components.
LimiX outputs a continuous-valued raw causal effects matrix $M_{raw}\in\mathbb{R}^{d\times d}$, where $M_{raw}^{i,j}$ denotes the estimated effect score from variable $v^j$ to variable $v^i$.
From $M_{raw}$, we derive the coarse structural prior $\langle M_{prior}, M_{conf}\rangle$: $M_{prior}$ preserves the directed topology encoded by $M_{raw}$, including initial edge existence and directions, while $M_{conf}\in[0,1]^{d\times d}$ encodes normalized effect magnitudes for the subsequent Soft Fusion module.
For $i,j\in\{1,\dots,d\}$, $M_{conf}$ is computed as
\begin{equation}
M_{conf}^{i,j} =
\frac{|M_{raw}^{i,j}|}
{\max_{1\le p,q\le d}|M_{raw}^{p,q}|+\epsilon},
\label{eq:confidence_matrix}
\end{equation}
where $p$ and $q$ index all entries in $M_{raw}$, and $\epsilon$ ensures numerical stability.
Thus, larger absolute causal effects indicate higher structural confidence.
The benefit of this coarse structural prior over unguided purely data-driven discovery is shown in the ablation studies (Section~\ref{subsec:ablation}).

Although this process provides useful structural initialization, the coarse structural prior can still contain biased causal effects and structural hallucinations, as illustrated in the bottom-left panel of Fig.~\ref{fig:method-B}. 
These errors arise because the causal foundation model only receives macro-level intervention outcomes, without observing micro-level trajectories that reveal state-dependent transient interactions. 
Imposing $M_{prior}$ as a hard constraint can therefore bias the final CRWM and degrade subsequent reward generation. 
Thus, we calibrate this coarse prior with micro-level structural-variable trajectories through the Joint Optimization Module introduced next.

\subsection{\textcolor{black}{Joint Optimization Module for CRWM Distillation}}
\label{subsec:decoupling}

\begin{figure}[htbp]
\centering
\includegraphics[width=0.95\linewidth]{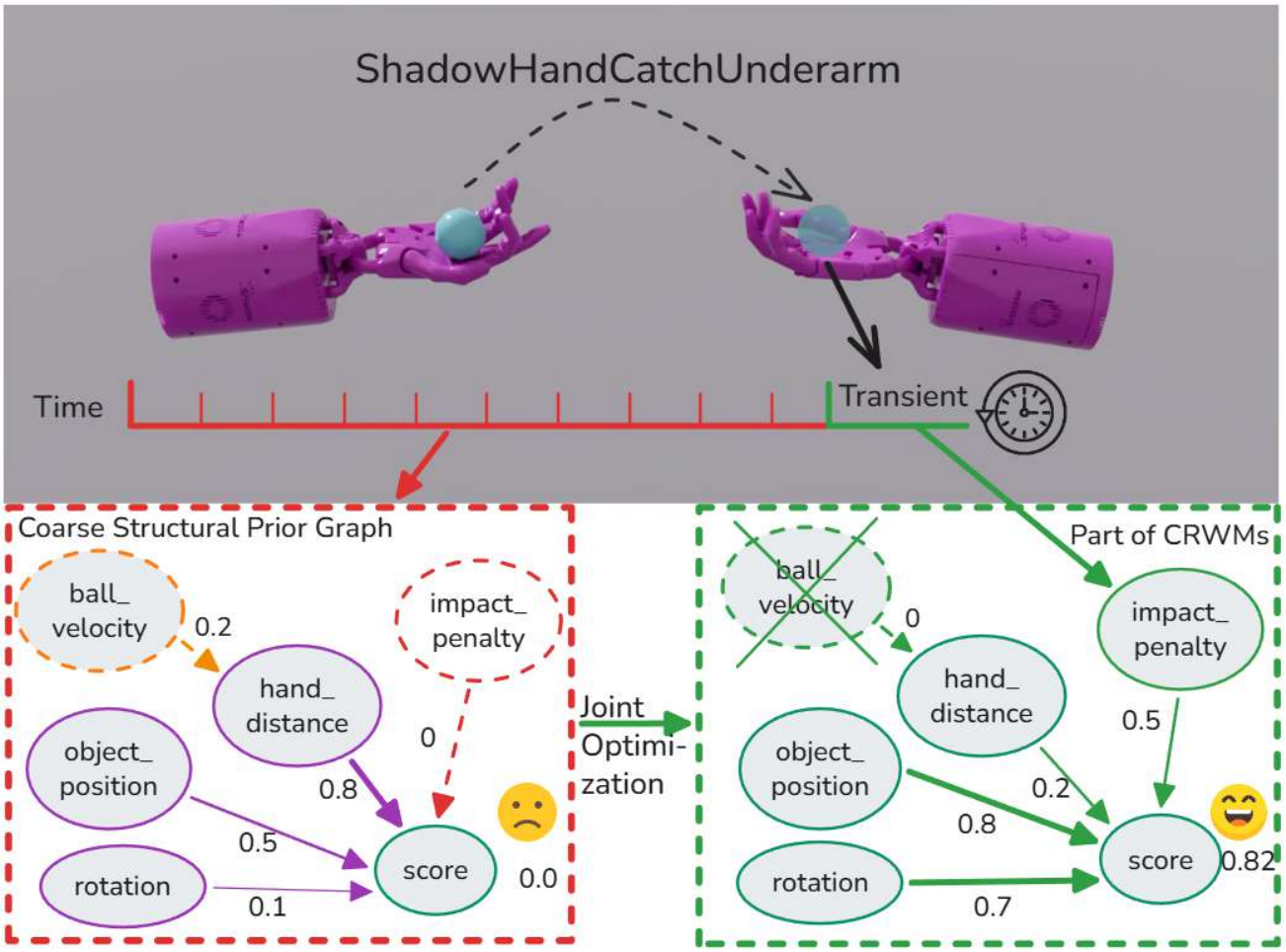}
\caption{\textbf{Joint Optimization for CRWM Distillation.} 
The top panel shows a physical trajectory in the \textit{ShadowHandCatchUnderarm} task, where ball-hand contact induces a state-dependent transient physical interaction. 
The bottom-left panel shows the coarse structural prior \(M_{prior}\) before joint optimization, which contains biased causal effects, a hallucinated \texttt{ball\_velocity} node, and a missing contact-related \texttt{impact\_penalty} relation. 
The bottom-right panel shows the final CRWM after joint optimization, where the hallucinated node is removed and the contact-related relation is included when supported by transition data. 
Overall, EMD accounts for state-dependent transient physical interactions during reconstruction, while Confidence-Aware Soft Fusion calibrates the coarse structural prior using observed transitions and structural confidence.}
\label{fig:method-B}
\end{figure}

The goal of this stage is to distill the final CRWM skeleton $M_{inv}^{*}$ by refining the coarse structural prior with micro-level trajectories. 
This stage addresses two issues left by Section~\ref{subsec:prior}. 
First, a single static topology can confuse persistent reward-relevant relations with state-dependent transient physical interactions, such as contact-induced effects that appear only under specific physical states. 
Fig.~\ref{fig:method-B} illustrates this limitation: the coarse structural prior incorrectly retains the transient correlate \texttt{ball\_velocity} while missing the contact-dependent \texttt{impact\_penalty} relation. 
A simplified analysis of this static-topology limitation is provided in the Supplementary Material, Sec.~S-B. 
Following the Independent Causal Mechanisms (ICM) principle~\cite{peters2017elements}, we therefore model reward-relevant transitions using a learnable causal skeleton together with state-dependent transient interactions. 
Second, as discussed in Section~\ref{subsec:prior}, the coarse structural prior $M_{prior}$ may contain structural hallucinations and biased causal effects, so directly enforcing it can bias the final CRWM.

To address these coupled challenges, we propose a joint continuous optimization framework that accounts for state-dependent transient physical interactions while calibrating the coarse structural prior.
This module optimizes the learnable causal skeleton $M_{inv}$ and obtains its converged form $M_{inv}^{*}$ as the final CRWM.
The CRWM distillation is driven by the unified objective:
\begin{equation}
\begin{split}
    \min_{M_{inv}, \theta} \mathcal{L} &= \mathbb{E}_{(s_{t-1}, s_t) \sim \mathcal{D}_{micro}} \left[ \mathcal{L}_{MSE}\Big(s_t, M_{inst}(s_{t-1}) s_{t-1}\Big) \right] \\
    &\quad + \lambda h(M_{inv}) + \mathcal{L}_{soft}.
\end{split}
\label{eq:objective}
\end{equation}
Here, $M_{inv}$ denotes the learnable causal skeleton whose converged form $M_{inv}^{*}$ is retained as the CRWM, and $\theta$ parameterizes the state-dependent transient component introduced by EMD. 
The reconstruction term compares each transition pair $(s_{t-1},s_t)\sim\mathcal{D}_{micro}$ with the locally linear prediction $M_{inst}(s_{t-1})s_{t-1}$, where $M_{inst}(s_{t-1})$ is the state-dependent instantaneous topology constructed below. 
The coefficient $\lambda$ weights the acyclicity regularization $h(M_{inv})$ (App.~\ref{app:hyperparameters}), and $\mathcal{L}_{soft}$ denotes the Confidence-Aware Soft Fusion loss for calibrating the coarse structural prior.

\textcolor{black}{Since \(M_{inv}\) is retained as the reward-relevant causal skeleton for reward generation, we regularize it as a directed acyclic structure to avoid circular dependencies in the causal prior provided to the LLM. 
This acyclicity constraint is applied only to $M_{inv}$, rather than the full time-lagged physical dynamics, because feedback loops and inertial couplings are handled through the state-dependent instantaneous topology during reconstruction. 
Specifically, we impose the DAGMA acyclicity constraint~\cite{bello2022dagma} on $M_{inv}$:}
\begin{equation}
    h(M_{inv}) = -\log \det (\alpha I - M_{inv} \odot M_{inv}) + d \log \alpha,
\label{eq:dagma}
\end{equation}
where $\alpha>0$ is the DAGMA scalar hyperparameter, $I$ is the identity matrix, $d=|\mathcal{V}|$ is the number of structural variables, and $\odot$ denotes the Hadamard product. 
Following the standard DAGMA setting, we set $\alpha=1$. 
The effect of this acyclicity regularization is further examined in Supplementary Material, Sec. S-D.
The objective in Eq.~\ref{eq:objective} is implemented by two modules introduced below: EMD constructs the state-dependent instantaneous topology $M_{inst}(s_{t-1})$, while Confidence-Aware Soft Fusion defines $\mathcal{L}_{soft}$ to calibrate the coarse structural prior.

\textbf{Explicit Mechanism Decoupling (EMD).} 
EMD constructs the state-dependent instantaneous topology $M_{inst}(s_{t-1})$ used in the reconstruction term of Eq.~\ref{eq:objective}.
Instead of using a single static topology for all structural-variable transitions, EMD constructs a state-dependent instantaneous topology by combining the learnable causal skeleton with state-dependent transient physical interactions.
Specifically, $M_{inv}\in\mathbb{R}^{d\times d}$ denotes the learnable causal skeleton retained for reward generation, and $M_{trans}(s_{t-1};\theta)\in\mathbb{R}^{d\times d}$ denotes the transient component parameterized by an MLP, where $s_{t-1}$ is the previous structural-variable state and $\theta$ denotes the MLP parameters. 
The instantaneous topology is constructed as
\begin{equation}
    M_{inst}(s_{t-1}) = M_{inv} \odot \Big( \mathbf{1}_{d \times d} + M_{trans}(s_{t-1}; \theta) \Big),
\label{eq:emd}
\end{equation}
where $\mathbf{1}_{d\times d}$ is an all-ones matrix, and $\odot$ denotes the Hadamard product.
This multiplicative form allows $M_{trans}(s_{t-1};\theta)$ to adjust candidate-edge strengths across physical states while anchoring the optimization to $M_{inv}$.

By constructing $M_{inst}(s_{t-1})$ at each step, EMD allows $M_{trans}(s_{t-1};\theta)$ to account for short-lived variations during reconstruction, while $M_{inv}$ remains the skeleton retained for reward generation. 
Since $M_{inst}(s_{t-1})$ is multiplicatively gated by $M_{inv}$, the transient component cannot freely introduce unrelated edges outside the learnable skeleton. 
This design helps stabilize the extracted reward-relevant skeleton while accounting for state-dependent transient interactions in $\mathcal{D}_{micro}$. 
The effect of this decoupled formulation over static causal modeling is analyzed in the Supplementary Material, Sec.~S-B, and the overall contribution of EMD is evaluated in Section~\ref{subsec:ablation} (CRWM w.o. EMD).

\textbf{Confidence-Aware Soft Fusion.} 
While EMD accounts for state-dependent transient physical interactions in the reconstruction term, a reliable causal skeleton also requires calibrating the coarse structural prior $M_{prior}$.
Using $M_{prior}$ as a rigid structural mask would force the optimization to inherit its structural hallucinations and biased causal effects.
We therefore define the Confidence-Aware Soft Fusion loss as
\begin{equation}
    \mathcal{L}_{soft} = \gamma \| M_{conf} \odot (M_{inv} - M_{prior}) \|_F^2, 
\label{eq:soft_fusion}
\end{equation} 
where $\gamma$ controls the loss strength, $M_{conf}$ is the structural confidence matrix, and $\|\cdot\|_F$ denotes the Frobenius norm.
The setting of $\gamma$ is discussed in App.~\ref{app:hyperparameters}, with its sensitivity analysis shown in Fig.~\ref{fig:gama_compare}.

This term penalizes deviations from the coarse structural prior in a confidence-aware manner, imposing stronger regularization on high-confidence edges while keeping low-confidence edges more flexible.
Together with the reconstruction term in Eq.~\ref{eq:objective}, Soft Fusion softly calibrates the coarse structural prior using micro-level structural-variable trajectories.
Thus, Soft Fusion allows $M_{inv}$ to deviate from unreliable parts of $M_{prior}$ while retaining high-confidence structural guidance. 
The qualitative changes before and after joint optimization are illustrated in Fig.~\ref{fig:method-B}, and the contribution of Soft Fusion is evaluated in Section~\ref{subsec:ablation} (CRWM w.o. Soft-Fusion).

Since standard unconstrained optimizers cannot directly enforce the equality constraint $h(M_{inv})=0$, we solve Eq.~\ref{eq:objective} using the Augmented Lagrangian Method (ALM)~\cite{hestenes1969multiplier}. 
This optimization balances transition reconstruction, DAG acyclicity, and confidence-aware prior calibration. 
Upon convergence, the optimized causal skeleton is extracted as $M_{inv}^{*}$, which is retained as the final CRWM. 
A visualization of the extracted CRWM is provided in Fig.~\ref{fig:global_crwm}.

\subsection{Causal Automated Reward Design (Causal-ARD)}
\label{subsec:zero_shot_generation}
After CRWM distillation, the optimized skeleton $M_{inv}^{*}$ must be converted into a form that can guide executable reward synthesis. 
Causal-ARD addresses this interface problem by translating the continuous CRWM matrix into a structured causal topology prompt and using it to guide LLM-based reward code generation. 
This stage connects the learned causal structure with zero-shot reward generation: the CRWM provides reward-relevant causal links, while Causal-ARD uses these links to prune spurious reward terms.

\begin{figure}[htbp]
\centering
\includegraphics[width=0.9\linewidth]{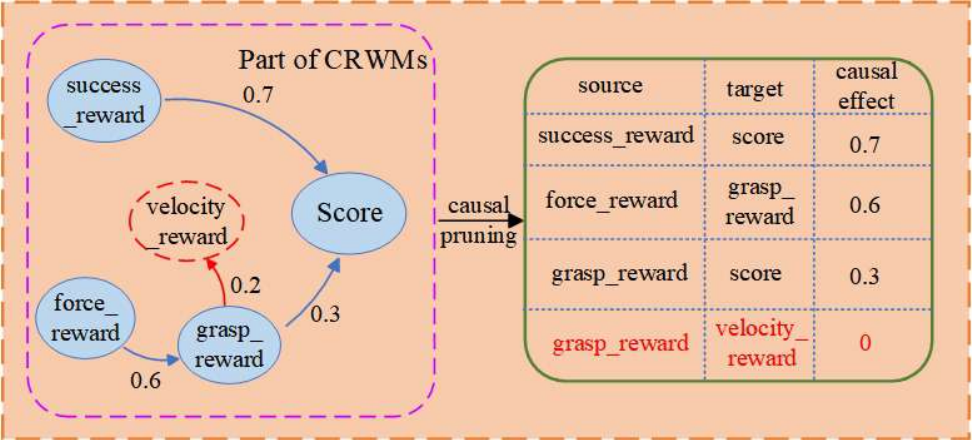}
\caption{\textbf{Translation of the CRWM into explicit topological triplets.} 
The extracted causal edges from the CRWM are serialized into a structured tabular format with \textit{source}, \textit{target}, and \textit{causal effect}. 
These triplets are embedded into the causal topology prompt to guide explicit causal pruning during zero-shot reward generation.}
\label{fig:graph_translation}
\end{figure}

\textbf{Causal Graph to Prompt Translation.} 
Since LLMs are better suited to structured text than continuous numerical matrices due to tokenization inefficiencies and spatial reasoning limits~\cite{wang2023can, fatemi2024talk}, Causal-ARD employs a deterministic serialization protocol.
To extract active causal edges, we apply one-dimensional 2-means clustering to the absolute edge causal effects $\mathcal{W} = \{ |M_{inv}^{*,i,j}| \}_{i,j=1}^d$, where $j$ and $i$ denote the source and target nodes, respectively.
This partitions $\mathcal{W}$ into noise and signal clusters by minimizing intra-cluster variance.
We set the extraction threshold as the midpoint of their centroids, $\tau = \frac{\mu_{noise} + \mu_{signal}}{2}$, where $\mu_{noise}$ and $\mu_{signal}$ are the corresponding cluster means, separating dominant causal effects from numerical noise.
The extracted active edges are serialized into explicit topological triplets, i.e., \textit{source}, \textit{target}, and \textit{causal effect}, as illustrated in Fig.~\ref{fig:graph_translation}.
These triplets are embedded into the system instructions to construct the \textit{Causal Topology Prompt}, with the full prompt provided in App.~\ref{app:prompt}.

\textbf{Causal Pruning.} 
Conditioned on the \textit{Causal Topology Prompt}, the LLM generates the reward function under explicit causal constraints.
The prompt identifies reward-relevant causal links and spurious correlates that should not contribute to the reward.
During code generation, the LLM performs causal pruning by assigning zero coefficients to reward terms associated with identified spurious correlates.
As highlighted by the red annotations in Fig.~\ref{fig:graph_translation}, the reward component of an identified spurious node is set to zero, preventing spurious correlates from entering the generated reward code.

\textbf{Executable Output of Causal-ARD.} 
Ultimately, for an unseen target task, the LLM receives a structured input context consisting of (1) the textual task description, (2) the raw observation space, and (3) the Causal Topology Prompt.
\textcolor{black}{Notably, reward components whose required embodiment-specific variables are absent from the observation space, such as finger-related variables in ManiSkill2 or real-world tasks, are not used by the LLM during reward design.}
Guided by this causal prior, the LLM generates a Python reward function in a single pass.
By producing executable code for the unseen environment without iterative trial-and-error, this zero-shot process avoids computationally expensive evolutionary RL loops.
The generated reward function can be directly executed by the underlying RL algorithm, as validated in Section~\ref{exp:main_results}.
The overall procedure, including CRWM distillation and Causal-ARD, is summarized in Alg.~\ref{alg:crwm_short}.

\begin{algorithm}[htbp]
\caption{CRWM distillation and Causal-ARD}
\label{alg:crwm_short}
\begin{algorithmic}[1]
\renewcommand{\algorithmicrequire}{\textbf{Input:}}
\renewcommand{\algorithmicensure}{\textbf{Output:}}

\REQUIRE An interventional dataset $\mathcal{D}_{off} = \{\mathcal{D}_{macro}, \mathcal{D}_{micro}\}$, a task description $\mathcal{T}$, observation nodes $\mathcal{V}$, a causal foundation model $\mathcal{M}_{f}$, and an LLM $\mathcal{M}_{LLM}$
\ENSURE A zero-shot reward function represented in Python code $R^*$

\STATE Extract the structural prior and confidence: $M_{prior}, M_{conf} \leftarrow \mathcal{M}_{f}(\mathcal{V}, \mathcal{D}_{macro})$
\STATE Initialize the invariant skeleton $M_{inv} \in \mathbb{R}^{d \times d}$ and the transient modulator parameters $\theta$

\WHILE{not converged}
    \STATE Sample transition batches $(s_{t-1}, s_t) \sim \mathcal{D}_{micro}$
    \STATE Construct the instantaneous topology: $M_{inst}(s_{t-1}) \leftarrow M_{inv} \odot \Big( \mathbf{1}_{d \times d} + M_{trans}(s_{t-1}; \theta) \Big)$
    \STATE Compute the reconstruction loss: $\mathcal{L}_{MSE} \leftarrow \| s_t - M_{inst}(s_{t-1}) s_{t-1} \|_2^2$
    \STATE Compute the joint objective:
    \STATE \quad $\mathcal{L}_{\text{total}} \leftarrow \mathcal{L}_{MSE} + \lambda h(M_{inv}) + \gamma \| M_{conf} \odot (M_{inv} - M_{prior}) \|_F^2$
    \STATE Update $M_{inv}$ and $\theta$ via $\nabla \mathcal{L}_{\text{total}}$ using the ALM
\ENDWHILE

\STATE Extract the invariant skeleton: $M_{inv}^* \leftarrow M_{inv}$
\STATE Compute an adaptive threshold $\tau$ via 1D 2-means clustering on $\{|M_{inv}^{*, i,j}|\}$
\STATE Extract the active causal triplets: $\mathcal{E} \leftarrow \{ (v_i, v_j, M_{inv}^{*, i,j}) \mid |M_{inv}^{*, i,j}| \ge \tau \}$
\STATE Serialize $\mathcal{E}$ to construct the causal topology prompt $\mathcal{P}_{causal}$
\STATE Causal reward generation: $R^* \leftarrow \mathcal{M}_{LLM}(\mathcal{T}, \mathcal{V}, \mathcal{P}_{causal})$

\RETURN $R^*$
\end{algorithmic}
\end{algorithm}

%% file: TNNLS/sec/experiment/main_exp.tex
\section{experiments}

In this section, we evaluate the efficacy of the CRWM through extensive experiments on zero-shot task performance, computational efficiency, and generalization across diverse environments and robotic morphologies.
We further demonstrate its practical feasibility by zero-shot synthesizing reward functions for real-world robotic tasks and validating their alignment offline.
To isolate architectural gains from underlying semantic capabilities, all LLM-driven methods, including ours and the baselines, use the same foundation model, \texttt{deepseek-v3-2-251201}~\cite{liu2024deepseek} (see Supplementary Material, Sec. S-F-A for results with other open-source LLMs).
We also use the pre-trained causal foundation model LimiX~\cite{zhang2025limix} across all experiments, with comparisons to alternative causal models provided in Supplementary Material, Sec. S-F-B.

\subsection{Experimental Setup}

\textcolor{black}{\textbf{Dexterity (Isaac Gym)~\cite{chen2022towards}:} We evaluate our approach on 20 tasks from the Shadow Hand manipulation suite~\cite{Ma2023EurekaHR} (detailed in Supplementary Material, Sec. S-A). These tasks feature high-dimensional state-action spaces and dense transient physical interactions. The abundance of redundant kinematic variables provides a testbed to evaluate the zero-shot task performance of the CRWM by assessing its causal pruning against spurious correlates.}

\textcolor{black}{\textbf{ManiSkill2:} To validate generalization across diverse environments, we utilize the ManiSkill2 benchmark~\cite{gu2023maniskill2} (detailed in Supplementary Material, Sec. S-A). This suite of robotic manipulation tasks serves to verify that the zero-shot task performance enabled by the CRWM remains consistent across varying task and robotic morphologies.}

\subsection{Baselines}

\textbf{Zero-Shot LLM (ZS-LLM):} A purely semantic baseline. Following established protocols~\cite{Ma2023EurekaHR}, we prompt the LLM with raw environment source code and task descriptions to synthesize reward functions in a single pass. This represents foundation model capabilities relying solely on semantic correlations, without explicit causal guidance or environmental feedback.

\textbf{Eureka~\cite{Ma2023EurekaHR}:} An evolutionary ARD framework utilizing LLMs to heuristically mutate and refine reward functions through computationally intensive trial-and-error interactions and environmental feedback.

\textbf{URDP~\cite{yang2025uncertainty}:} A bi-level optimization framework integrating LLM-driven structural refinement with uncertainty-aware Bayesian optimization. Although highly sample-efficient compared to Eureka, this posterior-correction method fundamentally lacks explicit causal mechanism decoupling.

\subsection{Evaluation Metrics}
We evaluate the automated reward design pipelines across two metrics to quantify both task performance and computational footprint. \textbf{(i) Success Rate (SR):} The primary indicator of task mastery and policy robustness. To ensure a fair comparison with the baseline methods, the SR for ManiSkill2 tasks is calculated utilizing the last 50\% of test episodes from each evaluation rollout, whereas full test results are aggregated for the Dexterity suite. \textbf{(ii) Evolutionary Search Iterations (ESI):} The number of feedback-driven reward refinement rounds after the initial generation batch. 
For zero-shot methods, all candidate reward functions are generated before environment evaluation. 
Evaluating this fixed candidate set to report task performance does not increase ESI, since the evaluation results are not used to mutate rewards, update prompts, or trigger another generation round.

\subsection{Results}
\label{exp:main_results}

\begin{figure*}[t] 
\centering

\subfloat[ShadowHandKettle\label{fig:kettle}]{
    \includegraphics[width=0.318\textwidth]{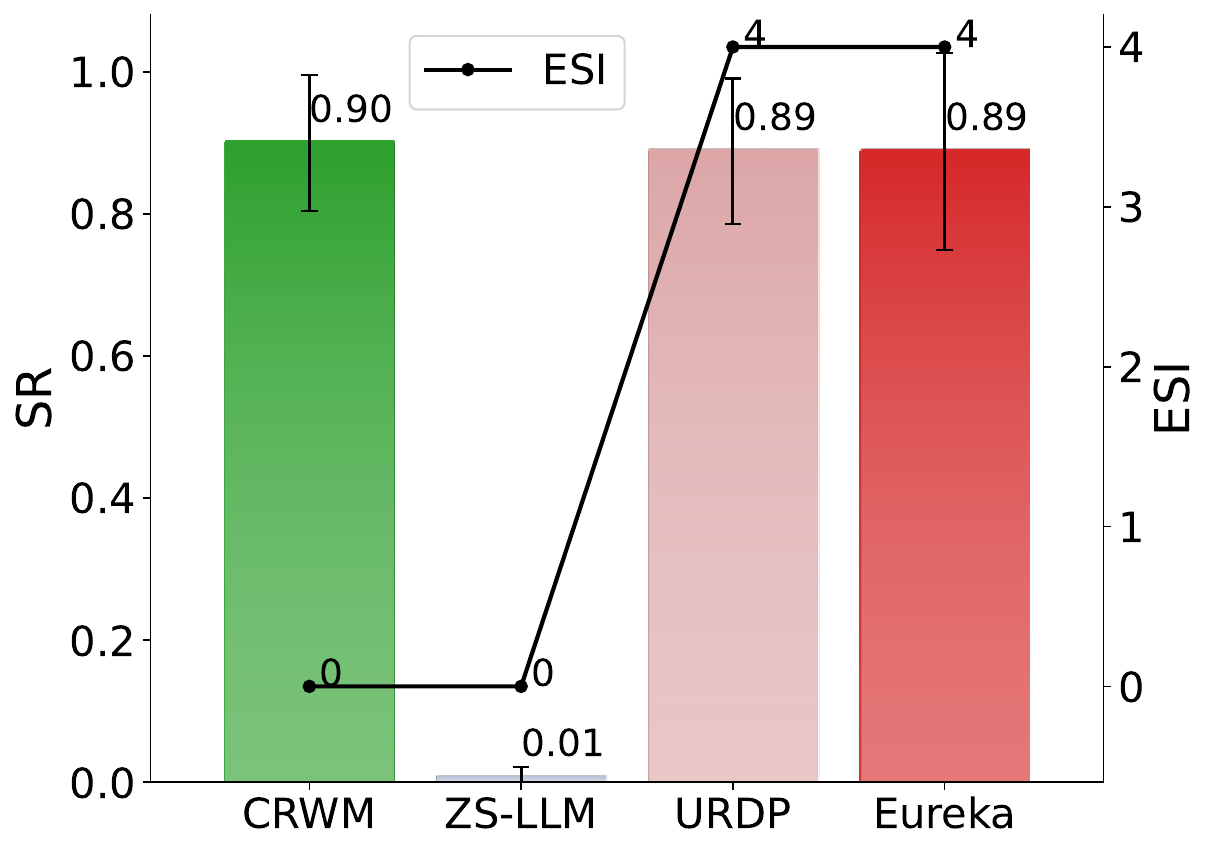} 
}\hfill 
\subfloat[ShadowHandCatchAbreast\label{fig:row1_2}]{
    \includegraphics[width=0.318\textwidth]{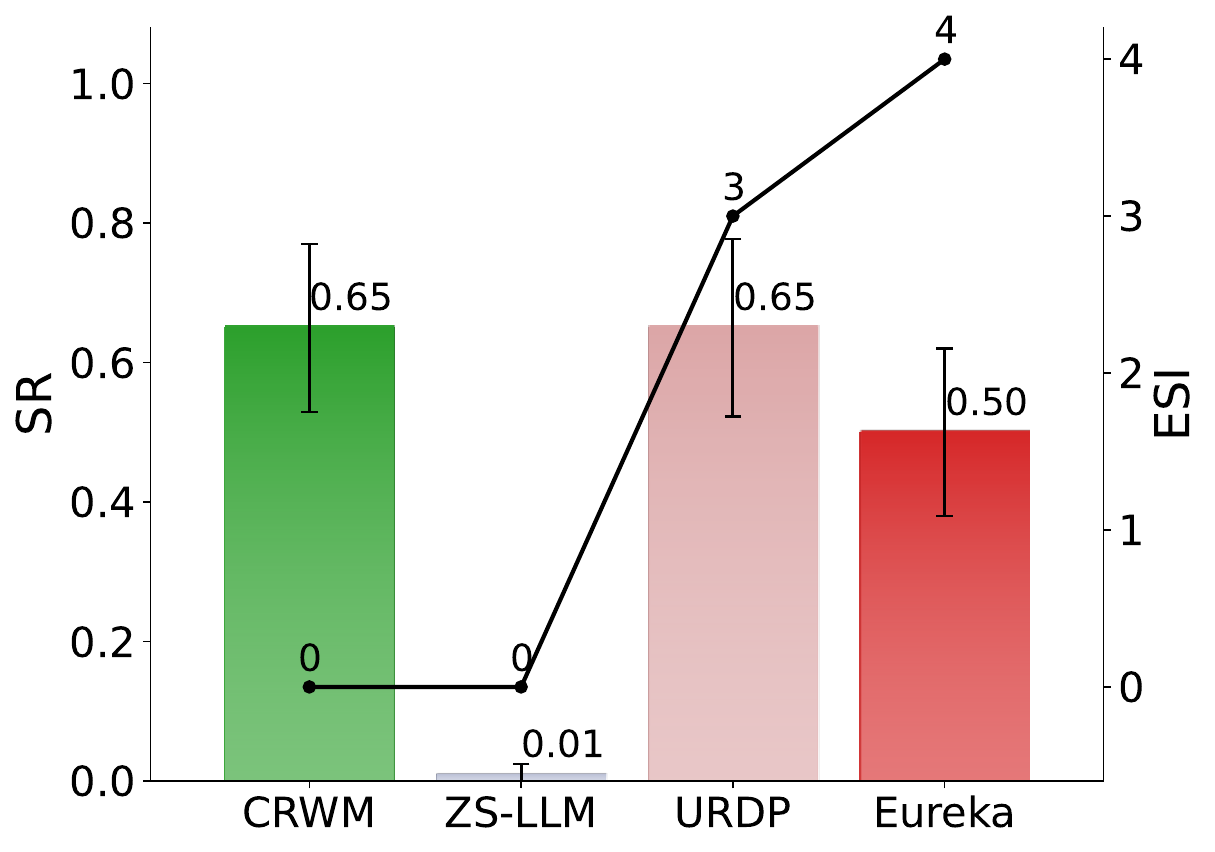}
}\hfill
\subfloat[ShadowHandCatchOver2Underarm\label{fig:row1_3}]{
    \includegraphics[width=0.318\textwidth]{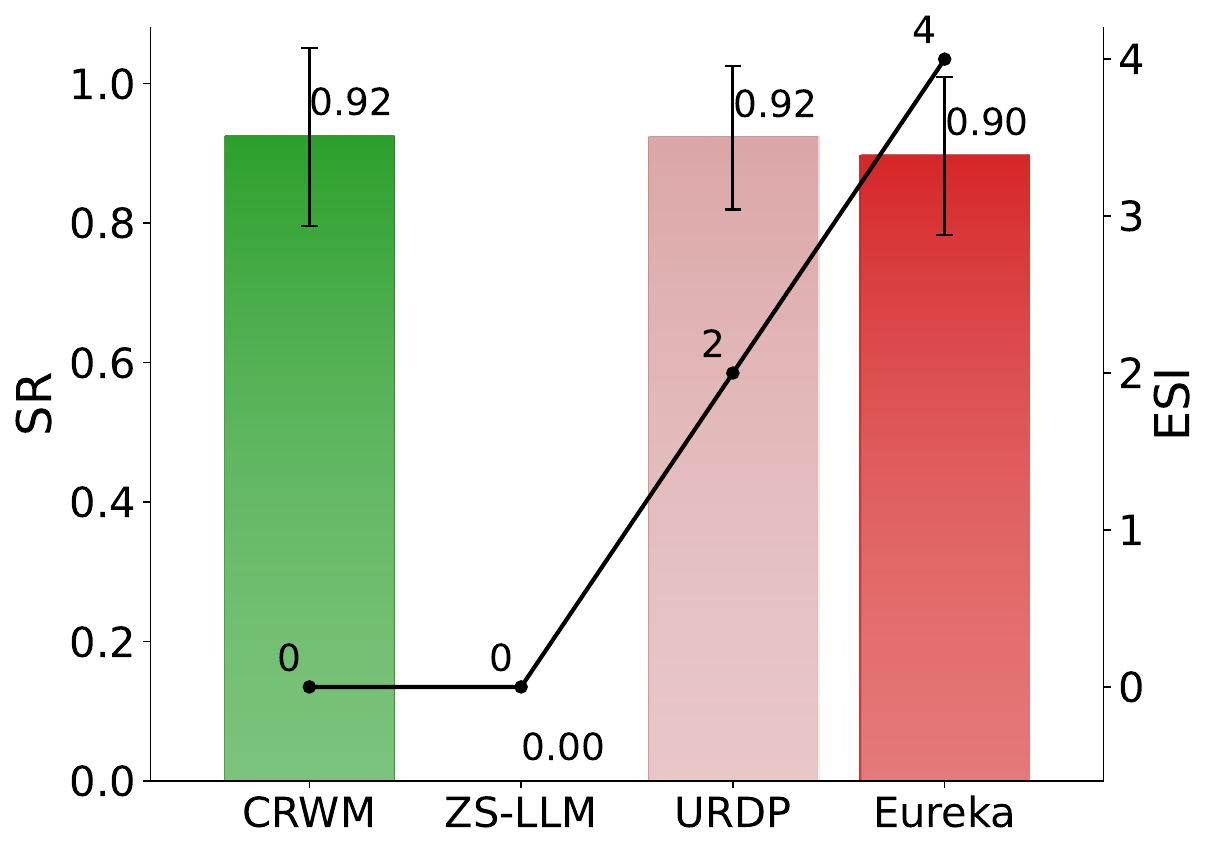}
}

\vspace{0ex} 

\subfloat[ShadowHandDoorCloseOutward\label{fig:row2_1}]{
    \includegraphics[width=0.318\textwidth]{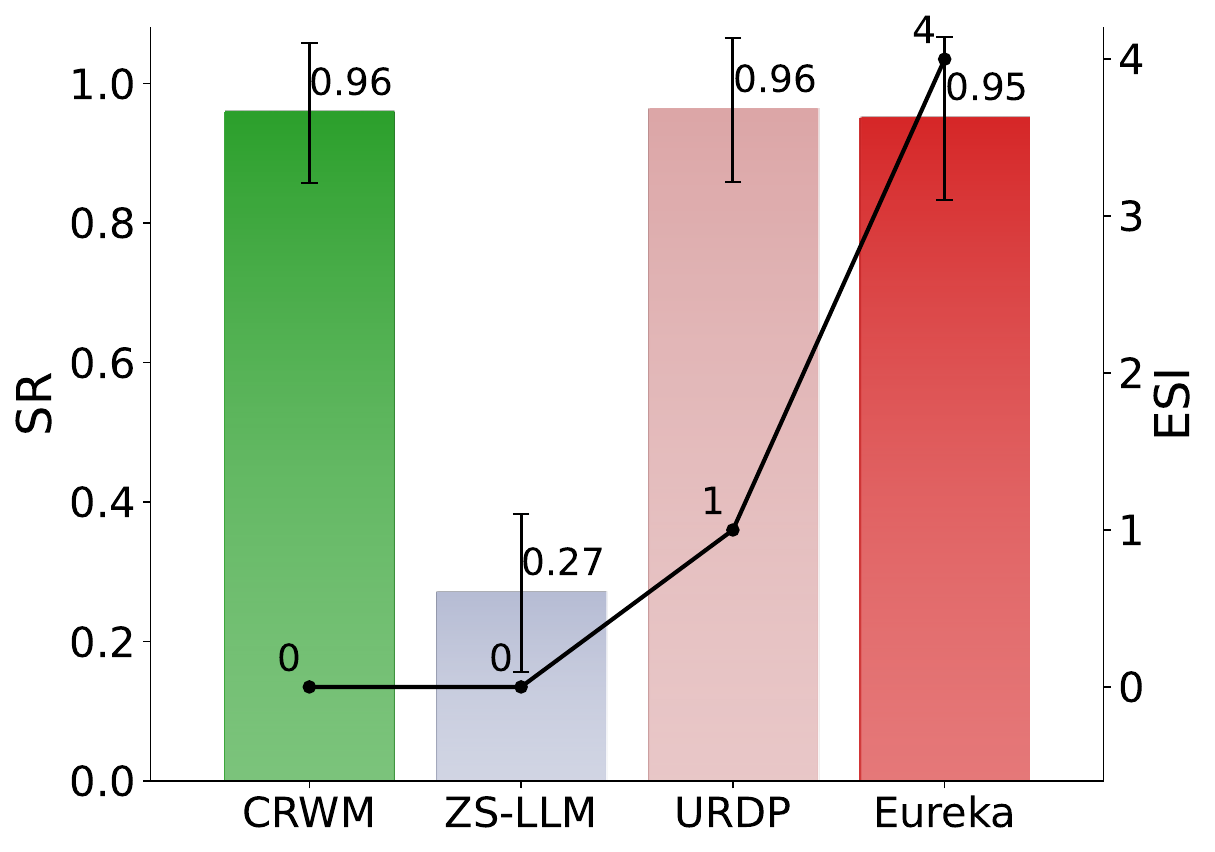}
}\hfill
\subfloat[ShadowHandPen\label{fig:row2_2}]{
    \includegraphics[width=0.318\textwidth]{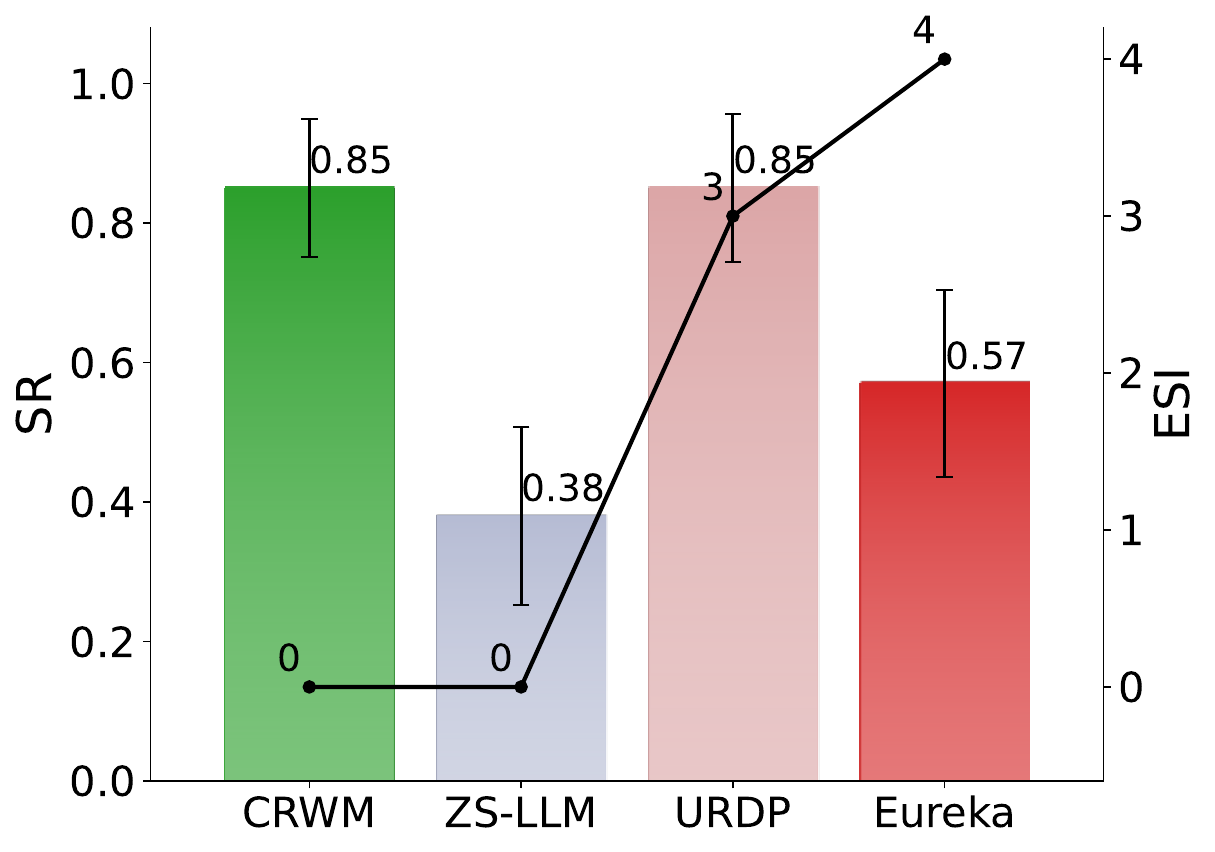}
}\hfill
\subfloat[ShadowHandLiftunderarm\label{fig:row2_3}]{
    \includegraphics[width=0.318\textwidth]{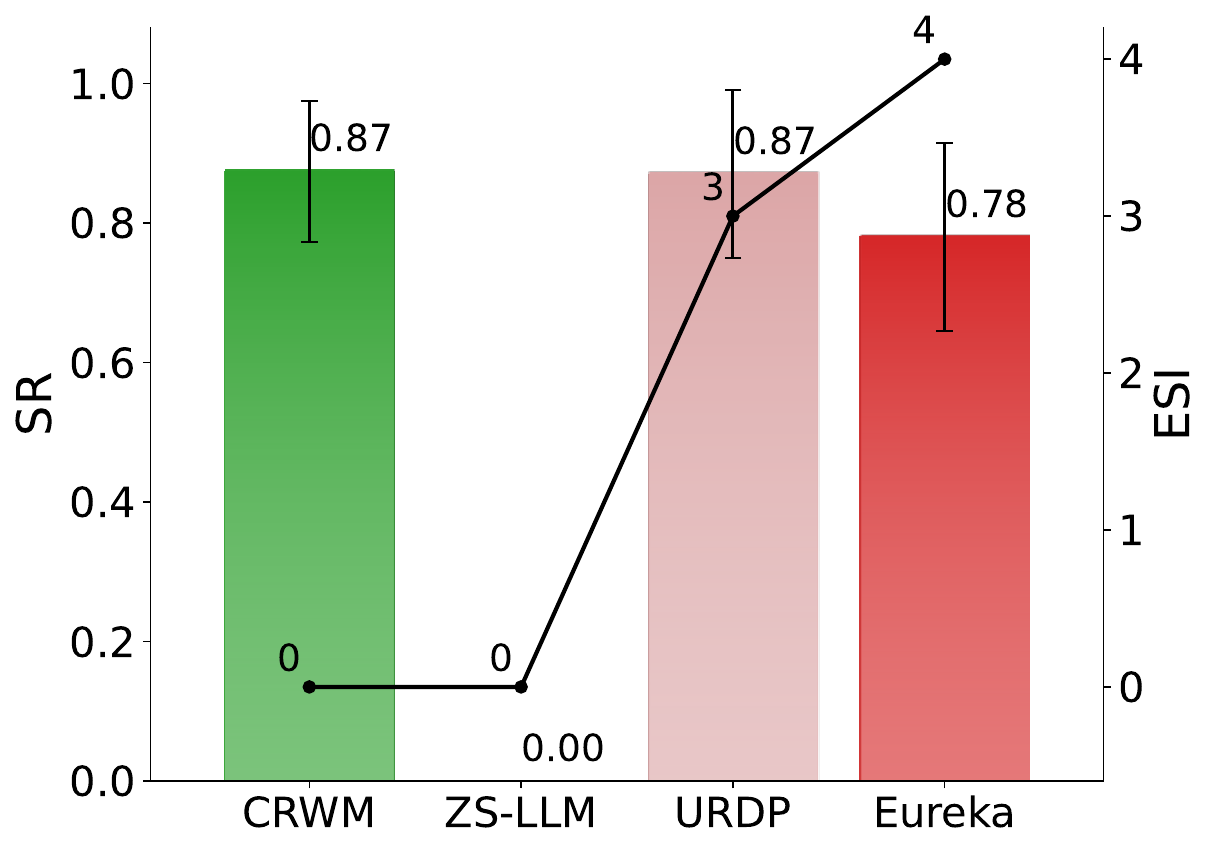}
}

\caption{\textbf{Zero-shot task performance across 6 unseen dexterous manipulation tasks.} 
The bar charts, corresponding to the left y-axis, represent the final SR of the generated reward functions. 
The solid black line, corresponding to the right y-axis, denotes the ESI required to achieve the reported performance. 
For readability, ``CRWM'' denotes our full zero-shot reward generation pipeline, where reward generation is guided by the CRWM. 
Across all unseen tasks, this pipeline achieves performance comparable to state-of-the-art iterative methods with zero evolutionary iterations ($\text{ESI}=0$), avoiding the costly trial-and-error loops required by Eureka and URDP. 
In contrast, the unguided ZS-LLM baseline exhibits inferior performance, particularly on dynamic and long-horizon tasks.}
\label{fig:main_results}
\end{figure*}

To evaluate the zero-shot task performance enabled by the CRWM, we conduct experiments on 20 Dexterity continuous control tasks.
We use a 14/6 task split, where 14 tasks serve as the pre-training set for distilling the CRWM and the remaining 6 unseen tasks form the zero-shot evaluation suite.
This split tests whether reward-relevant structures distilled from pre-training tasks can support reward generation on held-out tasks without task-specific feedback.
On these unseen tasks, we compare our full zero-shot reward generation pipeline against the semantic baseline ZS-LLM and iterative ARD frameworks, Eureka and URDP.
In this pipeline, Causal-ARD generates reward functions guided by the CRWM; for consistency with Fig.~\ref{fig:main_results}, we denote the full pipeline as CRWM in the experimental plots.
During evaluation, each round prompts the LLM to generate 16 candidate reward functions in parallel.
Both CRWM and ZS-LLM operate in the zero-shot setting ($\text{ESI}=0$), generating executable reward functions without subsequent physical trial-and-error, whereas the reported ESI for Eureka and URDP denotes the evolutionary iterations required to reach their converged SR.
The offline pre-training cost for constructing the CRWM is reported separately in App.~\ref{app:compute_cost}.

Fig.~\ref{fig:main_results} compares task performance and computational cost across the 6 unseen tasks using a dual-axis design.
Bars on the left y-axis report the final SR ($\uparrow$) achieved by the generated rewards, while lines on the right y-axis indicate the ESI ($\downarrow$) required to reach that performance.
CRWM achieves state-of-the-art-level task performance in a single reward-generation pass with no evolutionary iterations ($\text{ESI}=0$).
This comparison yields three key insights as follows.

\textbf{Causal Priors Improve Zero-Shot LLM Reward Generation.} 
The weak performance of ZS-LLM shows that semantic reasoning alone is insufficient for reliable reward generation in physical control tasks.
Without explicit causal guidance, ZS-LLM often introduces semantically plausible but causally irrelevant reward terms, leading to near-zero SRs on challenging unseen tasks such as \textit{ShadowHandLiftUnderarm} and \textit{ShadowHandKettle}.
In contrast, the CRWM-guided reward generation uses the causal topology to prune spurious reward components before policy training, providing necessary causal guidance for robust zero-shot reward generation..

\textbf{SOTA-Level Zero-Shot Performance.} 
Across the six unseen dexterous manipulation tasks, the CRWM-guided reward generation achieves performance comparable to iterative ARD baselines.
For example, it attains SRs of $0.92$ and $0.87$ on the challenging \textit{Catch Over 2 Underarm} and \textit{Lift Underarm} tasks, respectively.
These results outperform the multi-iteration Eureka baseline and match URDP, which requires up to 3 evolutionary search iterations to converge.
Notably, this is achieved in a zero-shot setting, without feedback-driven reward refinement or additional generation rounds.

\textbf{Zero-Shot Efficiency on Unseen Tasks.} 
Fig.~\ref{fig:main_results} also highlights the computational efficiency of our method on the unseen evaluation suite.
Iterative ARD baselines require repeated evolutionary search to reach their converged SR, with URDP requiring 1--4 ESIs and Eureka requiring 4 ESIs across the evaluated tasks.
Each ESI trains and evaluates candidate reward functions through large-scale RL environment interactions, which is time-consuming in continuous control tasks and takes approximately one hour on eight NVIDIA A800 GPUs in our setting.
In contrast, the CRWM-guided reward generation achieves competitive task performance with $\text{ESI}=0$, generating executable rewards for unseen tasks without feedback-driven refinement.

%% file: TNNLS/sec/experiment/ablation.tex
\subsection{Ablation Experiments}
\label{subsec:ablation}
\begin{figure*}[t] 
\centering
\includegraphics[width=1\textwidth]{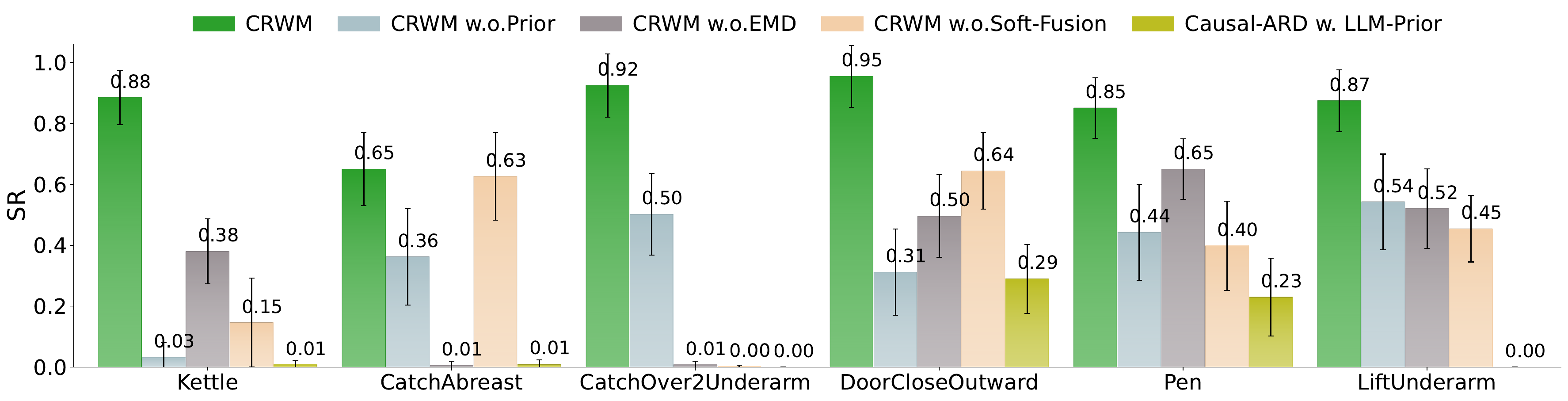} 
\caption{\textbf{Ablation studies across 6 unseen zero-shot tasks.} 
Performance drops when key CRWM components are removed or when the learned topology is replaced by an LLM-inferred semantic prior.}
\label{fig:ablation}
\end{figure*}

M construction, we evaluate three ablated variants across the unseen zero-shot tasks (Fig.~\ref{fig:ablation}). 
Each variant is used within the same Causal-ARD pipeline to measure its impact on task performance:
1) \textbf{CRWM w.o. Prior}: removes the coarse structural prior, relying only on offline transition data.
2) \textbf{CRWM w.o. EMD}: removes the state-dependent interaction term, reducing the model to a single static structure.
3) \textbf{CRWM w.o. Soft-Fusion}: replaces the soft regularization with a rigid structural mask.
4) \textcolor{black}{\textbf{Causal-ARD w. LLM-Prior}: uses the same Causal-ARD reward generation procedure as CRWM, but replaces the CRWM topology with a semantic causal topology inferred by the LLM from the target task description and observation space.}

\textbf{Necessity of the Structural Prior.} 
The \textit{CRWM w.o. Prior} variant consistently underperforms across all unseen tasks, indicating that offline transition data alone is insufficient for stable structure estimation.
Without structural guidance, the model tends to converge to suboptimal solutions in the high-dimensional space, causing the average SR to drop from $0.85$ to $0.36$, with the largest drop exceeding $0.7$.
These results show that the structural prior provides essential guidance for learning a reliable causal structure.

\textbf{Effect of EMD.} 
Removing EMD leads to noticeably unstable performance, with the \textit{CRWM w.o. EMD} variant struggling to maintain consistent results across tasks.
Without EMD, the model relies on a single static structure to explain all transitions, reducing its ability to capture data variations.
As a result, the average SR drops from $0.85$ to $0.35$, with the largest drop exceeding $0.8$.
This suggests that EMD improves the stability of the learned structure.

\textbf{Soft Fusion Enables Adaptive Use of the Prior.} 
When the structural prior is imposed as a rigid constraint, the \textit{CRWM w.o. Soft-Fusion} variant shows clear performance degradation across multiple tasks.
In this setting, the model becomes less adaptive to observed transitions, leading to suboptimal structure estimation.
Empirically, the average SR drops from $0.85$ to $0.38$, with the largest drop reaching around $0.75$.
These results indicate that Soft Fusion enables flexible integration between the prior structure and data, which is important for strong performance.

\textcolor{black}{\textbf{LLM-Inferred Priors Are Not Sufficient.} 
The \textit{Causal-ARD w. LLM-Prior} variant performs substantially worse than CRWM across unseen tasks.
Although it uses the same Causal-ARD reward generation procedure, its causal topology is inferred only from the target task description and observation space, without offline interventional data or micro-level trajectories.
Thus, the inferred semantic topology can be incomplete or misleading, causing causal pruning to remove useful reward terms or retain spurious correlations.
Empirically, its average SR drops from $0.85$ to $0.09$, with the largest drop exceeding $0.9$.
This indicates that the main performance gain comes from the CRWM itself: pairing the same Causal-ARD procedure with an LLM-inferred topology leads to substantial performance degradation.}

\subsection{Cross-Embodiment and Cross-Environment Generalization}
\label{subsec:generalization}
We evaluate the cross-embodiment and cross-environment generalization ability of the CRWM.
Unlike the 14/6 split for unseen-task evaluation within Dexterity, this experiment uses the CRWM constructed from the full 20-task Dexterity suite.
The goal is to examine whether structural variables and causal relations learned from Dexterity can transfer to a different manipulation benchmark.
Specifically, we apply the CRWM to three ManiSkill2 manipulation tasks whose reward-related variables are represented by existing CRWM structural variables, as detailed under \textit{Compositional Coverage} in App.~\ref{app:task_details}.
This setting introduces a clear domain shift from a high-DoF multi-fingered ShadowHand to single-arm manipulators in a different physics engine.

\begin{figure*}[t]
\centering
\subfloat[PickCube]{
    \includegraphics[width=0.318\textwidth]{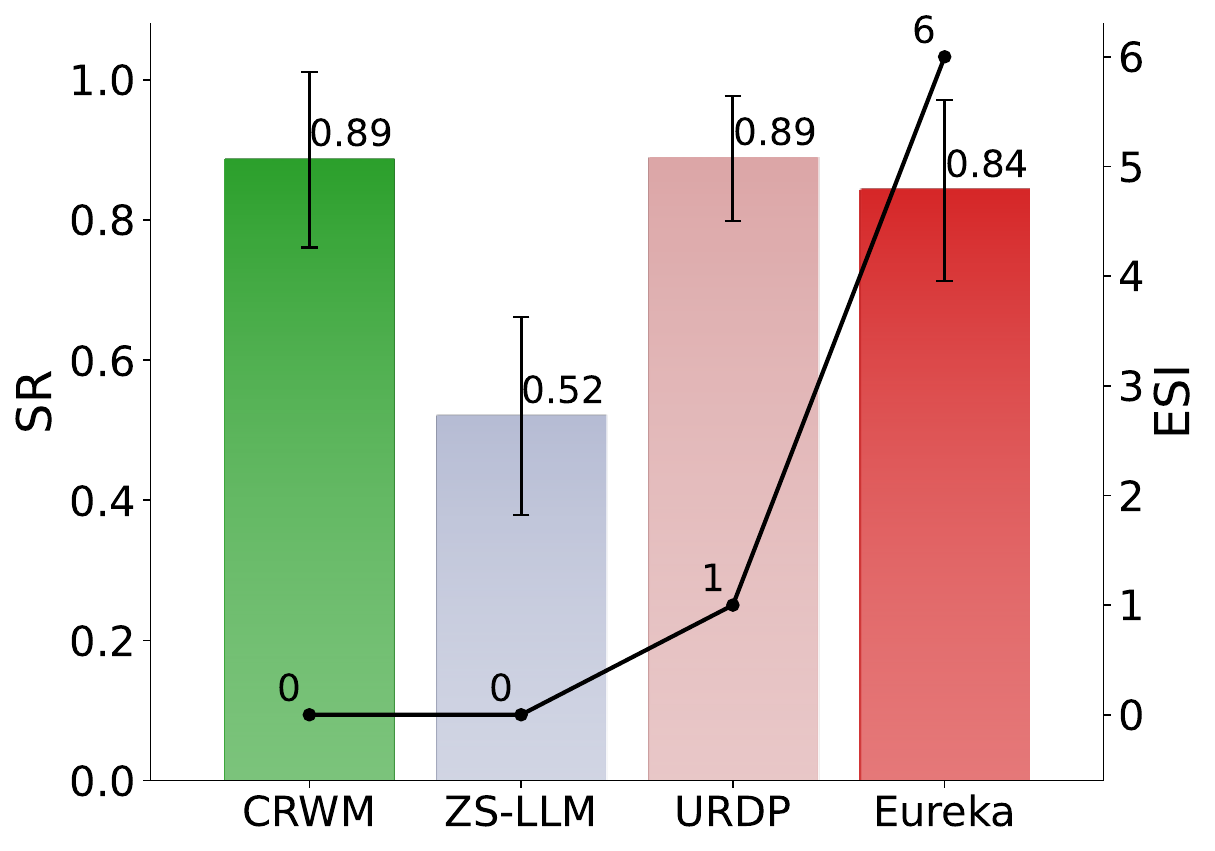}
}\hfill
\subfloat[TurnFaucet]{
    \includegraphics[width=0.318\textwidth]{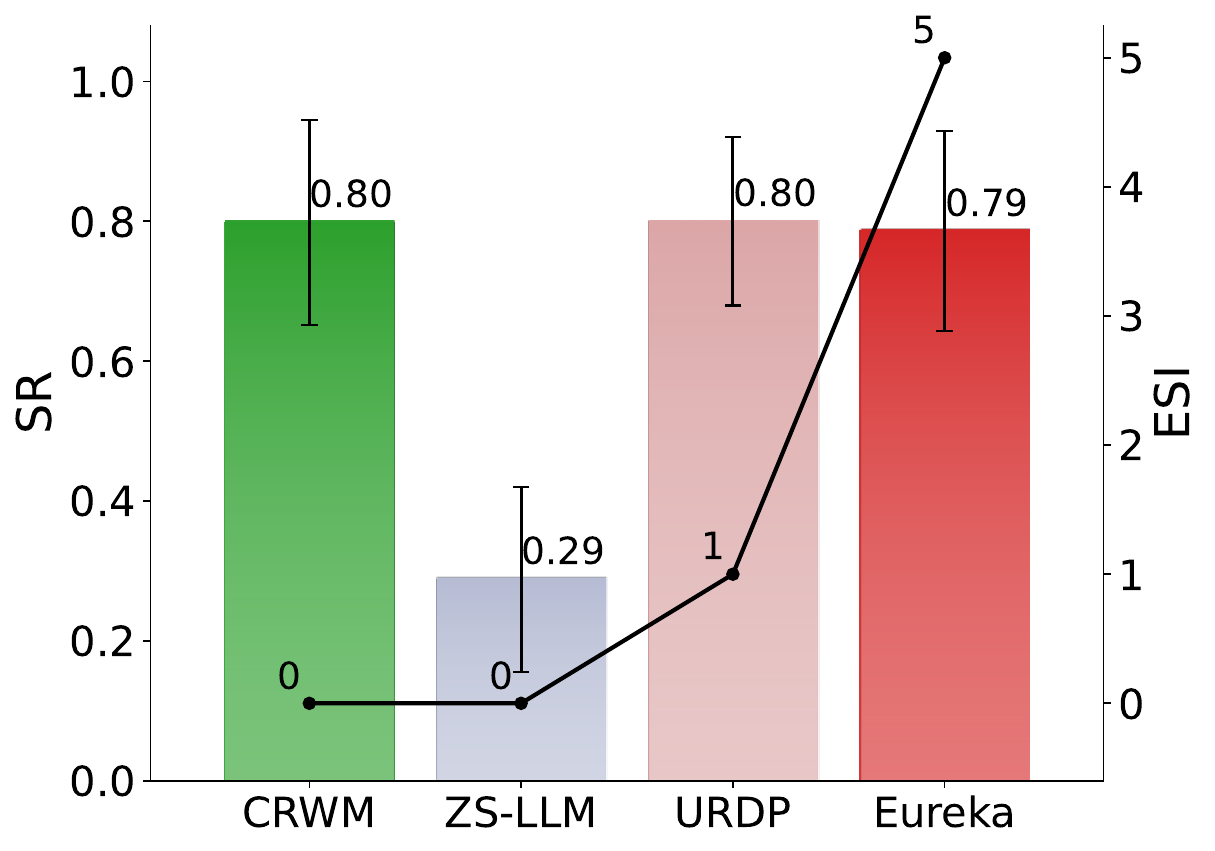}
}\hfill
\subfloat[OpenCabinetDrawer]{
    \includegraphics[width=0.318\textwidth]{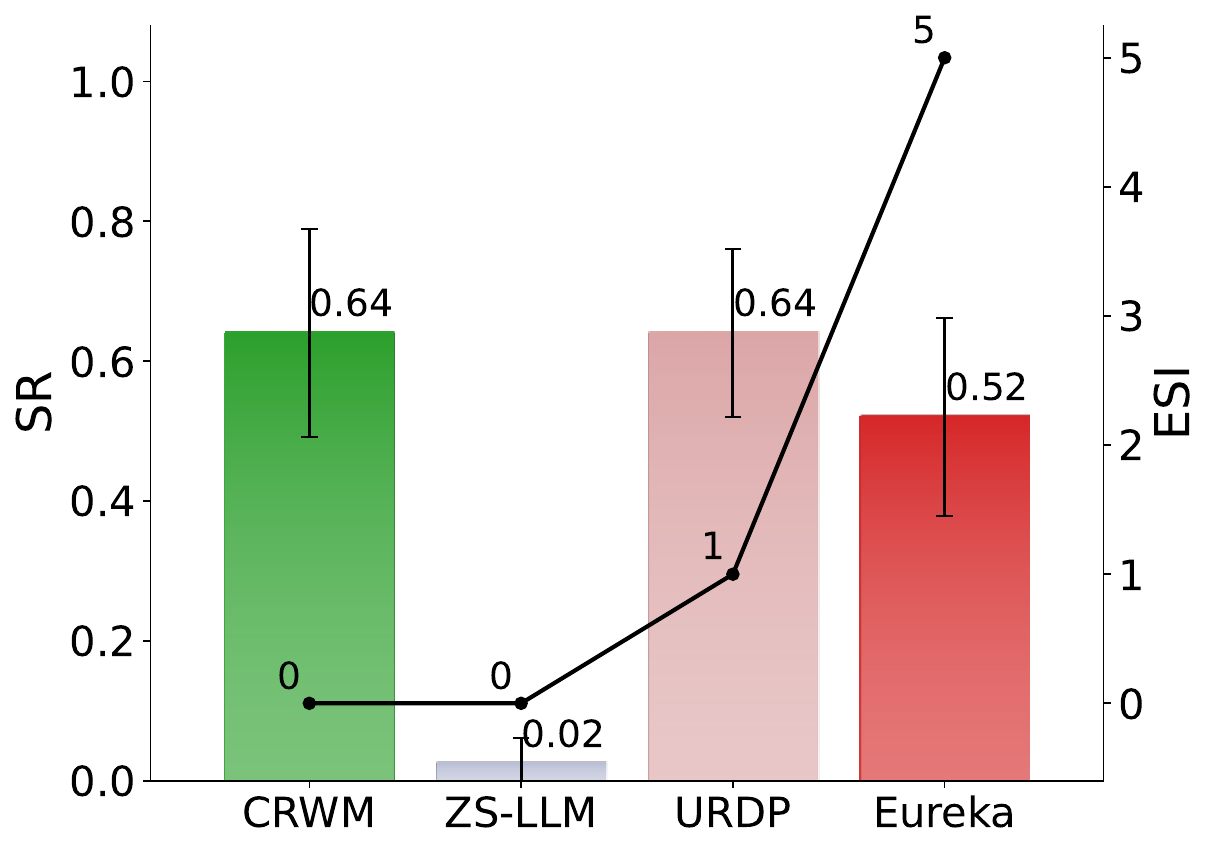}
}
\caption{\textbf{Cross-domain generalization in ManiSkill2.} 
CRWM is applied to three unseen tasks with different robot embodiments and physics environments. 
It achieves strong success rates in a zero-shot setting, without requiring iterative environment interaction.}
\label{fig:results_generalization}
\end{figure*}

The CRWM is applied to these new environments without interaction data or task-specific adaptation.
As shown in Fig.~\ref{fig:results_generalization}, it achieves strong zero-shot performance ($\text{ESI}=0$) across all tasks, with success rates of 0.64 on PickCube, 0.89 on OpenCabinetDoor, and 0.80 on TurnFaucet.
In comparison, iterative methods such as Eureka require multiple rounds of environment interaction, up to 6 ESIs, to reach similar performance and still remain less consistent.

These results indicate that the CRWM captures transferable causal structure across tasks and robot embodiments, enabling effective reward construction under substantial domain shifts without iterative interaction.

\subsection{Sim-to-Real Experiments}
\label{subsec:real_world}

We conduct real-world experiments to evaluate whether reward functions guided by the CRWM remain effective on physical hardware under the zero-shot setting.
This experiment focuses on reward alignment rather than downstream policy optimization.
We use the CASBOT W1 mobile manipulator~\cite{casbot2024} and adopt \textit{Offline Trajectory Evaluation}~\cite{gleave2021quantifying} to avoid dynamics gaps and hardware risks from direct sim-to-real deployment.
Human-teleoperated kinematic trajectories are fed into the generated reward functions to assess alignment independently of RL optimization and sim-to-real discrepancies.

\begin{figure}[htbp]
\centering
\includegraphics[width=0.85\linewidth]{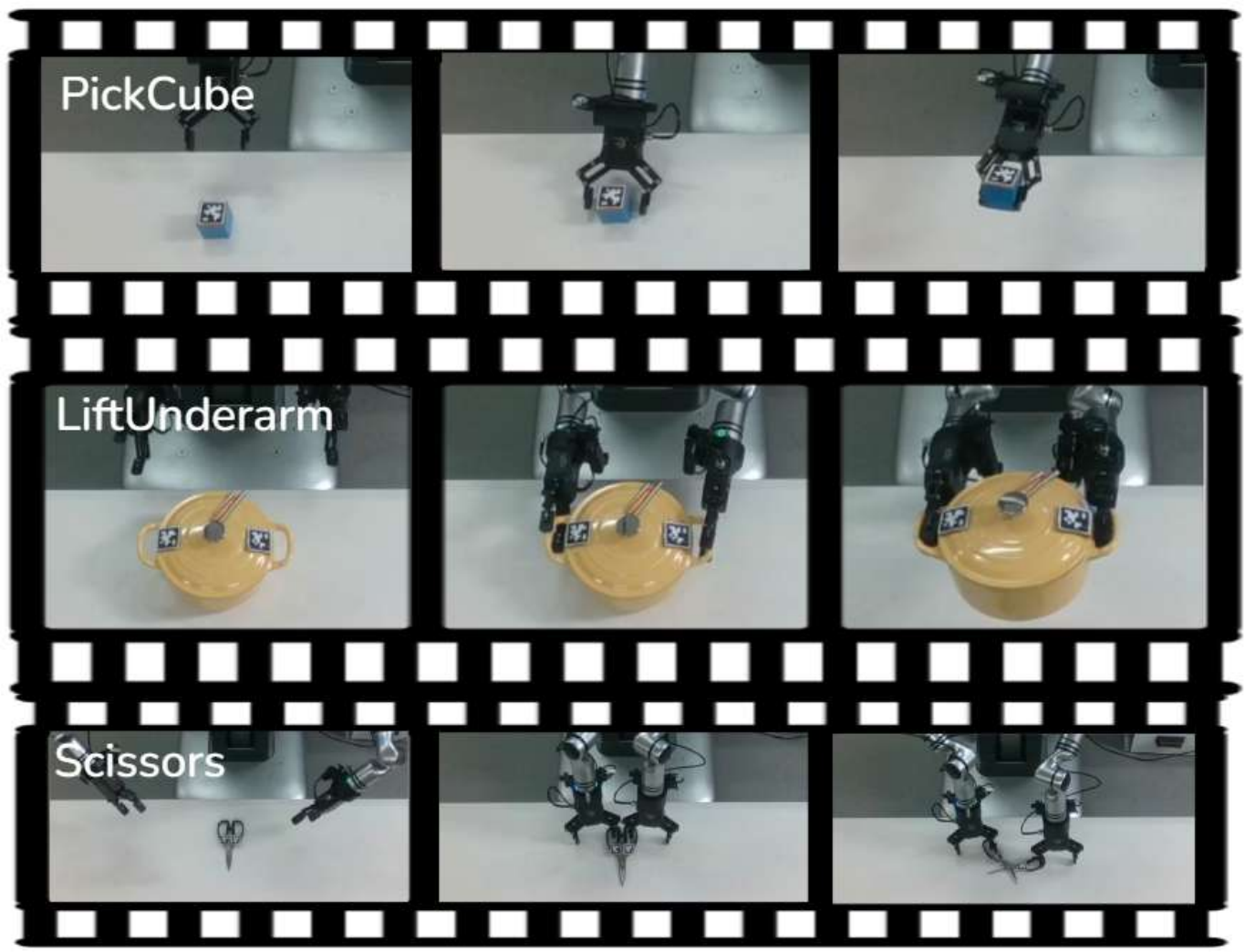}
\caption{\textbf{Real-world manipulation tasks.} PickCube: picking up a black cube. LiftUnderarm: lifting a pot. Scissors: open scissors.}
\label{fig:real_world_tasks}
\end{figure}

To quantify zero-shot alignment, we collect a balanced human-teleoperated trajectory dataset across three tasks: PickCube, LiftUnderarm, and Scissors (Fig.~\ref{fig:real_world_tasks}).
Each task contains 25 successful and 25 failed trajectories, yielding 625 successful-failed trajectory pairs for comparison.
Since reward functions differ in scale, their absolute values are not directly comparable.
We evaluate alignment using \textit{Pairwise Distinguishability Accuracy}, equivalent to the Area Under the ROC Curve (AUC)~\cite{fawcett2006introduction}.
For each successful-failed pair, a well-aligned reward function should assign a higher time-normalized mean reward $\bar{R}(\tau)$ to the successful trajectory.

As shown in Table~\ref{tab:auc_results}, the ZS-LLM baseline achieves limited distinguishability (average AUC $\approx$ 53.9\%). 
In contrast, reward functions generated under the CRWM guidance achieve high alignment accuracy across all tasks (average AUC $\approx$ 96.5\%). 
These results show that CRWM supports zero-shot reward generation that remains valid on physical hardware.

\begin{table}[htbp]
\centering
\caption{Zero-shot reward alignment accuracy (AUC) on real-world trajectories.}
\label{tab:auc_results}
\begin{tabular}{@{}lccc@{}}
\toprule
\textbf{Method} & \textbf{PickCube} & \textbf{LiftUnderarm} & \textbf{Scissors} \\ \midrule
ZS-LLM (AUC \%$\uparrow$) & 62.5\% & 47.9\% & 51.3\% \\
\textbf{CRWM} (AUC \%$\uparrow$) & \textbf{99.6\%} & \textbf{95.3\%} & \textbf{94.5\%} \\ \bottomrule
\end{tabular}
\end{table}

%% file: TNNLS/sec/discussion/discuss.tex
\section{Discussion}
\label{sec:discuss}

Fig.~\ref{fig:three_failures} in Section~\ref{sec:introduction} illustrates two common issues observed in our initial study: Optimization Collapse and Specification Gaming. To examine whether these issues are isolated cases or broader ARD limitations, we conduct a quantitative analysis.

Table~\ref{tab:failure_stats} reports the frequency of these issues across reward functions generated by baselines (Eureka and URDP) over 20 dexterous manipulation tasks. The results show that reward functions without explicit causal structure frequently exhibit these patterns:

\textbf{Optimization Collapse:} A considerable portion of baseline rewards ($21.2\%$ for Eureka and $12.6\%$ for URDP) fail to provide useful learning signals, causing both cumulative reward and SR to remain near zero during training.
The red curves in the blue boxes of Fig.~\ref{fig:hacking} illustrate this behavior, with additional examples in Supplementary Material, Fig. S-1.

\textbf{Specification Gaming:} A non-negligible fraction of baseline rewards ($10.4\%$ for Eureka and $4.7\%$ for URDP) are affected by spurious correlations, leading agents to exploit misaligned rewards and achieve high reward values while failing the task.
The red curves in the orange boxes of Fig.~\ref{fig:hacking} show this behavior, with more examples in Supplementary Material, Fig. S-1.
In contrast, CRWM-guided rewards show consistent alignment between reward signals and task outcomes, as indicated by the green curves in Fig.~\ref{fig:hacking} and Supplementary Material, Fig. S-1, suggesting that causal structure improves reward alignment in LLM-based ARD.

\begin{table}[ht]
\centering
\caption{\textbf{Statistical prevalence of reward failure modes.} The table reports the percentage of generated reward functions that fall into Optimization Collapse and Specification Gaming across all 20 dexterity tasks for the baseline ARD paradigms (Eureka and URDP).}
\label{tab:failure_stats}
\begin{tabular}{lcc}
\toprule
\textbf{Method} & \textbf{Optimization Collapse (\%) $\downarrow$} & \textbf{Specification Gaming (\%) $\downarrow$} \\
\midrule
Eureka & 21.2\% & 10.4\% \\
URDP   & 12.6\% & 4.7\% \\
\bottomrule
\end{tabular}
\end{table}

\begin{figure}
    \centering
    \includegraphics[width=0.98\linewidth]{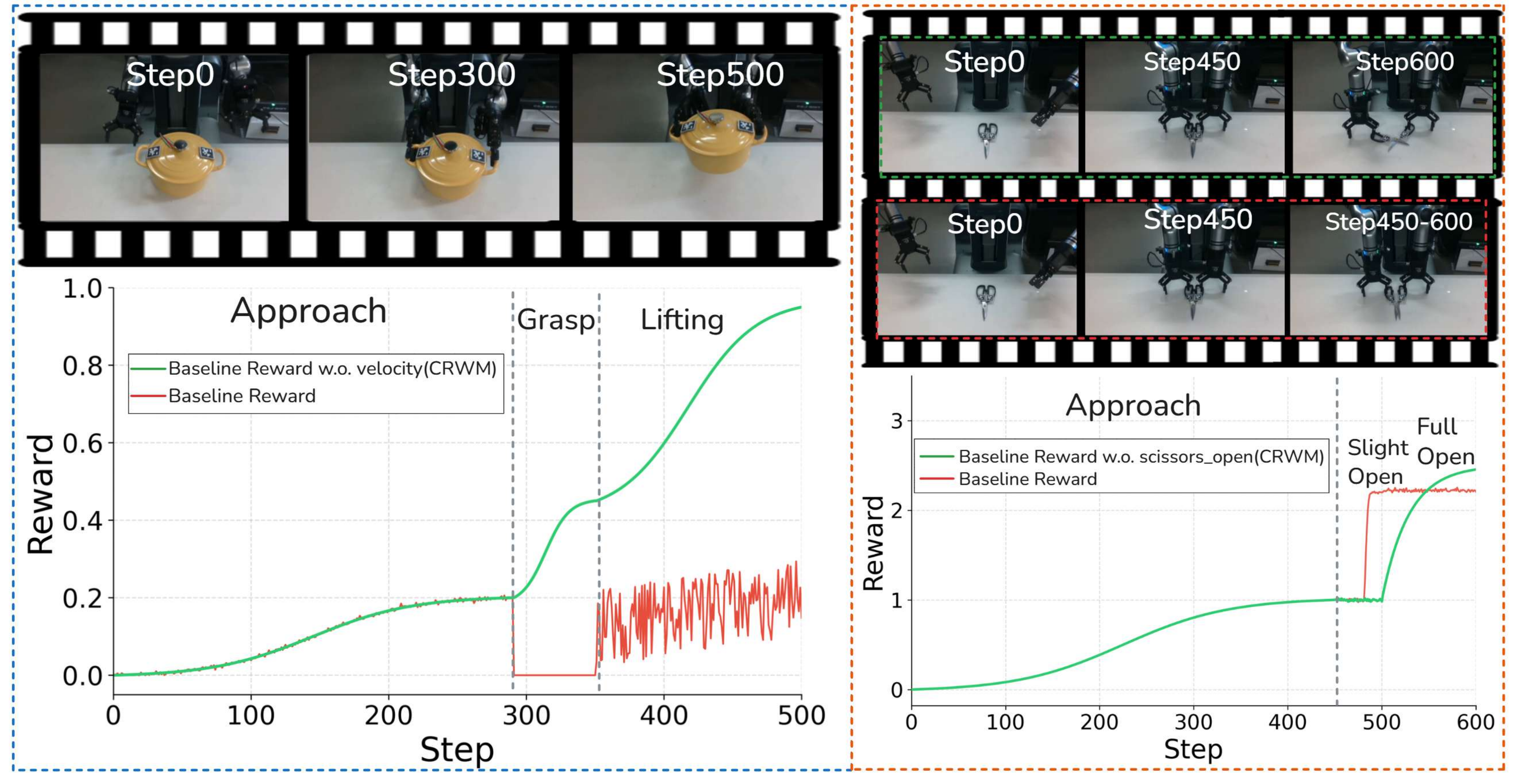}
\caption{\textbf{Real-world examples of Optimization Collapse and Specification Gaming.} 
Top (blue box): Optimization Collapse, where the baseline reward (red curve) fails to provide useful learning signals, while the CRWM (green curve) continues to reflect task progress. 
Bottom (orange box): Specification Gaming, where the baseline reward saturates prematurely despite incomplete task execution, whereas the CRWM remains aligned with the task outcome.}
    \label{fig:hacking}
\end{figure}

%% file: TNNLS/sec/appendix/appen.tex
\clearpage
\section{Visualization of the CRWM}
\label{app:crwm_visualization}

\begin{figure*}[t]
    \centering
    \includegraphics[width=0.9\textwidth]{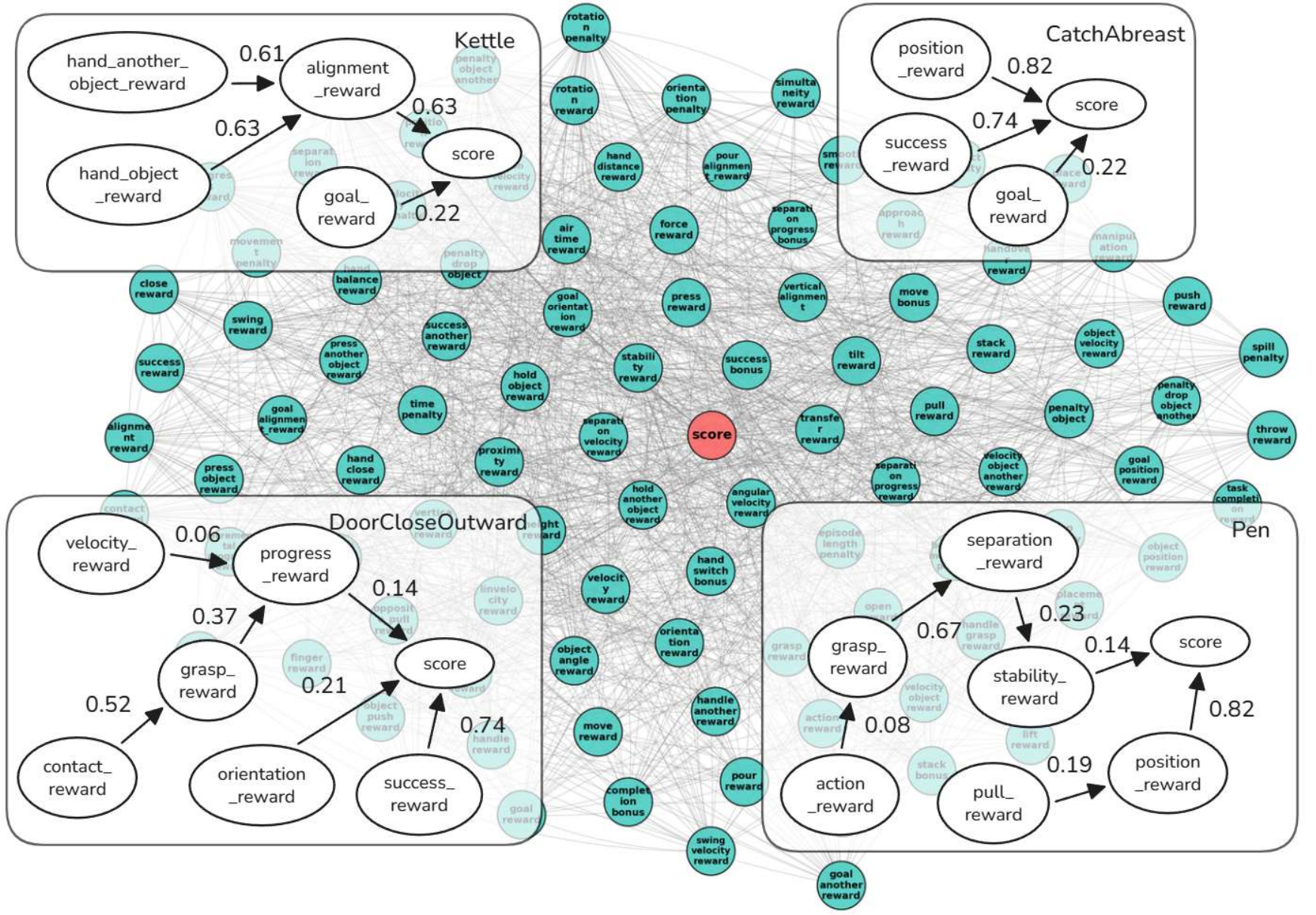} 
\caption{\textbf{Visualization of the CRWM.} 
The background graph shows the task-invariant causal skeleton, where the red node denotes the \texttt{score}. 
The four insets present subgraphs corresponding to different tasks, illustrating the relevant structures within the CRWM. 
Directed edges indicate causal relationships, and edge weights represent the corresponding causal effects.}
    \label{fig:global_crwm}
\end{figure*}

To illustrate the learned causal structure, we visualize the CRWM in Fig.~\ref{fig:global_crwm}. 
The background graph is the CRWM distilled from multi-task data, where the central red node denotes the fitness score. 
The four insets show subgraphs corresponding to the \textit{Kettle}, \textit{CatchAbreast}, \textit{DoorCloseOutward}, and \textit{Pen} tasks. 
These task-specific subgraphs are not learned as isolated graphs; instead, they are selected from the same CRWM according to the reward components relevant to each task. 
This visualization shows how different task appearances can share a common reward-relevant topology while using different subsets of the global component pool.

\section{Implementation Details}
\label{app:implementation_details}

\subsection{Causal-Guided System Prompt}
\label{app:prompt}

The following prompt is provided to the LLM as a system-level directive. It forces the model to generate Python code based on the CRWM, effectively bypassing spurious structural correlations.

\begin{tcolorbox}[
    enhanced,
    breakable,
    width=\linewidth,
    colback=gray!5, 
    colframe=black!75, 
    sharp corners,  
    boxrule=0.8pt, 
    left=6pt, right=6pt, top=8pt, bottom=8pt,
    fonttitle=\bfseries\small,
    coltitle=white, 
    colbacktitle=black!75, 
    title={System Prompt: CRWM-Guided Reward Synthesis}
]
\small
\raggedright
\noindent The following causal graph (format: \texttt{[source, target, weight]}) explicitly defines the physical dependencies between primitive reward components and the \texttt{score}.

\vspace{0.2cm}
\noindent \textbf{CRITICAL DIRECTIVE 1: Causal Pruning} \\
You MUST ONLY include primitive components that possess a valid causal path to the \texttt{score} node. Any variable or sub-path (e.g., $A \rightarrow B$) that fails to reach \texttt{score} is a spurious correlation. You \textbf{MUST exclude} these by setting their coefficients to \texttt{0.0} in the Python code.

\vspace{0.2cm}
\noindent \textbf{CRITICAL DIRECTIVE 2: Task-Specific Selection \& Adaptation} \\
Among all valid components that causally influence the \texttt{score}, you must proactively select the optimal combination of reward terms that best aligns with the specific objective of the target task. Furthermore, you can smoothly adapt the abstract component names from the causal graph to match the exact concrete variable names defined in the current environment's observation space.

\vspace{0.2cm}
\noindent \textbf{Graph Interpretation Rules}:
\begin{itemize}
    \setlength{\itemsep}{2pt}
    \item \textbf{Edges (\texttt{source} $\to$ \texttt{target})}: Defines a prerequisite relationship where the \texttt{source} physically drives the \texttt{target}.
    \item \textbf{Causal Weights ($0.0 - 1.0$)}: Quantifies structural importance. Use these as inductive priors to scale your reward coefficients.
    \item \textbf{Variables}: Comprises raw state observations and primitive reward terms.
\end{itemize}
\end{tcolorbox}

\subsection{Multi-Task Implementation Details}
\label{app:task_details}
This appendix provides implementation details for constructing the shared structural variable space used by the CRWM and performing joint optimization across pre-training tasks.

\textbf{Task-Specific Name Normalization.}
Different tasks may use task-specific names for semantically equivalent physical quantities, such as \texttt{scissors\_position}, \texttt{pot\_position}, or \texttt{bottle\_position}.
To support cross-task causal structure learning, we normalize these names into a shared set of atomic reward components.
For example, object-specific position terms are mapped to a unified component, such as \texttt{object\_position}.
This reduces redundant task-specific naming while preserving each variable's physical meaning.

\begin{figure}
    \centering
    \includegraphics[width=0.95\linewidth]{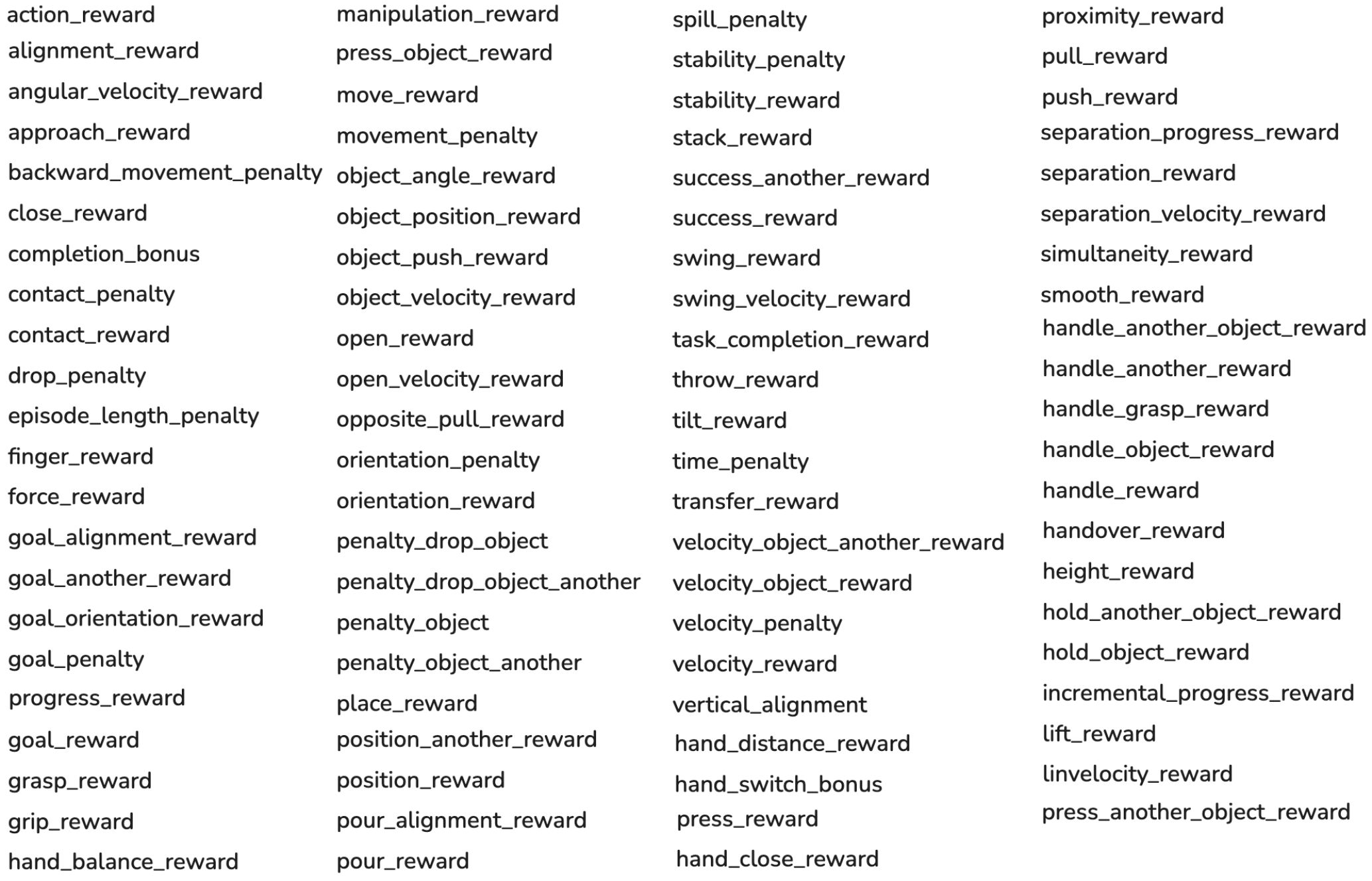}
    \caption{The global pool of normalized atomic reward components used to construct the shared structural variable space.}
    \label{fig:reward_cp}
\end{figure}

\textbf{Global Atomic Reward Component Pool.}
\textcolor{black}{After normalization, all atomic reward components are organized into a global component pool (Fig.~\ref{fig:reward_cp}), covering common reward-relevant factors in manipulation tasks, including spatial relations, contact, grasping, stability, velocity, orientation, and task progress.
Each pre-training task uses only a subset of this pool, and its task-specific state vector is aligned into the same $d$-dimensional structural variable space $s_t \in \mathbb{R}^d$ with unused variables zero-padded.
This alignment enables heterogeneous task trajectories to be combined into the unified micro-level dataset $\mathcal{D}_{micro}$.
When a target task requires reward components outside the current pool, the corresponding physical variables must be added to the pool and supported by additional task data before the CRWM can reason over them.}

\begin{figure}
    \centering
    \includegraphics[width=0.95\linewidth]{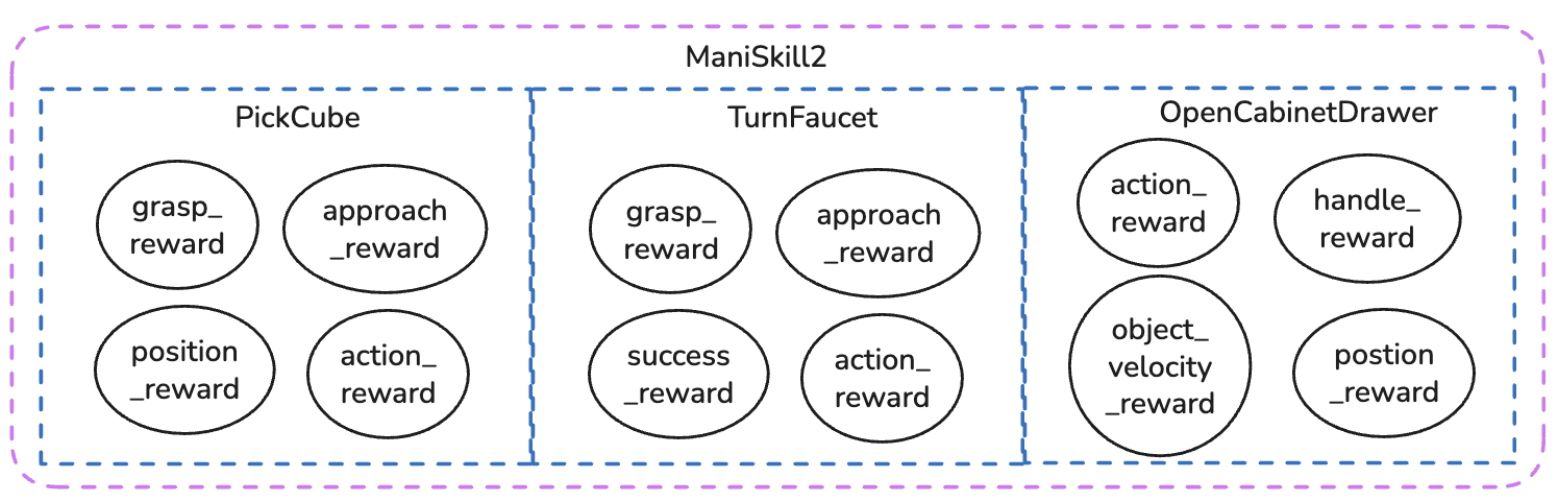}
    \caption{Reward component combinations used for the three ManiSkill2 generalization tasks. These components are instantiated from the global atomic reward component pool.}
    \label{fig:manisill2_rw}
\end{figure}

\textbf{Compositional Coverage.}
The global component pool is not intended to cover all possible physical variables, but provides a reusable set of kinematic and dynamic variables commonly used in robotic manipulation.
The CRWM generalizes to new tasks through new causal combinations of these existing structural variables, not by enumerating every possible variable.
In the ManiSkill2 generalization experiments, reward components for unseen tasks are selected from this pool, as shown in Fig.~\ref{fig:manisill2_rw}.

\textbf{Joint Distillation and Cross-Task Voting.}
During Joint Optimization, pre-training tasks are optimized jointly rather than sequentially.
At each training step, batches are sampled from the aggregated multi-task dataset, supporting the Confidence-Aware Soft Fusion module.
If a structural hallucination originates from task-specific naming or semantic bias, it is unlikely to be consistently supported by physical transitions across tasks.
Thus, reconstruction gradients accumulated across tasks can weaken such spurious edges in $M_{inv}$, while recurring physical relationships are reinforced in the shared structural variable space.

\subsection{CRWM Hyper-parameter Settings}
\label{app:hyperparameters}

In this section, we detail the three hyper-parameters governing the distillation process of the CRWM that warrant discussion, as shown in Table~\ref{tab:hyperparameters}.

\begin{table}[ht]
\centering
\caption{Hyperparameters of \textit{CRWM}.}
\label{tab:hyperparameters}
\begin{tabular}{lc}
\toprule
\textbf{Hyper-parameter} & \textbf{Value} \\
\midrule
Macro-level Sampling Size $N$ & 30 \\
Acyclicity Penalty Weight $\lambda$ & 0.1 \\
Soft Fusion Weight $\gamma$ & 1.0 \\
\bottomrule
\end{tabular}
\end{table}

\begin{figure}[htbp]
\centering
\includegraphics[width=0.95\linewidth]{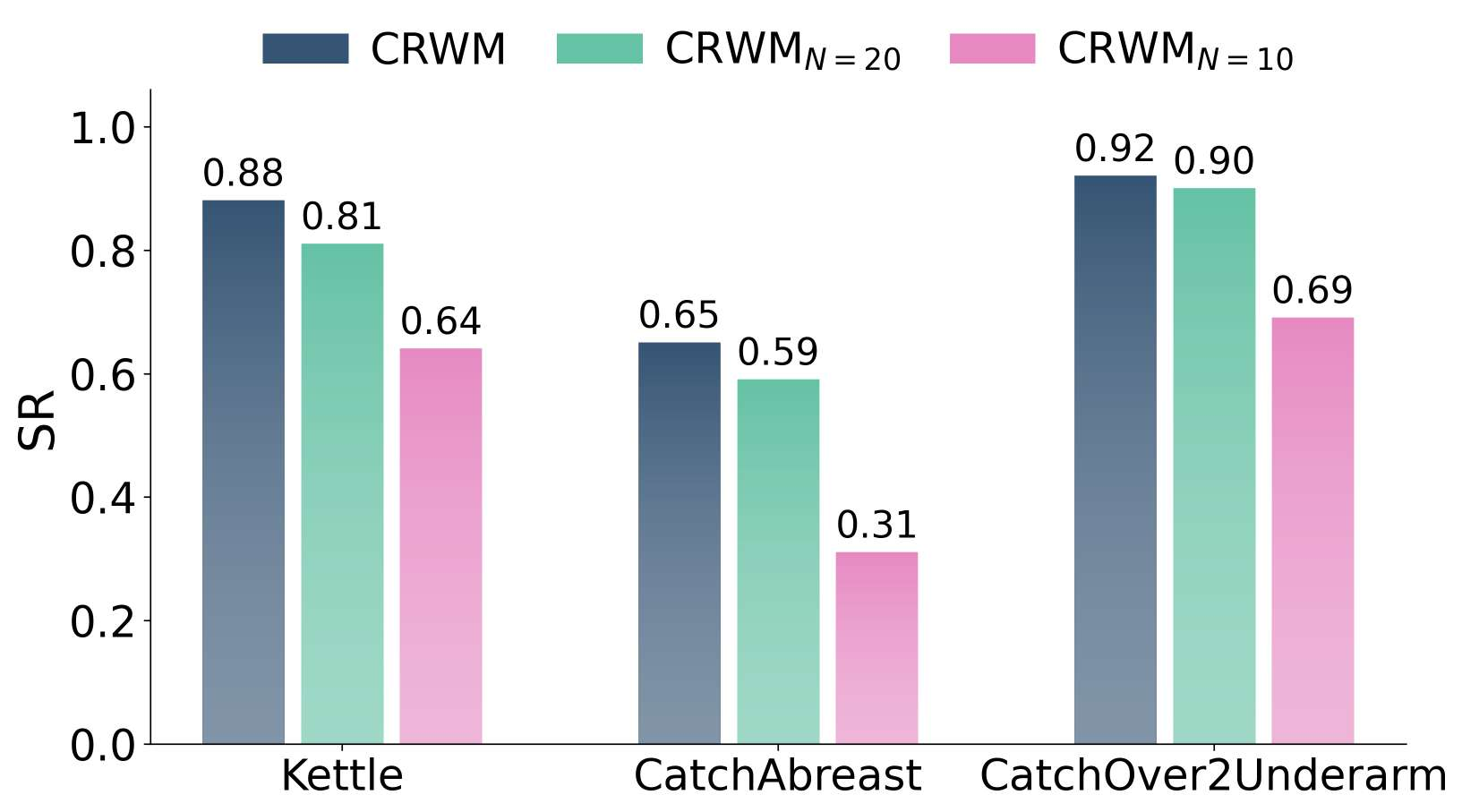}
\caption{\textbf{Sensitivity analysis of macro-level intervention sampling size $N$.} 
Reducing $N$ decreases the amount of offline interventional data and leads to lower SR, indicating that sufficient macro-level interventions and micro-level trajectories are important for CRWM.}
\label{fig:N_compare}
\end{figure}

\textbf{Macro-level Intervention Sampling Size ($N$):} 
This parameter controls the amount of offline interventional data collected before joint optimization.
By default, we set $N = 30$ for all tasks and reduce it to $20$ and $10$ to study its impact.
As shown in Fig.~\ref{fig:N_compare}, the Success Rate (SR) decreases as $N$ becomes smaller.
This trend reflects reduced data support during training: smaller $N$ limits both the macroscopic dataset ($\mathcal{D}_{macro}$) and microscopic trajectories ($\mathcal{D}_{micro}$).
With fewer samples, the structural prior becomes less reliable and joint optimization receives weaker physical signals, leading to less accurate reward alignment.

\textbf{DAGMA Acyclicity Penalty Weight ($\lambda$):} 
This parameter controls the strength of the Directed Acyclic Graph (DAG) constraint.
We solve the constrained optimization using the Augmented Lagrangian Method (ALM), initializing $\lambda$ to $0.1$.
During optimization, $\lambda$ is updated in the ALM outer loop to enforce the acyclicity constraint $h(M_{inv}) = 0$.
Since $\lambda$ is adjusted during training, we do not treat it as a fixed hyper-parameter or ablate predefined values.
This scheme enforces the DAG constraint without manual tuning.

\begin{figure}[htbp]
\centering
\includegraphics[width=0.95\linewidth]{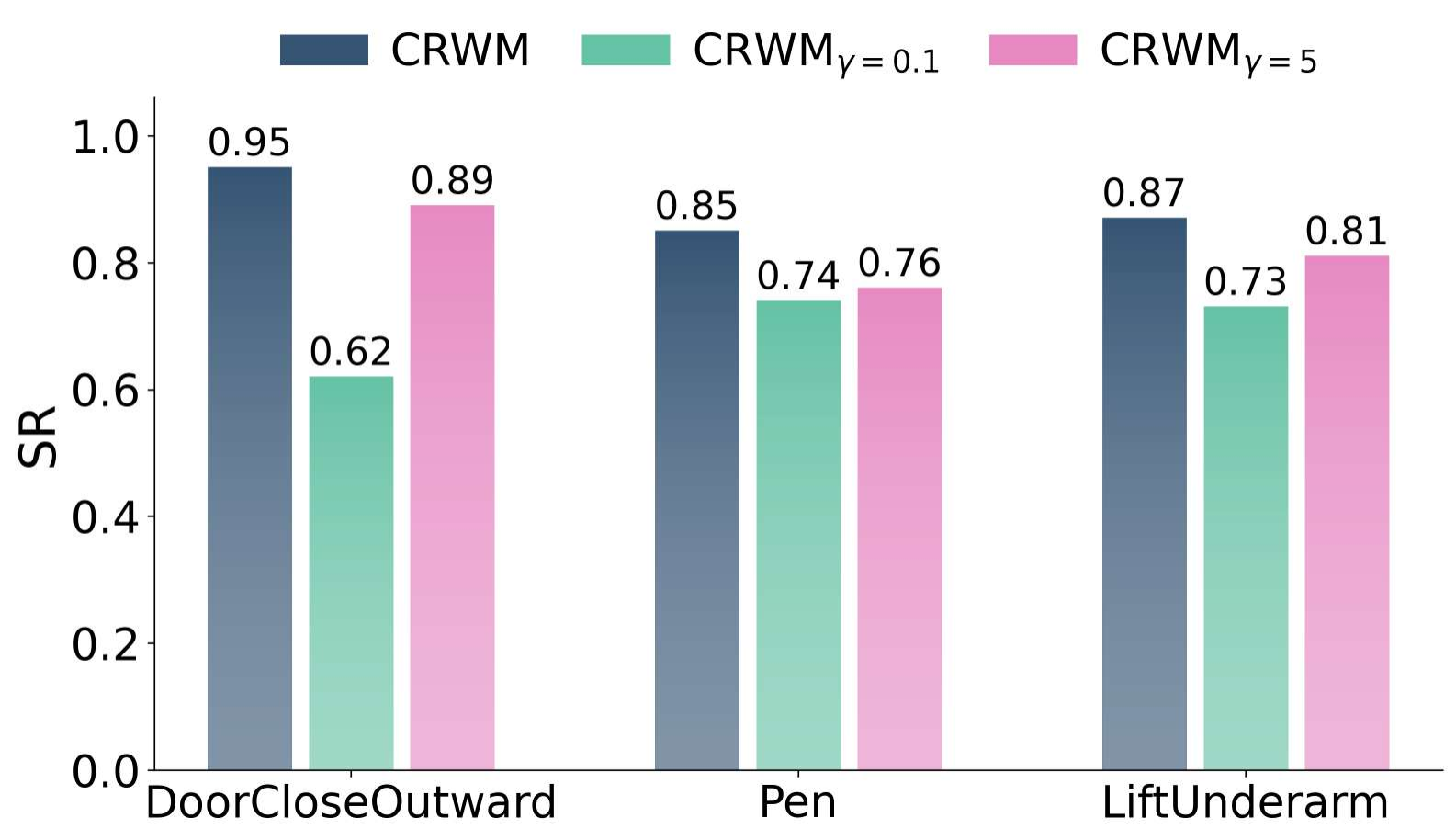}
\caption{\textbf{Sensitivity analysis of the Soft Fusion weight $\gamma$.} 
The default setting $\gamma=1.0$ achieves the best balance between the structural prior and physical transition data, while overly small or large values reduce the SR.}
\label{fig:gama_compare}
\end{figure}

\textbf{Confidence-Aware Soft Fusion Weight ($\gamma$):} 
This weight controls the balance between the structural prior ($\mathcal{L}_{soft}$) and the reconstruction term ($\mathcal{L}_{MSE}$). 
We set $\gamma = 1.0$ by default and evaluate $\gamma \in \{0.1, 1.0, 5.0\}$. 
As shown in Fig.~\ref{fig:gama_compare}, $\gamma = 1.0$ yields the best performance.
Reducing $\gamma$ to $0.1$ weakens the influence of the structural prior, leading to lower SR due to limited data support. 
Increasing $\gamma$ to $5.0$ over-constrains the model with the prior, reducing the ability of physical data to correct structural hallucinations.

\subsection{\textcolor{black}{Offline Pre-training Cost}}
\label{app:compute_cost}
The CRWM requires a one-time offline pre-training stage before zero-shot deployment.
In the Dexterity zero-shot evaluation, this stage uses only the 14 pre-training tasks, while the 6 held-out tasks are reserved for evaluation.
As summarized in Table~\ref{tab:compute_cost}, the ARD-based component construction process produces approximately 140 reward-function specifications; with $N=30$ reward-weight configurations per reward function, this yields 4200 policy-training runs for constructing $\mathcal{D}_{macro}$ and $\mathcal{D}_{micro}$.

This offline cost is incurred once during CRWM construction.
Afterward, reward generation for held-out tasks requires no evolutionary search or task-specific environment feedback, resulting in $\mathrm{ESI}=0$ during zero-shot deployment.
Thus, ESI in the main experiments measures deployment-time search cost, while the offline pre-training cost is reported separately here.

\begin{table}[htbp]
\centering
\caption{Offline pre-training cost for CRWM construction.}
\label{tab:compute_cost}
\resizebox{\linewidth}{!}{
\begin{tabular}{lc}
\toprule
\textbf{Item} & \textbf{Setting} \\
\midrule
Pre-training tasks & 14 \\
Reward-function specifications & 140 \\
Reward-weight configurations per reward function & $N=30$ \\
Total policy-training runs & 4200 \\
RL policy training iterations per run & 3000 \\
Hardware for data collection & 8 NVIDIA A800 GPUs \\
Parallel policy-training runs per batch & 80 \\
Wall-clock time per collection batch & $\sim$1 hour \\
Estimated collection batches & 50 \\
Deployment-time ESI after CRWM construction & 0 \\
\bottomrule
\end{tabular}
}
\end{table}

\section{Limitations and Discussion}
\textbf{Pre-training Cost.} 
The proposed framework relies on a pre-training stage to construct the CRWM from multi-task offline data.
This stage requires interventional data collection and multiple RL policy-training runs, introducing non-trivial computational cost.
Unlike iterative ARD methods that distribute this cost across tasks through online interaction, our approach concentrates it in an offline phase before deployment.
Although the resulting CRWM can be reused for zero-shot reward generation on new tasks, the upfront cost may become significant when scaling to larger task suites or more complex environments.
Reducing the data and computation required for this pre-training stage remains an important direction for future work.

\textcolor{black}{\textbf{Bounded Generalization by the Component Pool.}
The CRWM generalizes to new tasks by recombining reward-relevant structural variables in the global atomic reward component pool.
Thus, its zero-shot generalization is bounded by the pool coverage.
If a target task requires physical primitives absent from the pool, such as fluid dynamics variables or deformable-object constraints, the CRWM cannot directly reason over them.
Extending the CRWM with open-vocabulary physical variable discovery and automatic component-pool expansion remains an important direction for future work.}

\textbf{Sim-to-Real Coverage for Dual-Hand Manipulation.} 
Although our main experiments focus on dual-hand dexterous manipulation in simulation, we do not provide corresponding dual-hand sim-to-real validation due to the lack of suitable dual-hand robotic hardware in our current setup.
Extending the evaluation to real-world dual-hand systems is an important next step for further assessing the framework's practical applicability.

%% file: sublimentary/context/word.tex
\section{Detailed Task Specifications}
\label{app:env}

\begin{center}
\hrule height 0.8pt 
\vspace{5pt}
     Dexterity Environments
\end{center}
\hrule height 0.8pt 

\vspace{10pt}

\noindent
\begin{minipage}[t]{0.78\textwidth}
    \vspace{0pt} 
    Environment (obs dim, action dim)\\
    Task description 
\end{minipage}
\vspace{4pt}
\hrule 
\vspace{5pt}

\noindent
\begin{minipage}[t]{0.76\textwidth}
    \vspace{0pt} 
    \textbf{Over (398, 40)} \\
    This class corresponds to the HandOver task. This environment consists of two shadow hands with palms facing up, opposite each other, and an object that needs to be passed. In the beginning, the object will fall randomly in the area of the shadow hand on the right side. Then the hand holds the object and passes the object to the other hand. Note that the base of the hand is fixed. More importantly, the hand which holds the object initially cannot directly touch the target, nor can it directly roll the object to the other hand, so the object must be thrown up and stays in the air in the process.
\end{minipage}
\hfill
\begin{minipage}[t]{0.20\textwidth}
    \vspace{0pt} \centering
    \includegraphics[width=\linewidth]{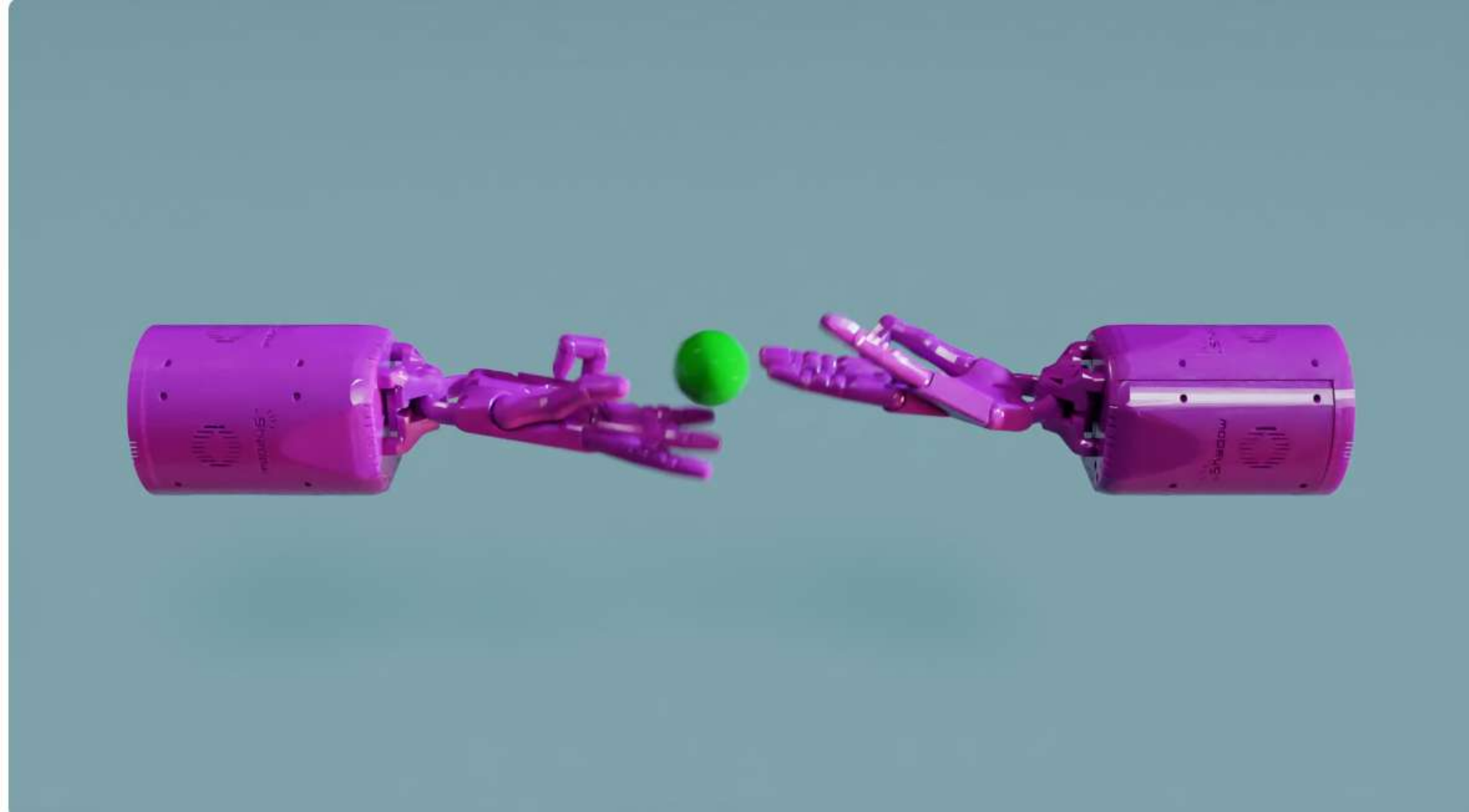} 
\end{minipage}
\vspace{8pt} \hrule \vspace{5pt}

\noindent
\begin{minipage}[t]{0.76\textwidth}
    \vspace{0pt}
    \textbf{DoorCloseInward (417, 52)} \\
    This class corresponds to the DoorCloseInward task. This environment require a closed door to be opened and the door can only be pushed outward or initially open inward. Both these two environments only need to do the push behavior, so it is relatively simple.
\end{minipage}
\hfill
\begin{minipage}[t]{0.20\textwidth}
    \vspace{0pt} \centering
    \includegraphics[width=\linewidth]{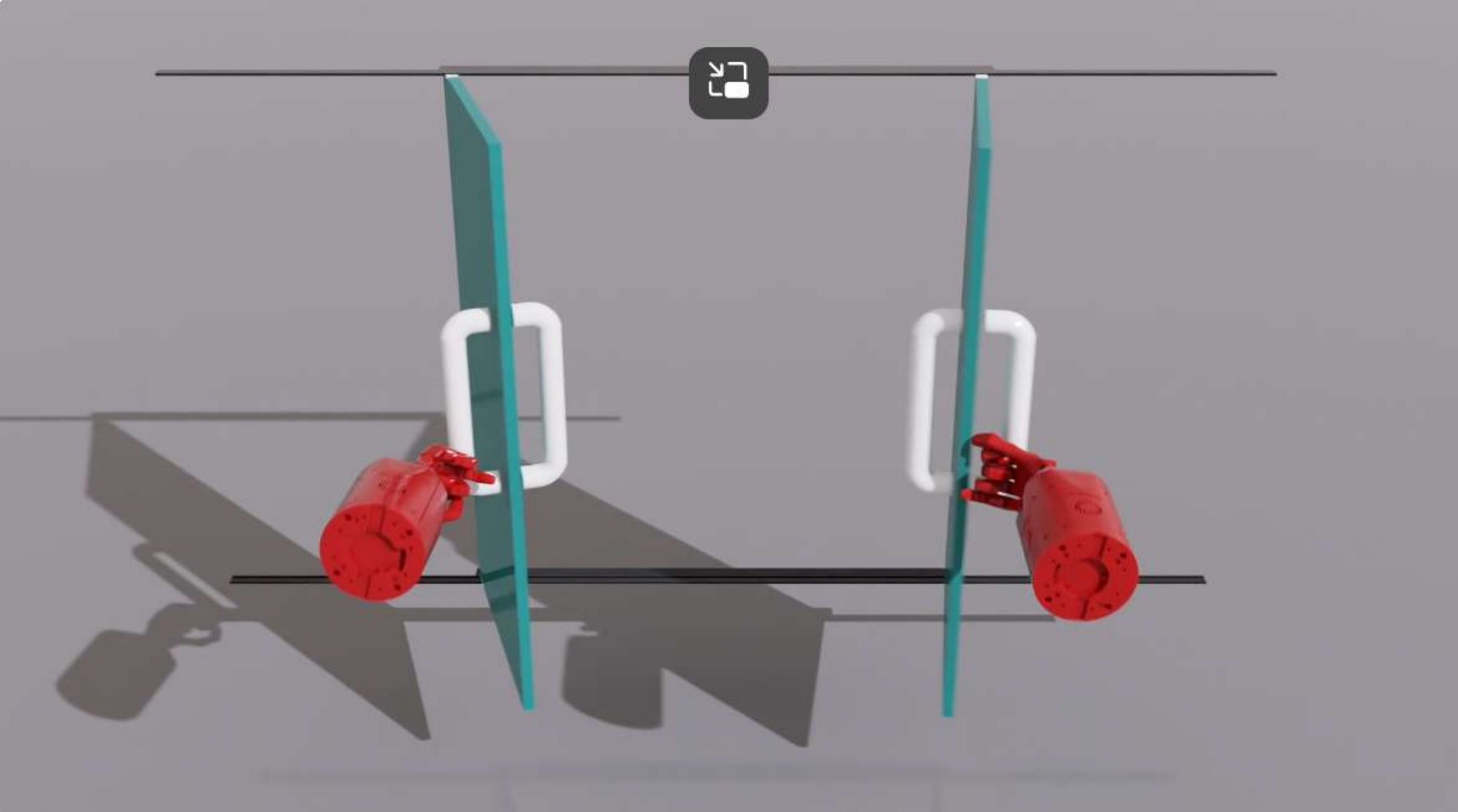}
\end{minipage}
\vspace{8pt} \hrule \vspace{5pt}

\noindent
\begin{minipage}[t]{0.76\textwidth}
    \vspace{0pt} 
    \textbf{DoorCloseOutward (417, 52)} \\
    This class corresponds to the DoorCloseOutward task. This environment also require a closed door to be opened and the door can only be pushed inward or initially open outward, but because they cannot complete the task by simply pushing, which need to catch the handle by hand and then open or close it, so it is relatively difficult.
\end{minipage}
\hfill
\begin{minipage}[t]{0.20\textwidth}
    \vspace{0pt} \centering
    \includegraphics[width=\linewidth]{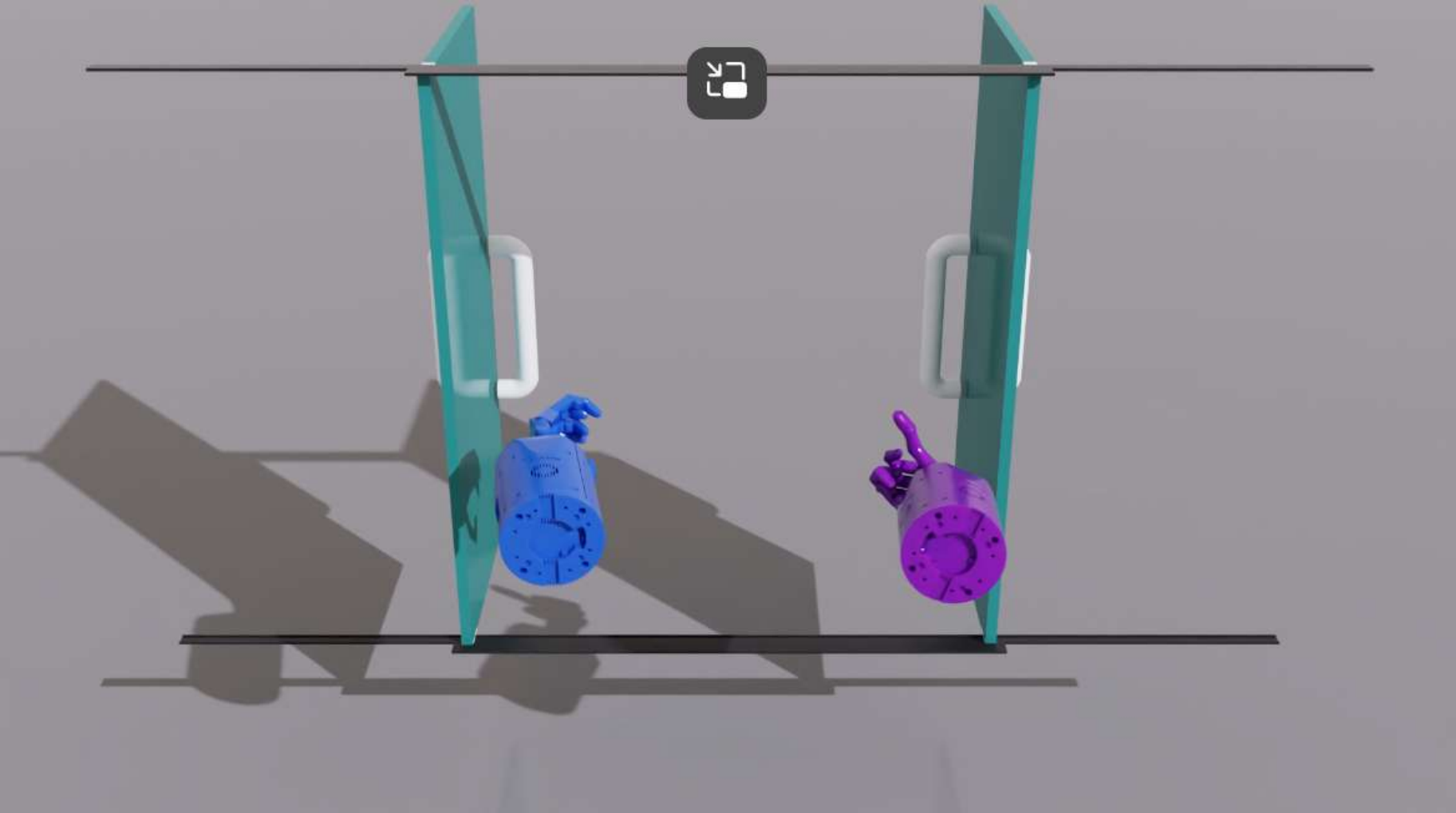} 
\end{minipage}
\vspace{8pt} \hrule \vspace{5pt}

\noindent
\begin{minipage}[t]{0.76\textwidth}
    \vspace{0pt} 
    \textbf{DoorOpenInward (417, 52)} \\
    This class corresponds to the DoorOpenInward task. This environment also require a opened door to be closed and the door can only be pushed inward or initially open outward, but because they cannot complete the task by simply pushing, which need to catch the handle by hand and then open or close it, so it is relatively difficult.
\end{minipage}
\hfill
\begin{minipage}[t]{0.20\textwidth}
    \vspace{0pt} \centering
    \includegraphics[width=\linewidth]{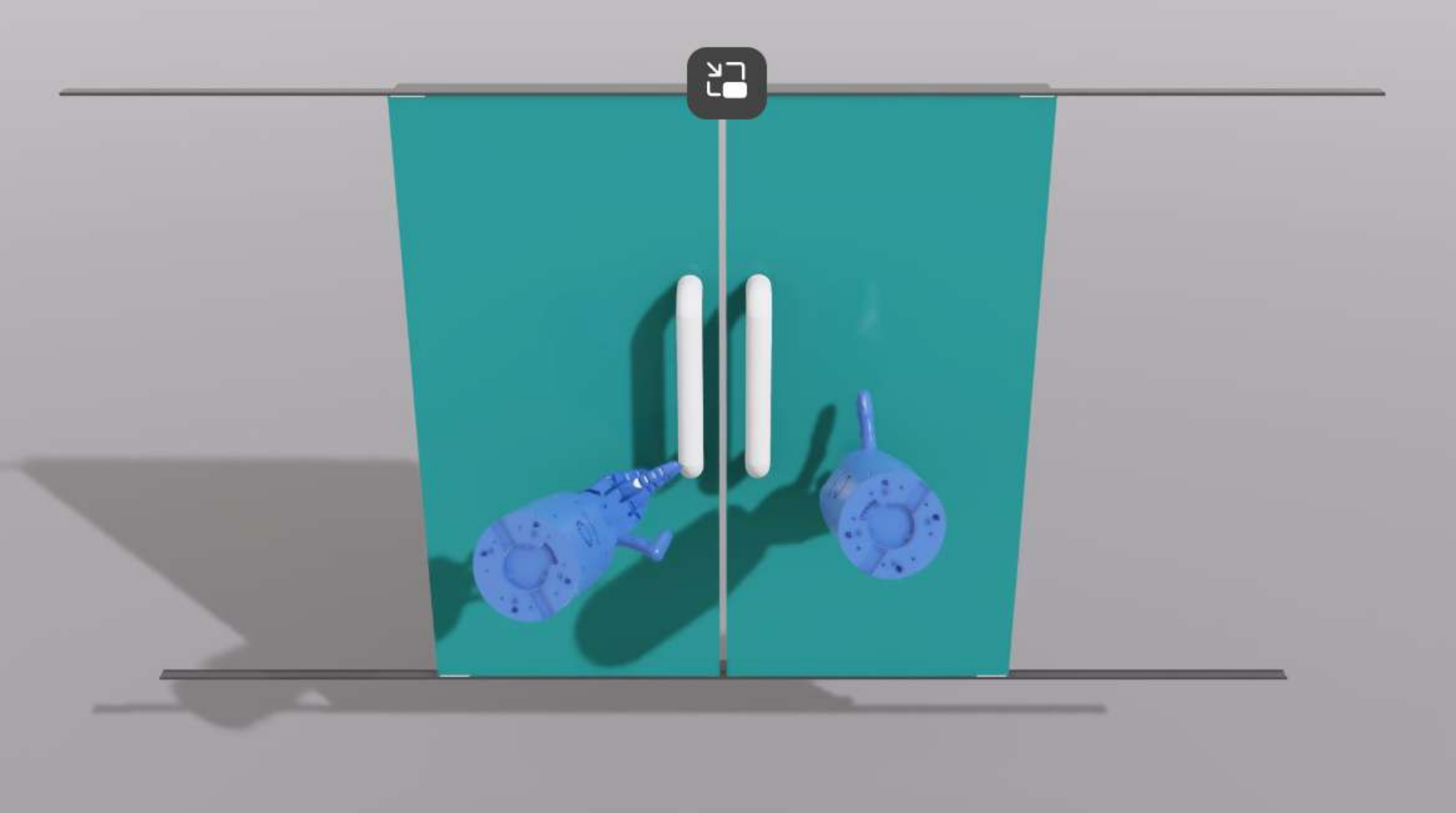} 
\end{minipage}
\vspace{8pt} \hrule \vspace{5pt}

\noindent
\begin{minipage}[t]{0.76\textwidth}
    \vspace{0pt} 
    \textbf{DoorOpenOutward (417, 52)} \\
    This class corresponds to the DoorOpenOutward task. This environment require a opened door to be closed and the door can only be pushed outward or initially open inward. Both these two environments only need to do the push behavior, so it is relatively simple.
\end{minipage}
\hfill
\begin{minipage}[t]{0.20\textwidth}
    \vspace{0pt} \centering
    \includegraphics[width=\linewidth]{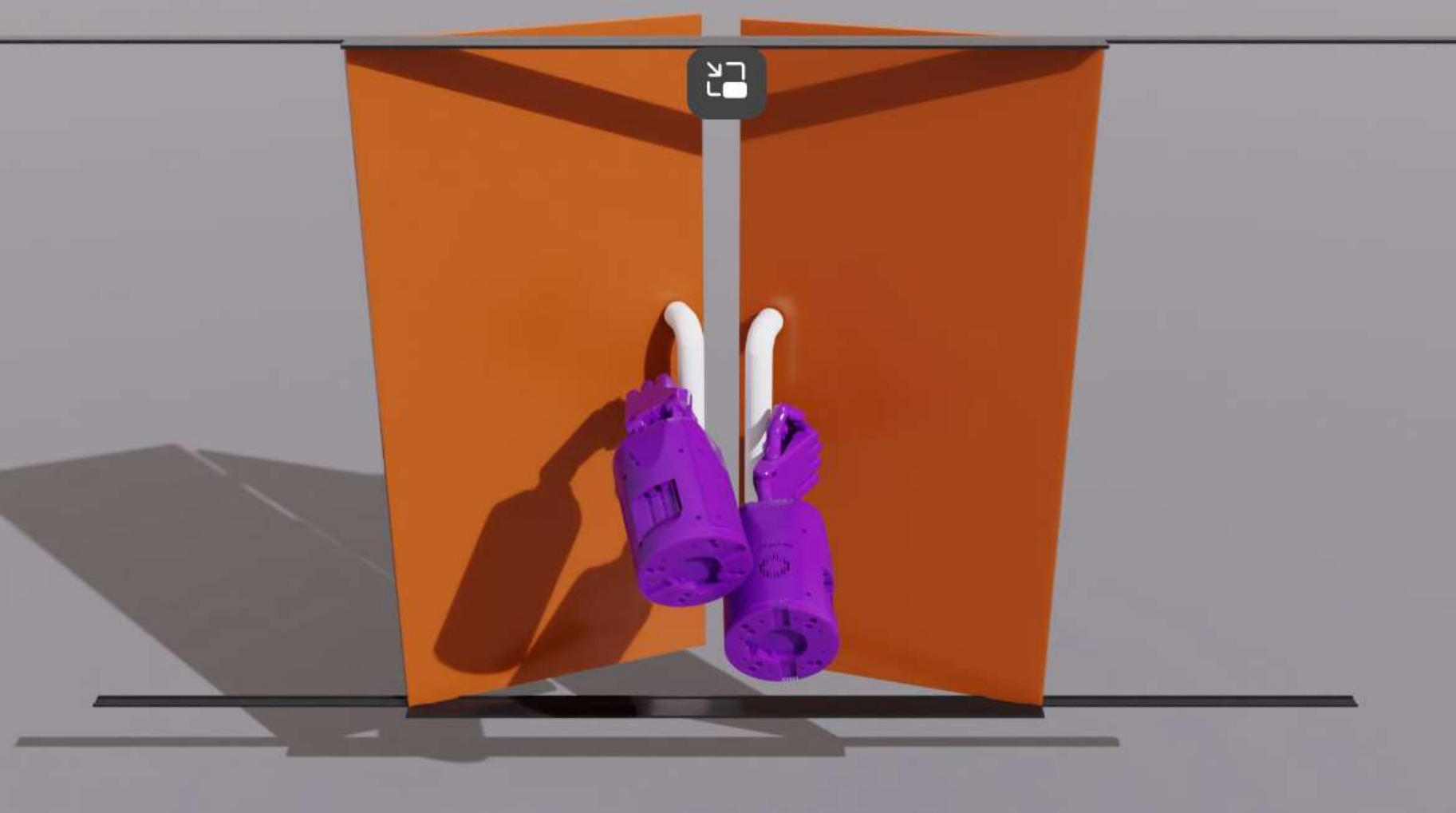} 
\end{minipage}
\vspace{8pt} \hrule \vspace{5pt}

\noindent
\begin{minipage}[t]{0.76\textwidth}
    \vspace{0pt} 
    \textbf{Scissors (417, 52)} \\
    This class corresponds to the Scissors task. This environment involves two hands and scissors, we need to use two hands to coordinate and open the scissors successfully.
\end{minipage}
\hfill
\begin{minipage}[t]{0.20\textwidth}
    \vspace{0pt} \centering
    \includegraphics[width=\linewidth]{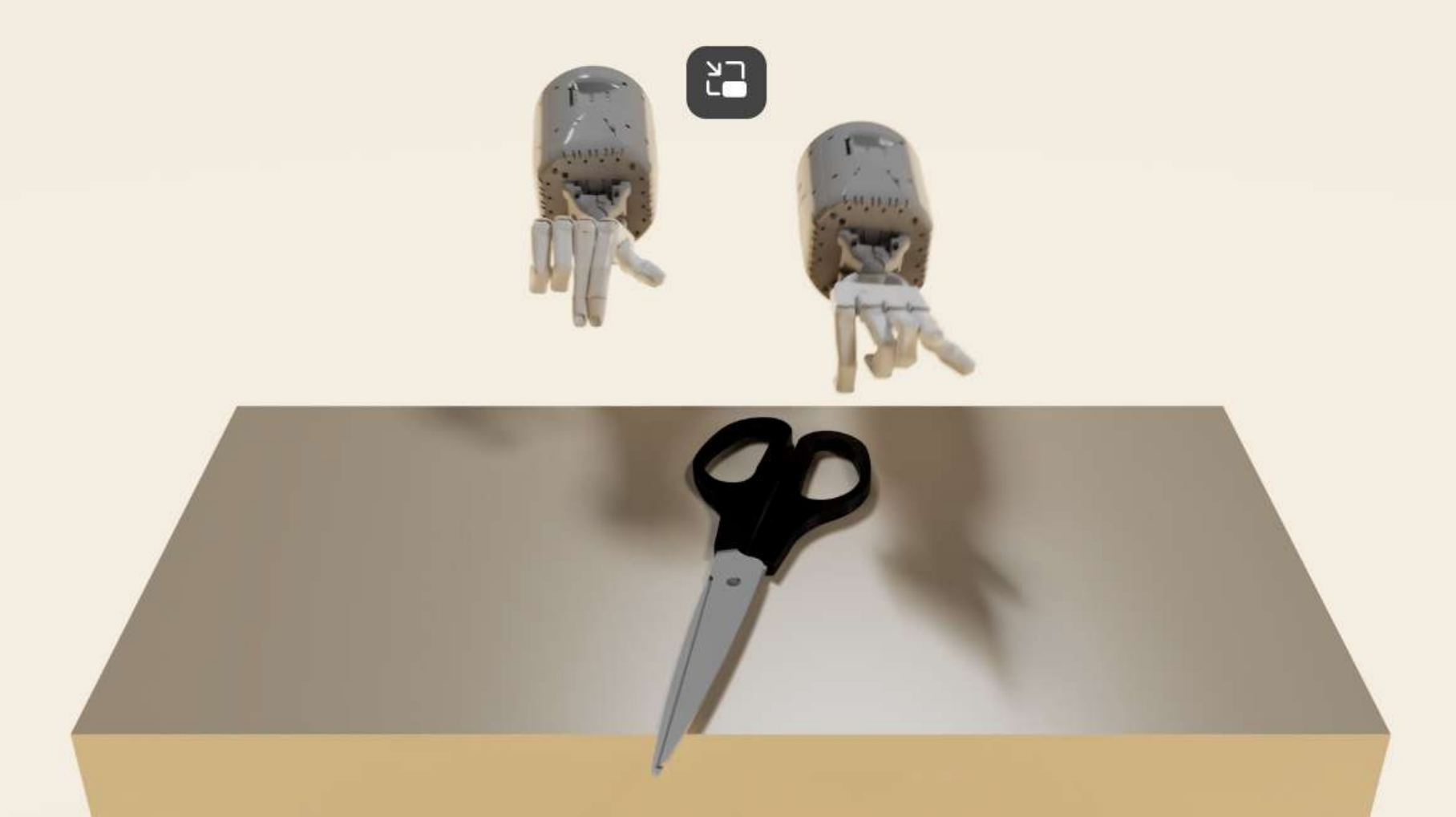} 
\end{minipage}
\vspace{8pt} \hrule \vspace{5pt}

\noindent
\begin{minipage}[t]{0.76\textwidth}
    \vspace{0pt} 
    \textbf{SwingCup (417, 52)} \\
    This class corresponds to the SwingCup task. This environment involves two hands and a dual handle cup, we need to use two hands to hold and swing the cup together in a synchronized motion.
\end{minipage}
\hfill
\begin{minipage}[t]{0.20\textwidth}
    \vspace{0pt} \centering
    \includegraphics[width=\linewidth]{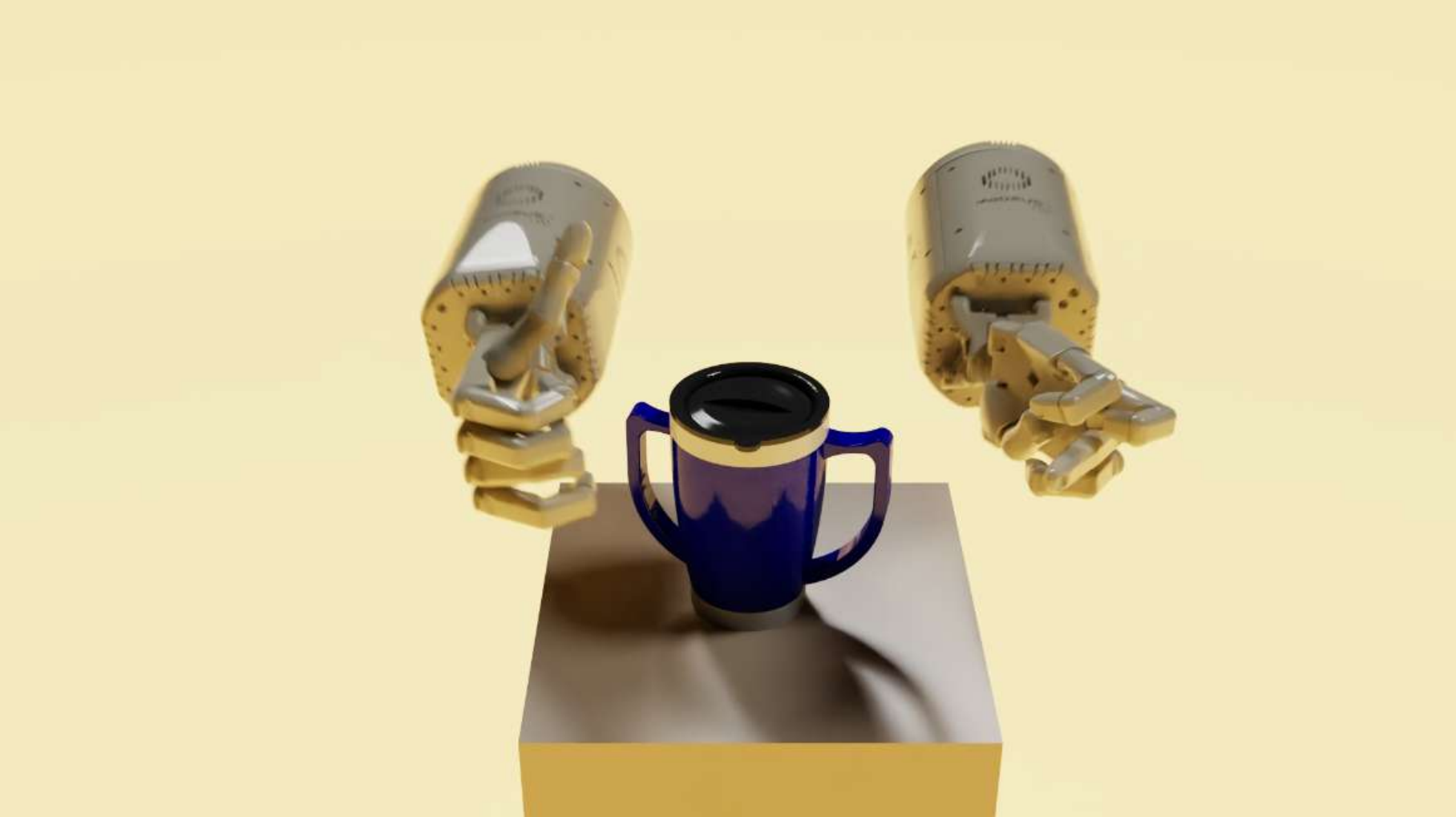} 
\end{minipage}
\vspace{8pt} \hrule \vspace{5pt}

\noindent
\begin{minipage}[t]{0.76\textwidth}
    \vspace{0pt} 
    \textbf{Switch (417, 52)} \\
    This class corresponds to the Switch task. This environment involves dual hands and a bottle, we need to use dual hand fingers to press the desired button with precision.
\end{minipage}
\hfill
\begin{minipage}[t]{0.20\textwidth}
    \vspace{0pt} \centering
    \includegraphics[width=\linewidth]{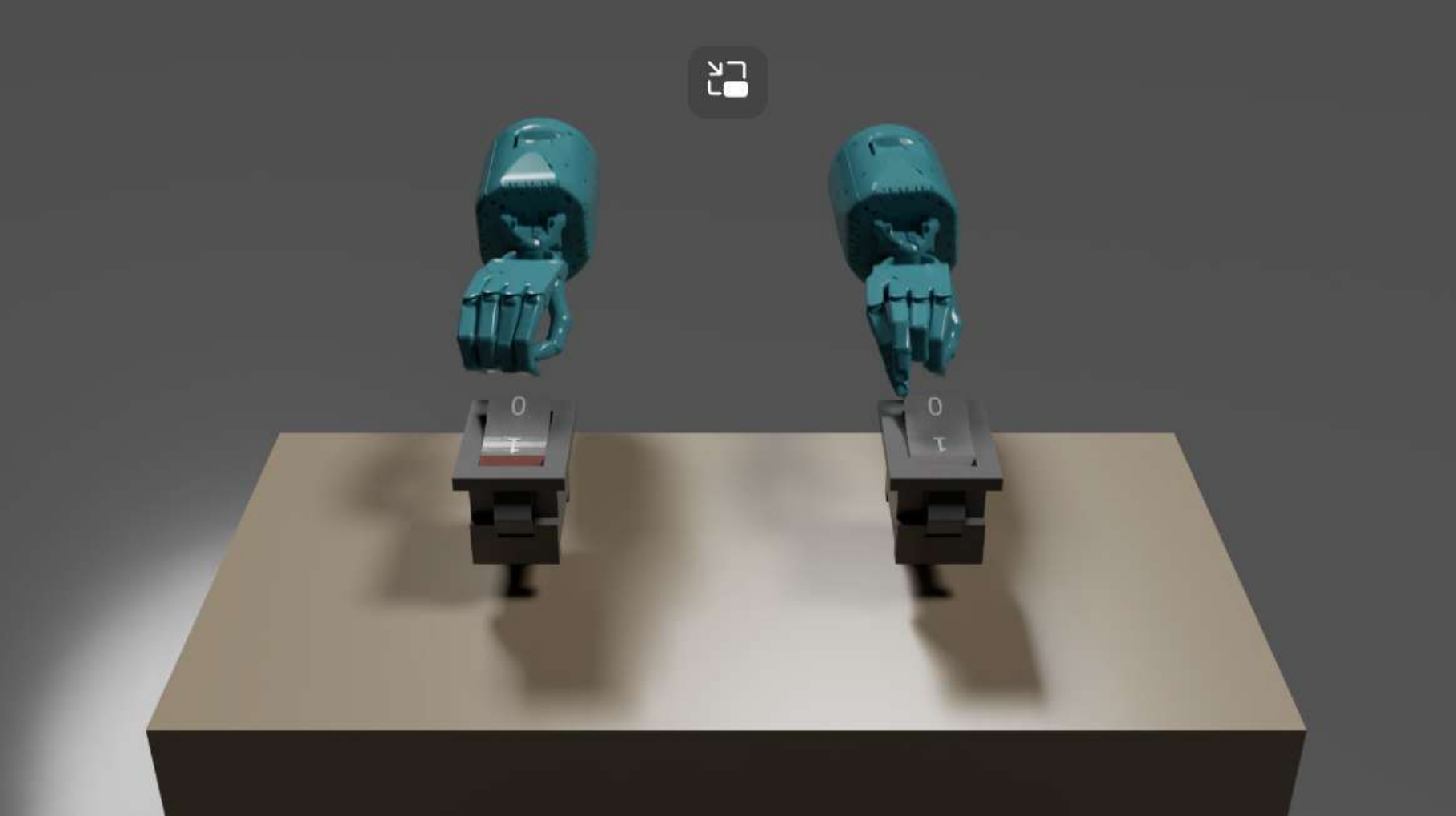} 
\end{minipage}
\vspace{8pt} \hrule \vspace{5pt}

\noindent
\begin{minipage}[t]{0.76\textwidth}
    \vspace{0pt} 
    \textbf{Kettle (417, 52)} \\
    The Kettle environment corresponds to the internal PourWater task. This environment involves two hands, a kettle, and a bucket. The agent needs to hold the kettle with one hand and the bucket with the other hand, and pour water from the kettle into the bucket.
\end{minipage}
\hfill
\begin{minipage}[t]{0.20\textwidth}
    \vspace{0pt} \centering
    \includegraphics[width=\linewidth]{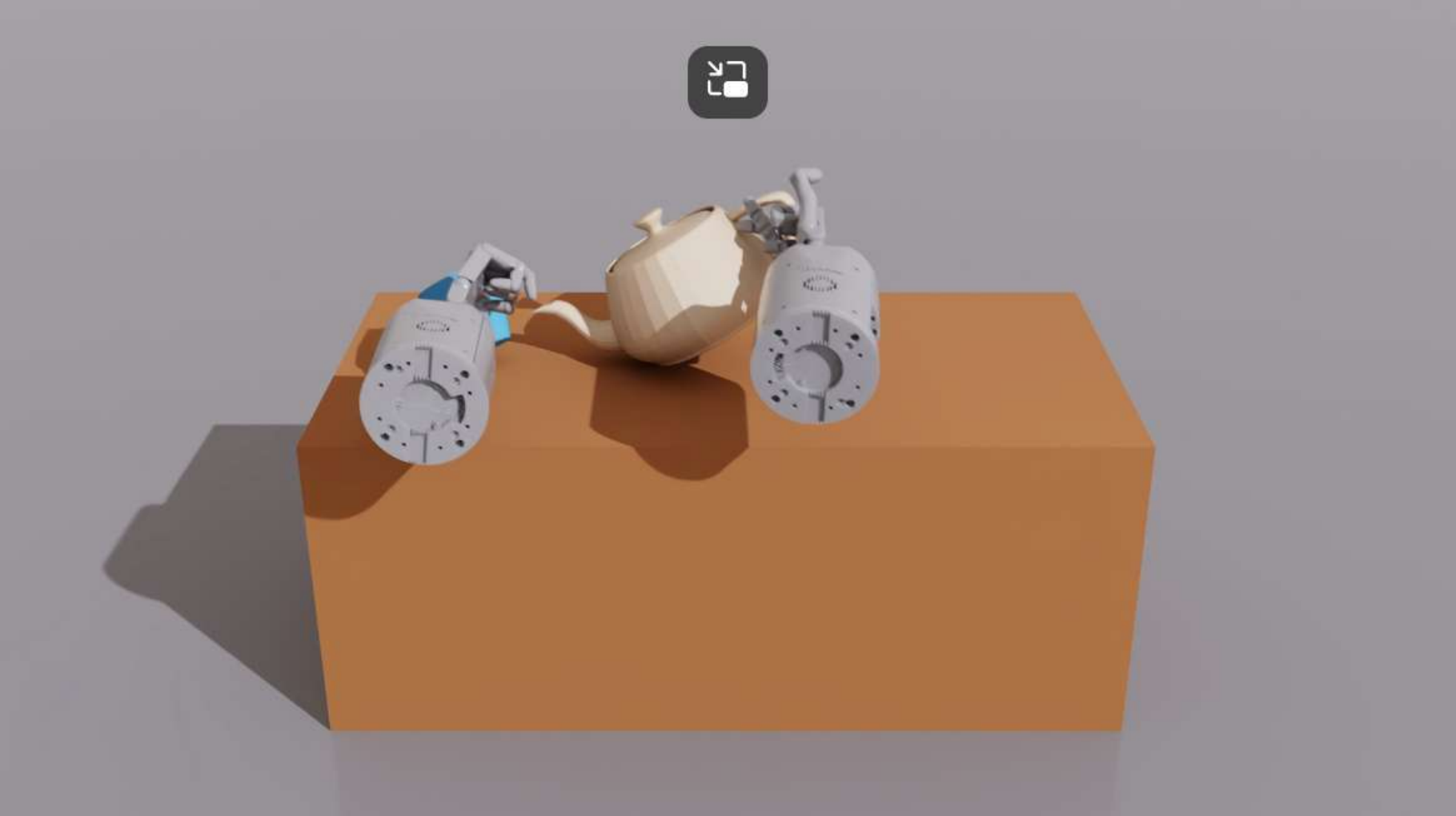} 
\end{minipage}
\vspace{8pt} \hrule \vspace{5pt}

\noindent
\begin{minipage}[t]{0.76\textwidth}
    \vspace{0pt} 
    \textbf{LiftUnderarm (417, 52)} \\
    This class corresponds to the LiftUnderarm task. This environment requires grasping the pot handle with two hands and lifting the pot to the designated position, simulating daily lifting skills.
\end{minipage}
\hfill
\begin{minipage}[t]{0.20\textwidth}
    \vspace{0pt} \centering
    \includegraphics[width=\linewidth]{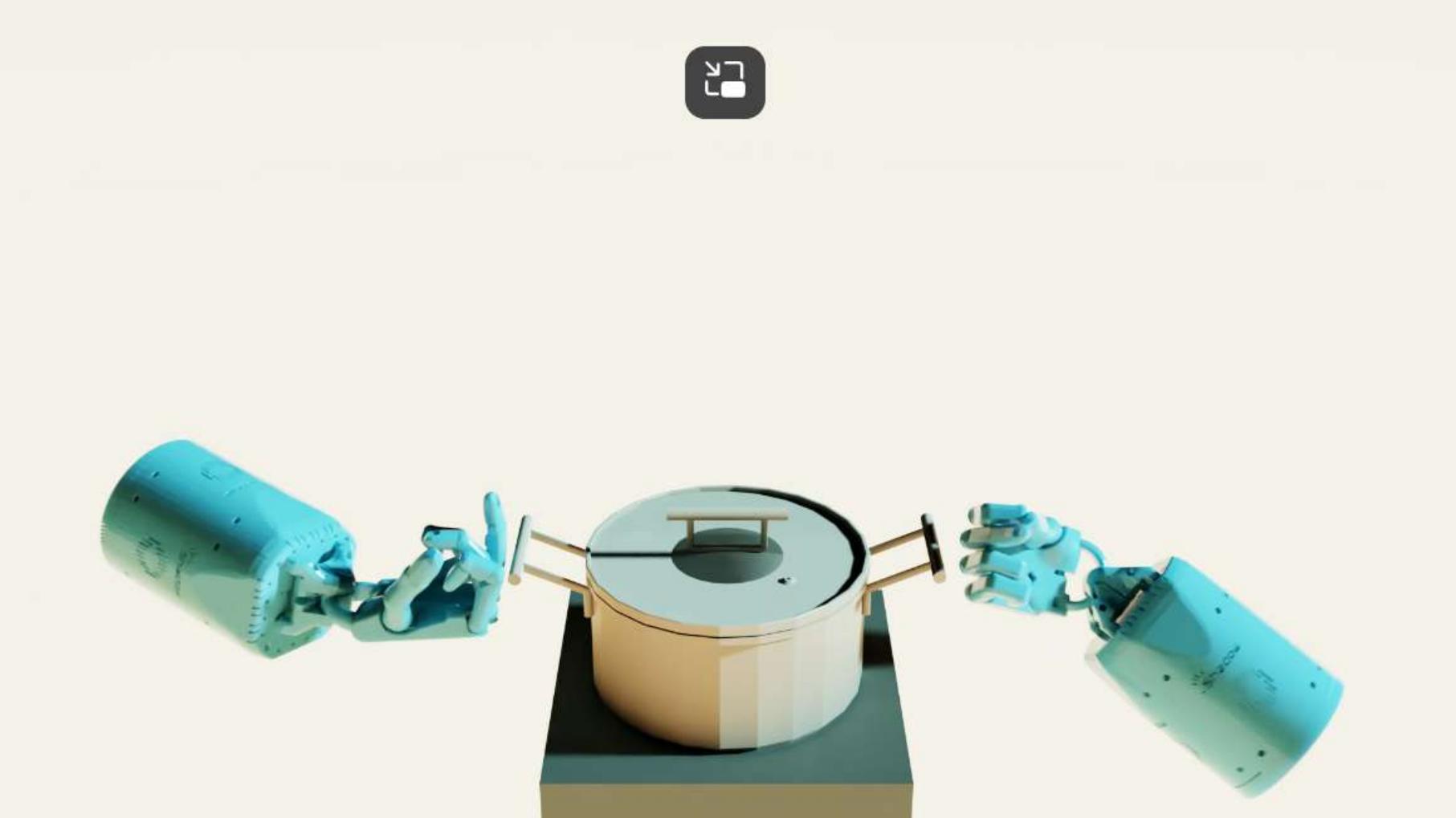} 
\end{minipage}
\vspace{8pt} \hrule \vspace{5pt}

\noindent
\begin{minipage}[t]{0.76\textwidth}
    \vspace{0pt} 
    \textbf{Pen (417, 52)} \\
    The Pen environment corresponds to the internal Open Pen Cap task. This environment involves two hands and a pen. The agent needs to use both hands to open the pen cap through fine-grained coordination.
\end{minipage}
\hfill
\begin{minipage}[t]{0.20\textwidth}
    \vspace{0pt} \centering
    \includegraphics[width=\linewidth]{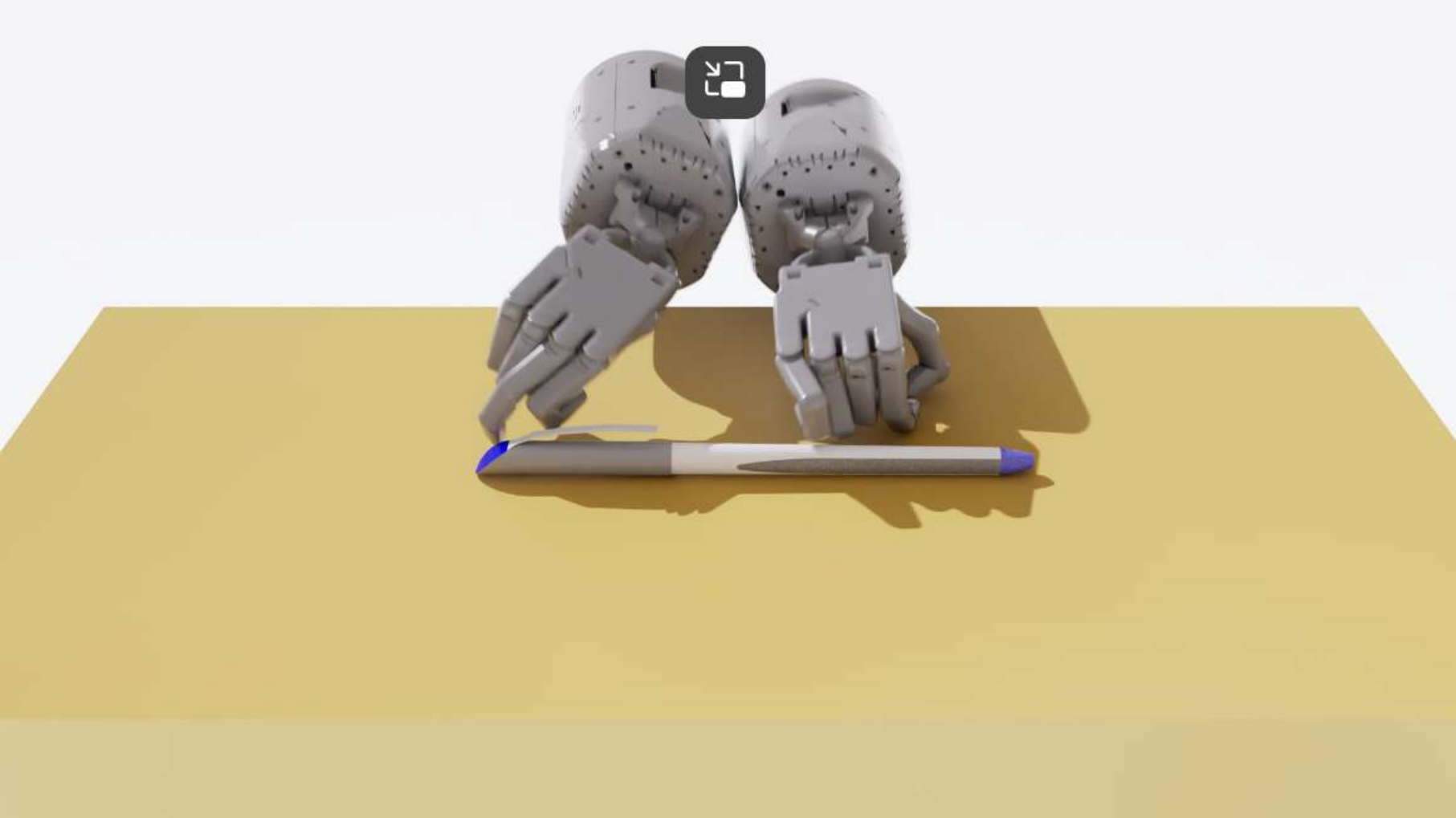} 
\end{minipage}
\vspace{8pt} \hrule \vspace{5pt}

\noindent
\begin{minipage}[t]{0.76\textwidth}
    \vspace{0pt} 
    \textbf{BottleCap (420, 52)} \\
    This class corresponds to the Bottle Cap task. This environment involves two hands and a bottle, we need to hold the bottle with one hand and open the bottle cap with the other hand without dropping the cap.
\end{minipage}
\hfill
\begin{minipage}[t]{0.20\textwidth}
    \vspace{0pt} \centering
    \includegraphics[width=\linewidth]{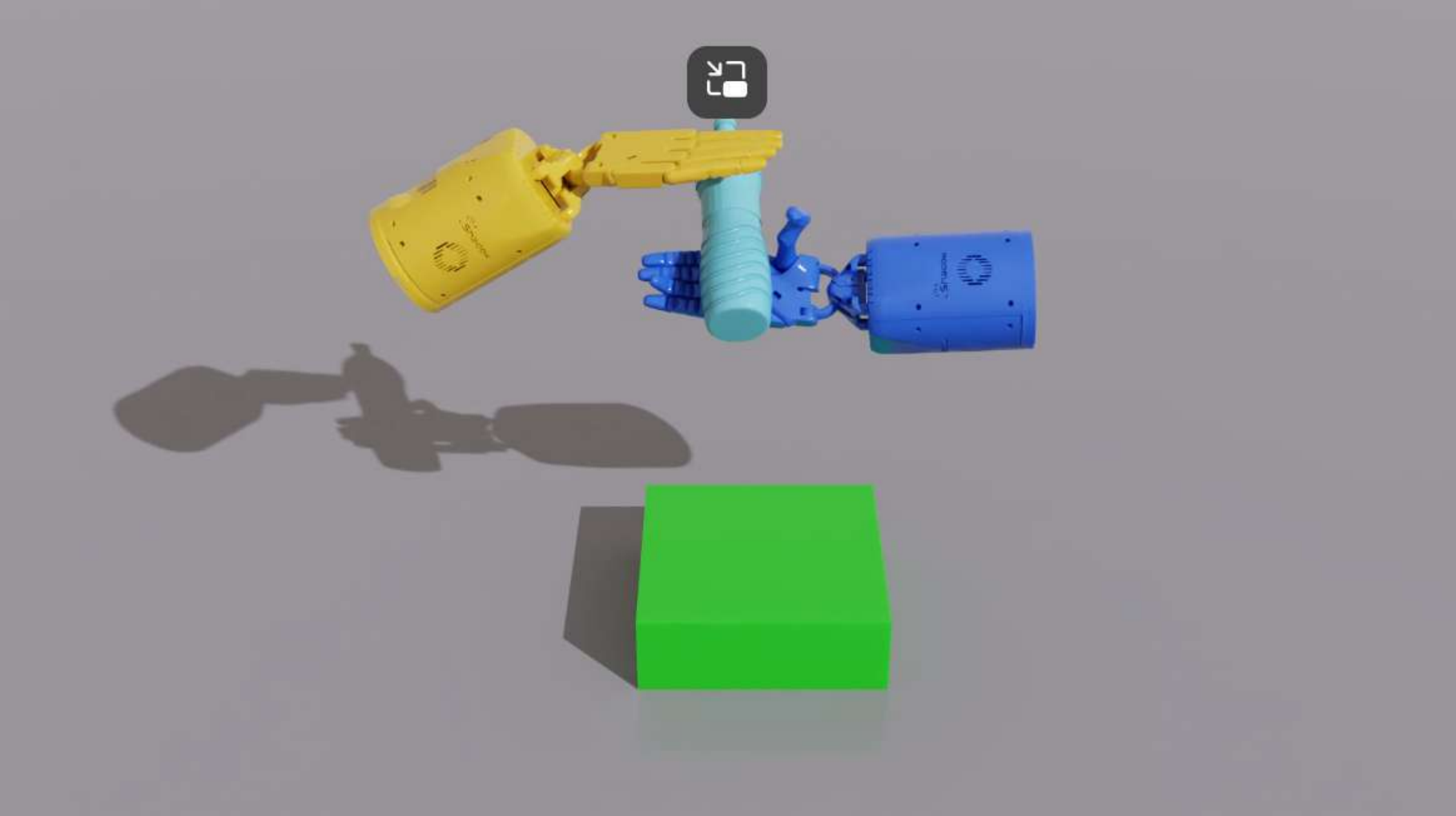} 
\end{minipage}
\vspace{8pt} \hrule \vspace{5pt}

\noindent
\begin{minipage}[t]{0.76\textwidth}
    \vspace{0pt} 
    \textbf{CatchAbreast (422, 52)} \\
    This environment consists of two shadow hands placed side by side and an object to be passed. It simulates two hands of the same person passing objects, requiring complex rotation and translation.
\end{minipage}
\hfill
\begin{minipage}[t]{0.20\textwidth}
    \vspace{0pt} \centering
    \includegraphics[width=\linewidth]{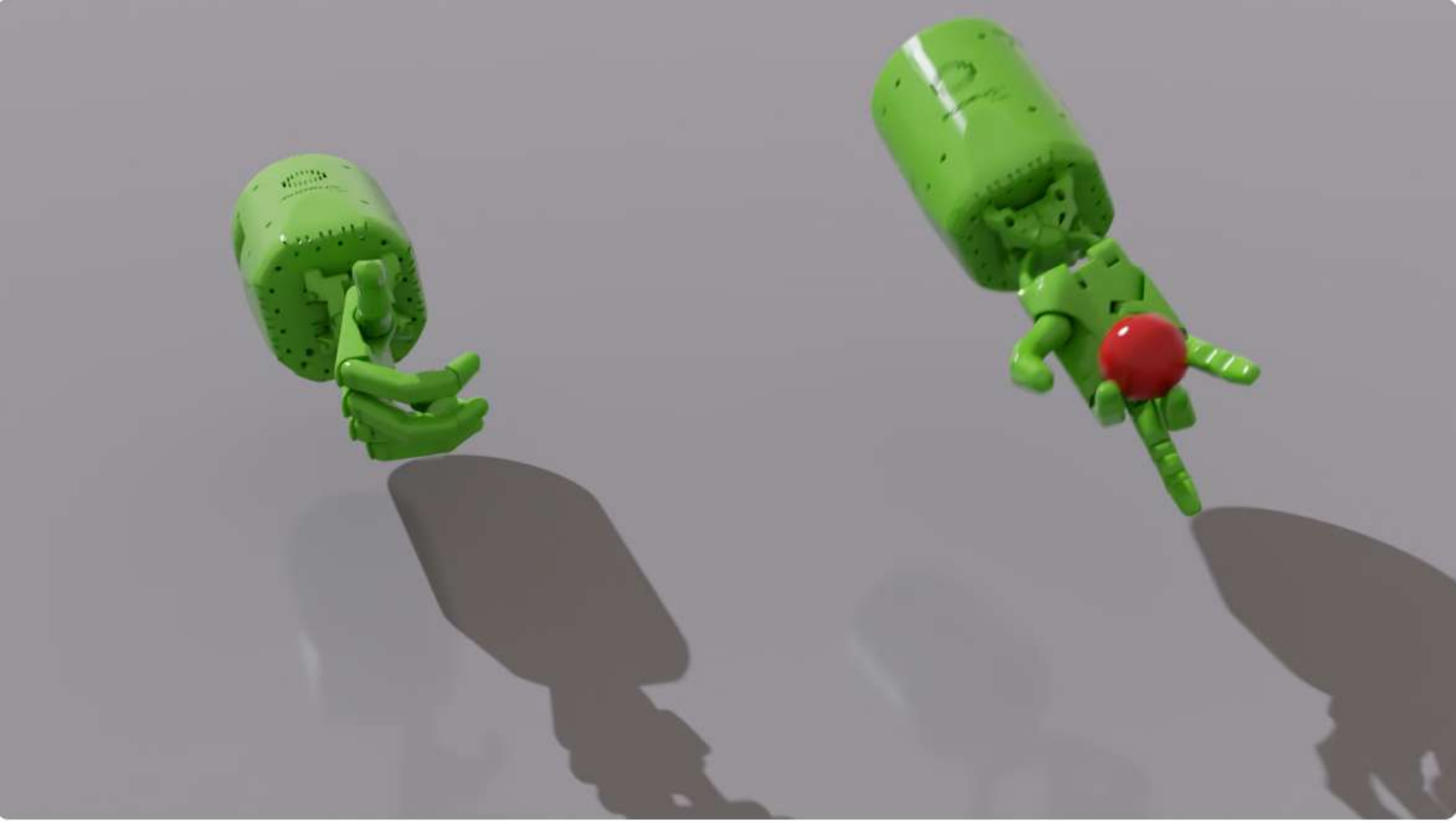} 
\end{minipage}
\vspace{8pt} \hrule \vspace{5pt}

\noindent
\begin{minipage}[t]{0.76\textwidth}
    \vspace{0pt} 
    \textbf{CatchOver2Underarm (422, 52)} \\
    This environment requires two objects to be thrown into the other hand at the same time, necessitating higher manipulation techniques than single-object tasks.
\end{minipage}
\hfill
\begin{minipage}[t]{0.20\textwidth}
    \vspace{0pt} \centering
    \includegraphics[width=\linewidth]{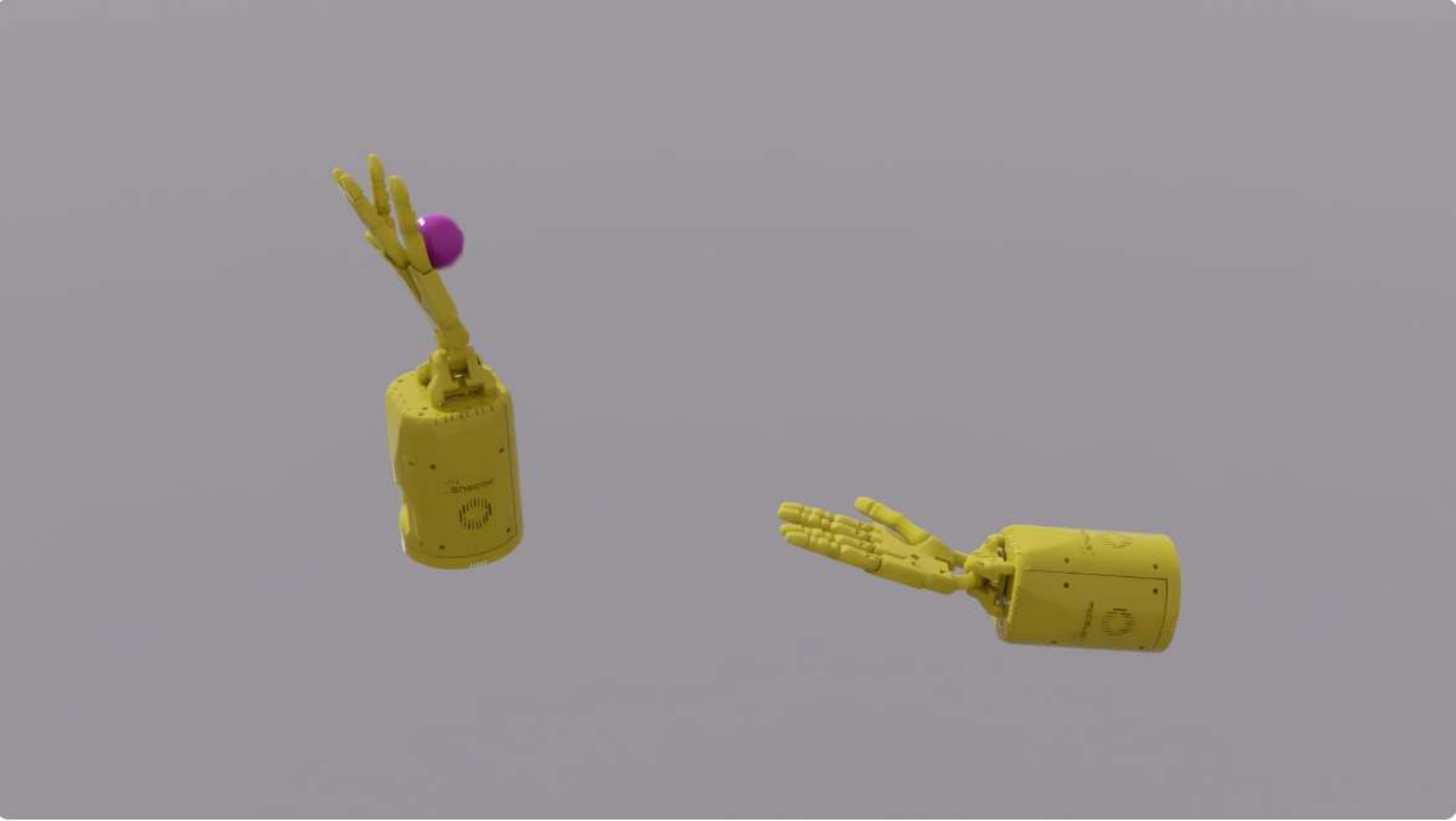} 
\end{minipage}
\vspace{8pt} \hrule \vspace{5pt}

\noindent
\begin{minipage}[t]{0.76\textwidth}
    \vspace{0pt} 
    \textbf{CatchUnderarm (422, 52)} \\
    In this task, two hands pass an object from one palm to the other. Unlike HandOver, the hands' translation and rotation degrees of freedom are added to the action space.
\end{minipage}
\hfill
\begin{minipage}[t]{0.20\textwidth}
    \vspace{0pt} \centering
    \includegraphics[width=\linewidth]{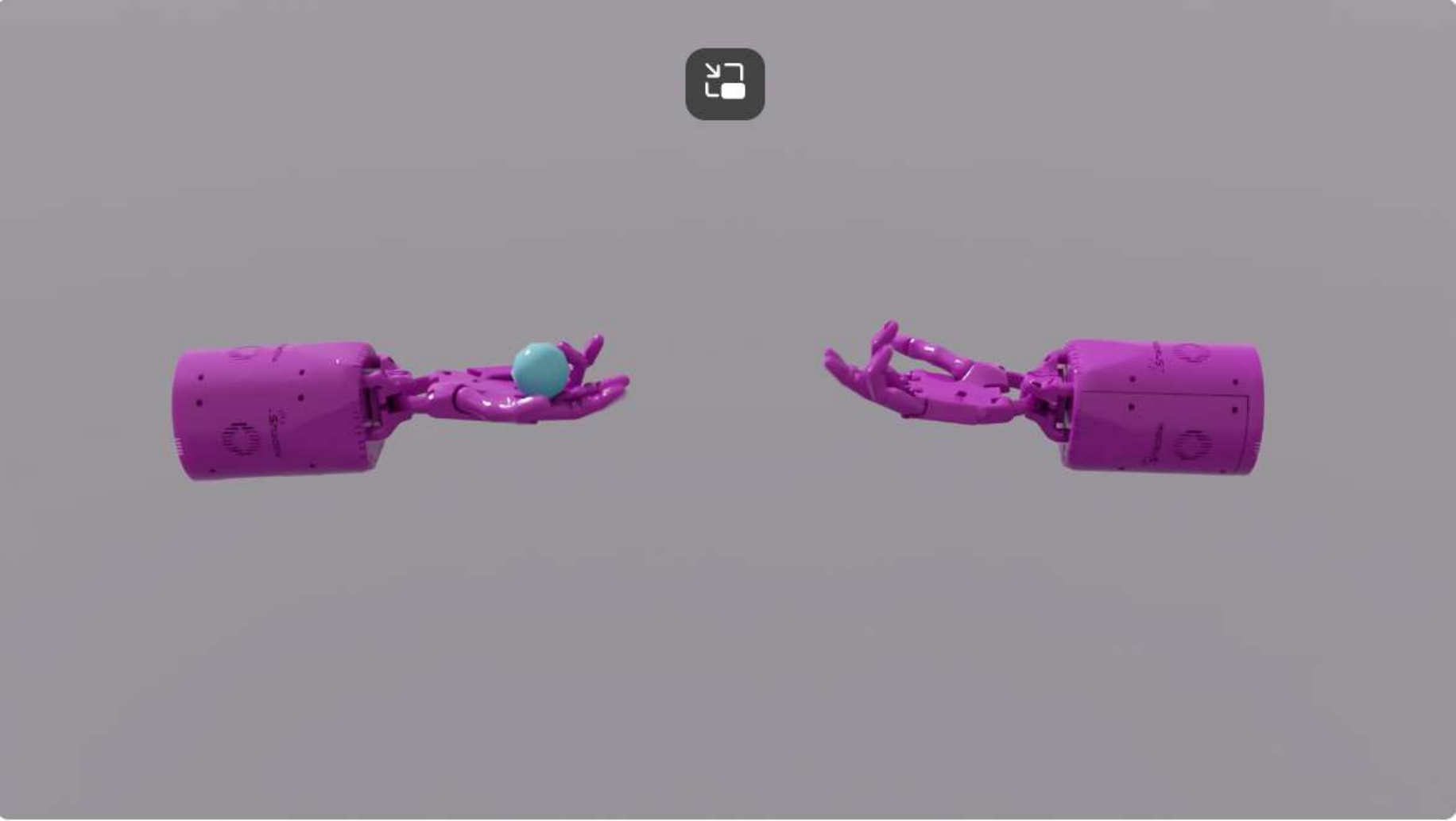} 
\end{minipage}
\vspace{8pt} \hrule \vspace{5pt}

\noindent
\begin{minipage}[t]{0.76\textwidth}
    \vspace{0pt} 
    \textbf{ReOrientation (422, 40)} \\
    Each hand holds an object and must independently reorient it to match a specified target orientation.
\end{minipage}
\hfill
\begin{minipage}[t]{0.20\textwidth}
    \vspace{0pt} \centering
    \includegraphics[width=\linewidth]{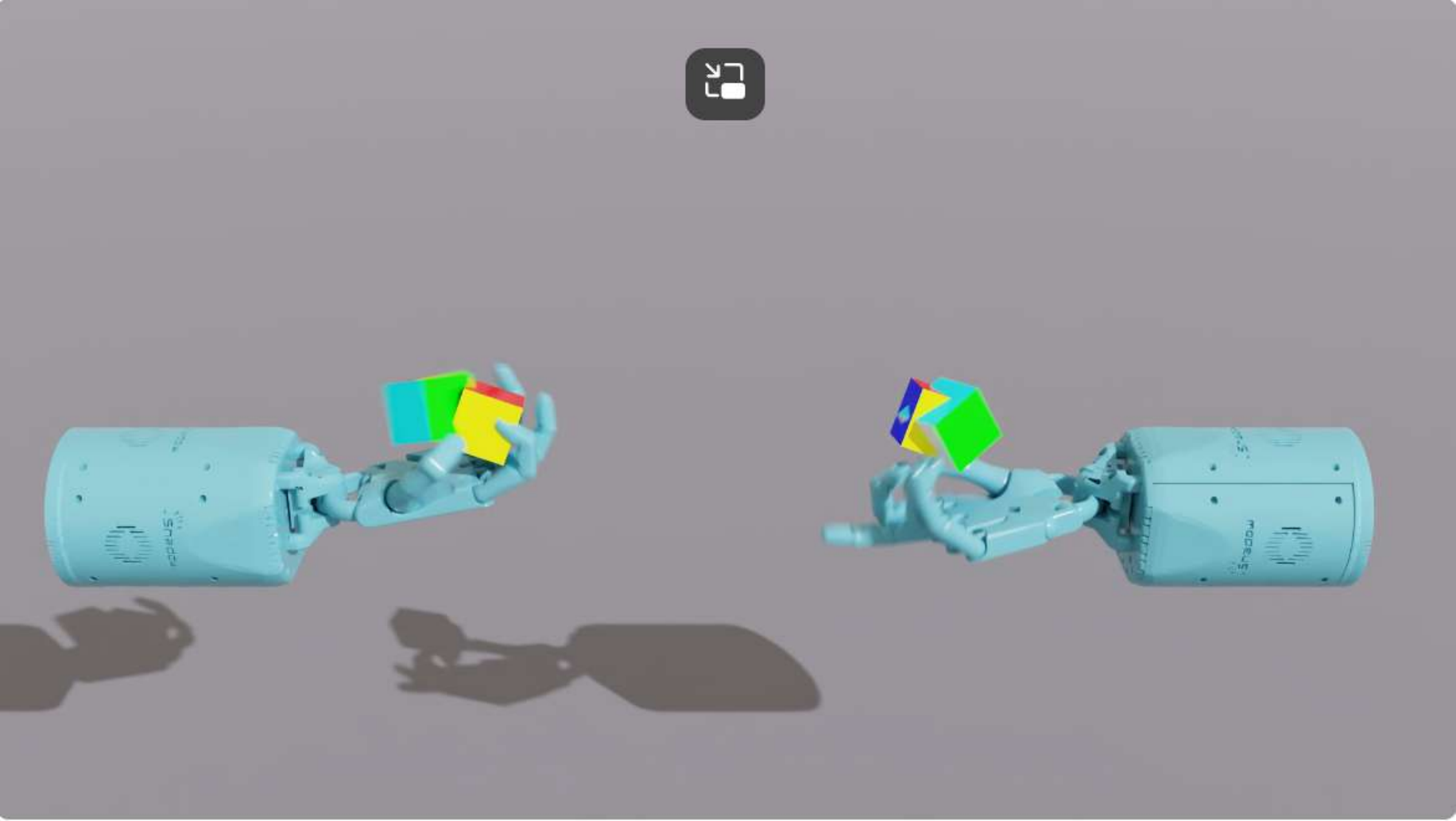} 
\end{minipage}
\vspace{8pt} \hrule \vspace{5pt}

\noindent
\begin{minipage}[t]{0.76\textwidth}
    \vspace{0pt} 
    \textbf{GraspAndPlace (425, 52)} \\
    This environment consists of dual-hands, an object and a bucket, requiring the agent to pick up the object and accurately place it into the bucket.
\end{minipage}
\hfill
\begin{minipage}[t]{0.20\textwidth}
    \vspace{0pt} \centering
    \includegraphics[width=\linewidth]{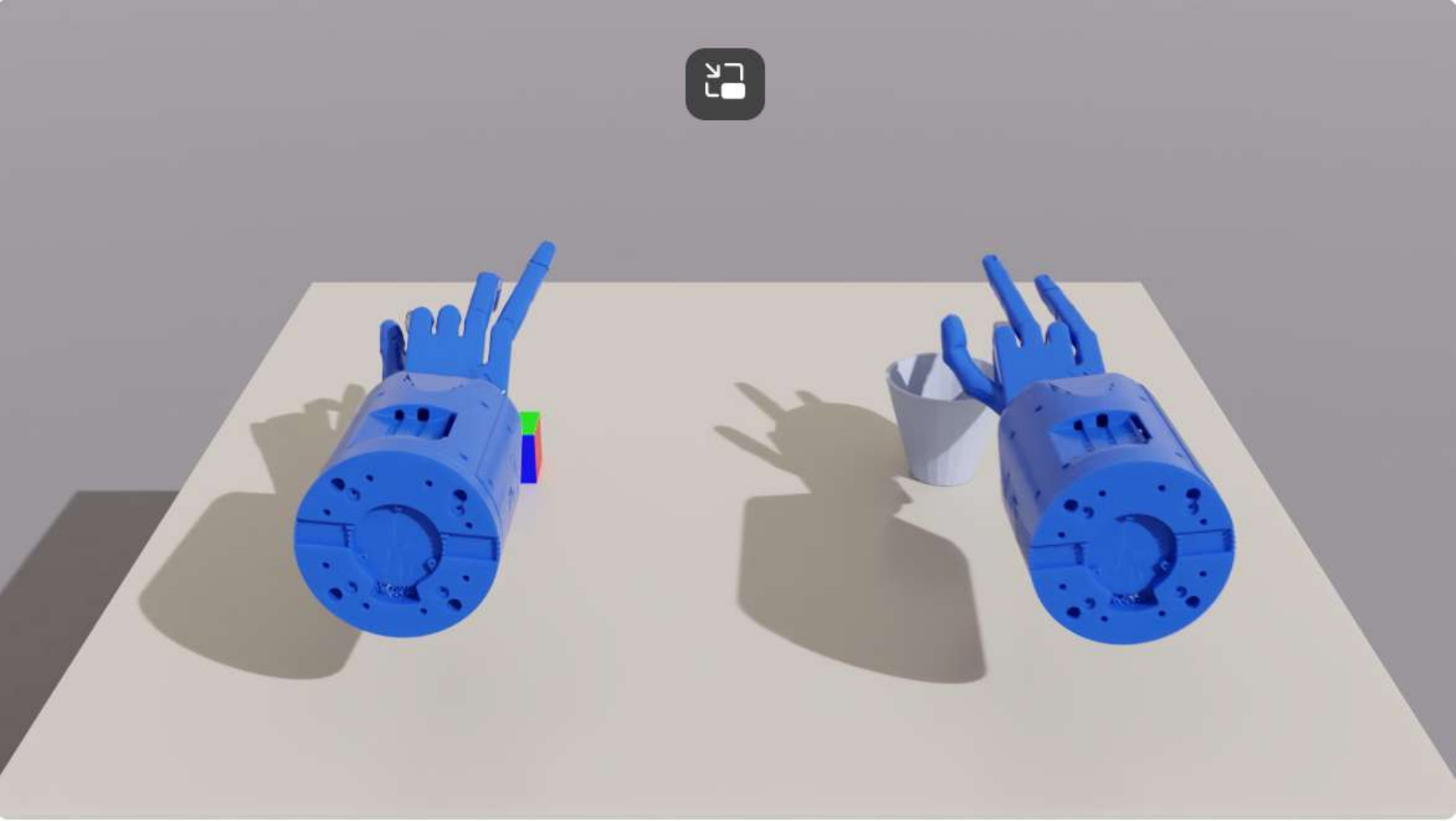} 
\end{minipage}
\vspace{8pt} \hrule \vspace{5pt}

\noindent
\begin{minipage}[t]{0.76\textwidth}
    \vspace{0pt} 
    \textbf{BlockStack (428, 52)} \\
    Involves dual hands and two blocks. The agent must stack the blocks to form a stable tower.
\end{minipage}
\hfill
\begin{minipage}[t]{0.20\textwidth}
    \vspace{0pt} \centering
    \includegraphics[width=\linewidth]{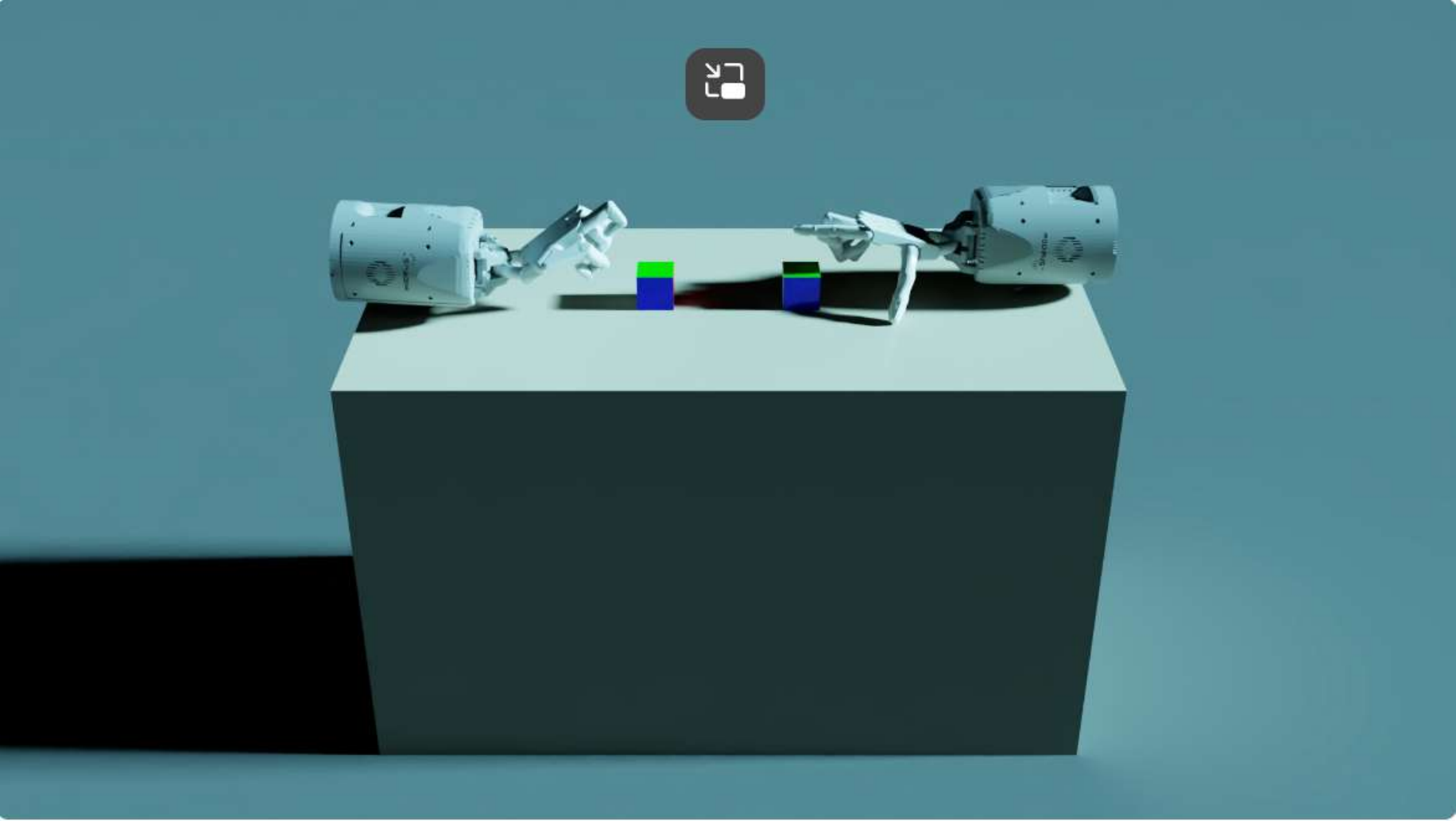} 
\end{minipage}
\vspace{8pt} \hrule \vspace{5pt}

\noindent
\begin{minipage}[t]{0.76\textwidth}
    \vspace{0pt} 
    \textbf{PushBlock (428, 52)} \\
    Two hands must separately reach and push two blocks to their desired goals. This is a basic coordination task.
\end{minipage}
\hfill
\begin{minipage}[t]{0.20\textwidth}
    \vspace{0pt} \centering
    \includegraphics[width=\linewidth]{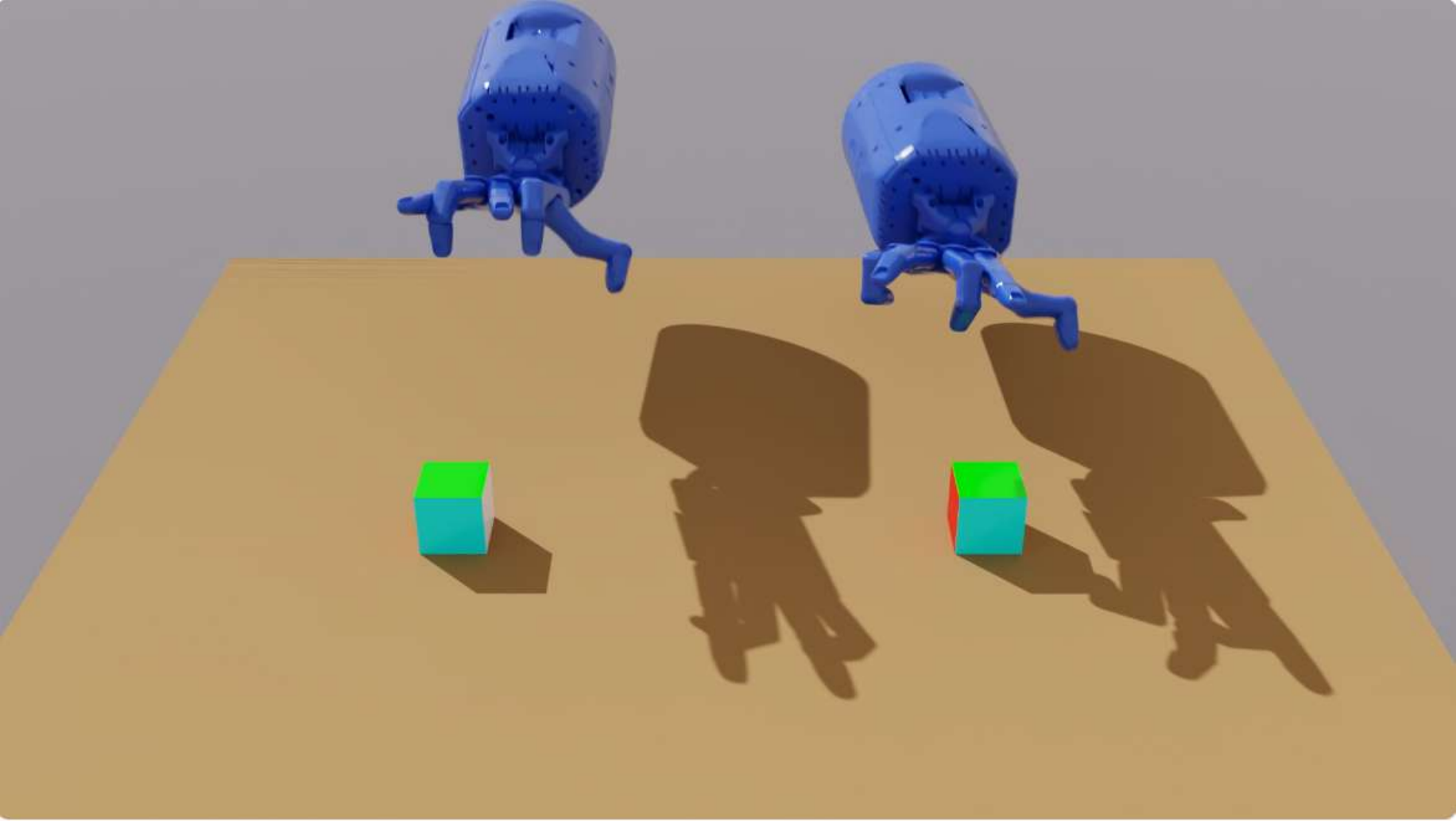} 
\end{minipage}
\vspace{8pt} \hrule \vspace{5pt}

\noindent
\begin{minipage}[t]{0.76\textwidth}
    \vspace{0pt} 
    \textbf{TwoCatchUnderarm (446, 52)} \\
    Similar to Catch Underarm but with two objects handled simultaneously, requiring extremely high precision and timing for the exchange.
\end{minipage}
\hfill
\begin{minipage}[t]{0.20\textwidth}
    \vspace{0pt} \centering
    \includegraphics[width=\linewidth]{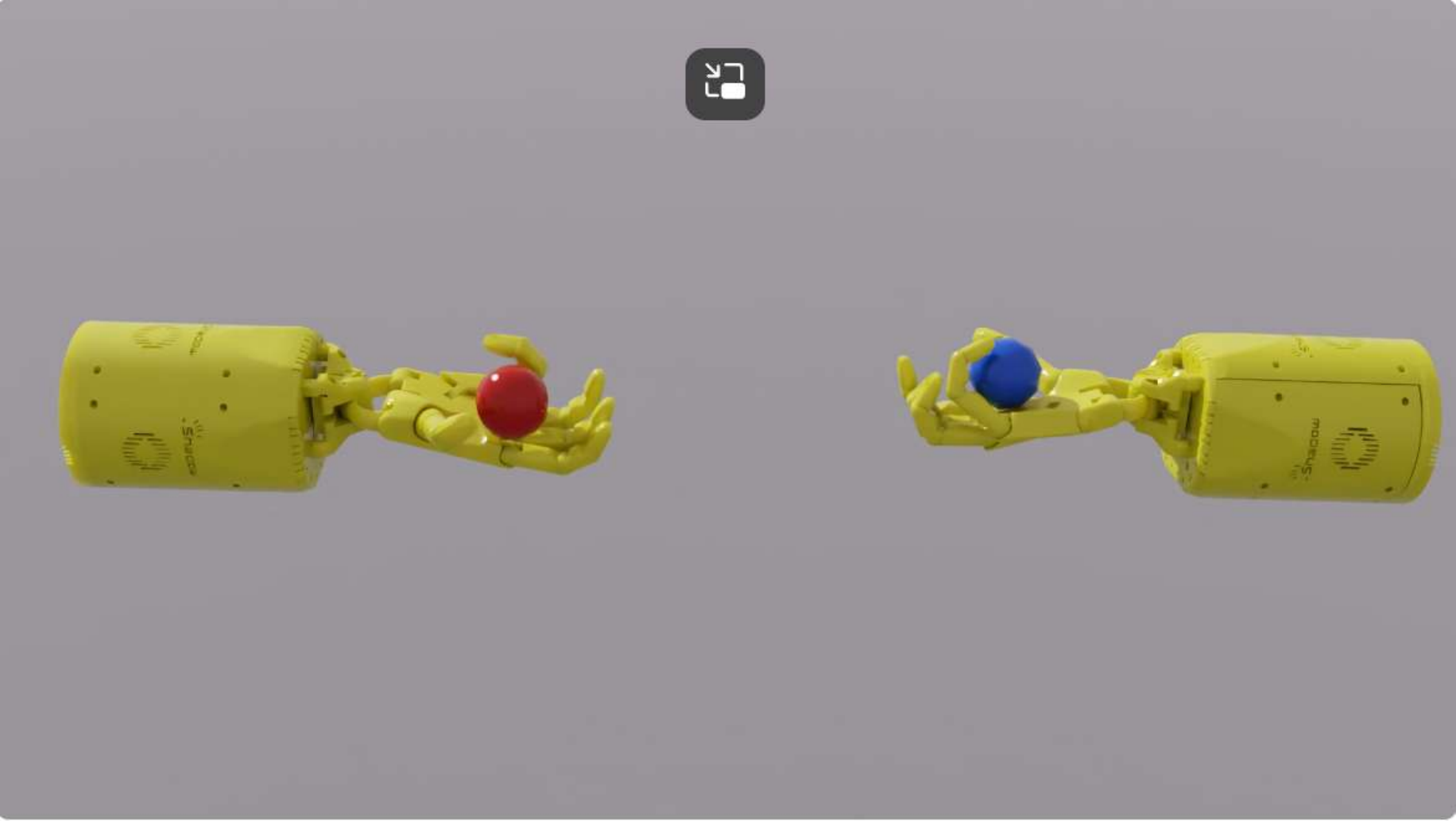} 
\end{minipage}
\vspace{8pt} \hrule \vspace{12pt}

\begin{center}
\hrule height 0.8pt 
\vspace{5pt}
     Maniskill2 Environments
\end{center}
\hrule height 0.8pt 

\vspace{10pt}

\noindent
\begin{minipage}[t]{0.78\textwidth}
    \vspace{0pt} 
    Environment (obs dim, action dim)\\
    Task description 
\end{minipage}
\vspace{4pt}
\hrule 
\vspace{12pt}

\noindent
\begin{minipage}[t]{0.76\textwidth}
    \vspace{0pt} 
    \textbf{PickCube-v0 (51, 7)} \\
    This class corresponds to the PickCube task in ManiSkill. This environment consists of a robot arm and a cube placed on the table. At the beginning, the cube appears at a random location and orientation. The agent must control the gripper to approach, grasp, and lift the cube above a threshold height. The challenge lies in object localization, precise control, and stable grasping
\end{minipage}
\hfill
\begin{minipage}[t]{0.20\textwidth}
    \vspace{0pt} \centering
    \includegraphics[width=\linewidth]{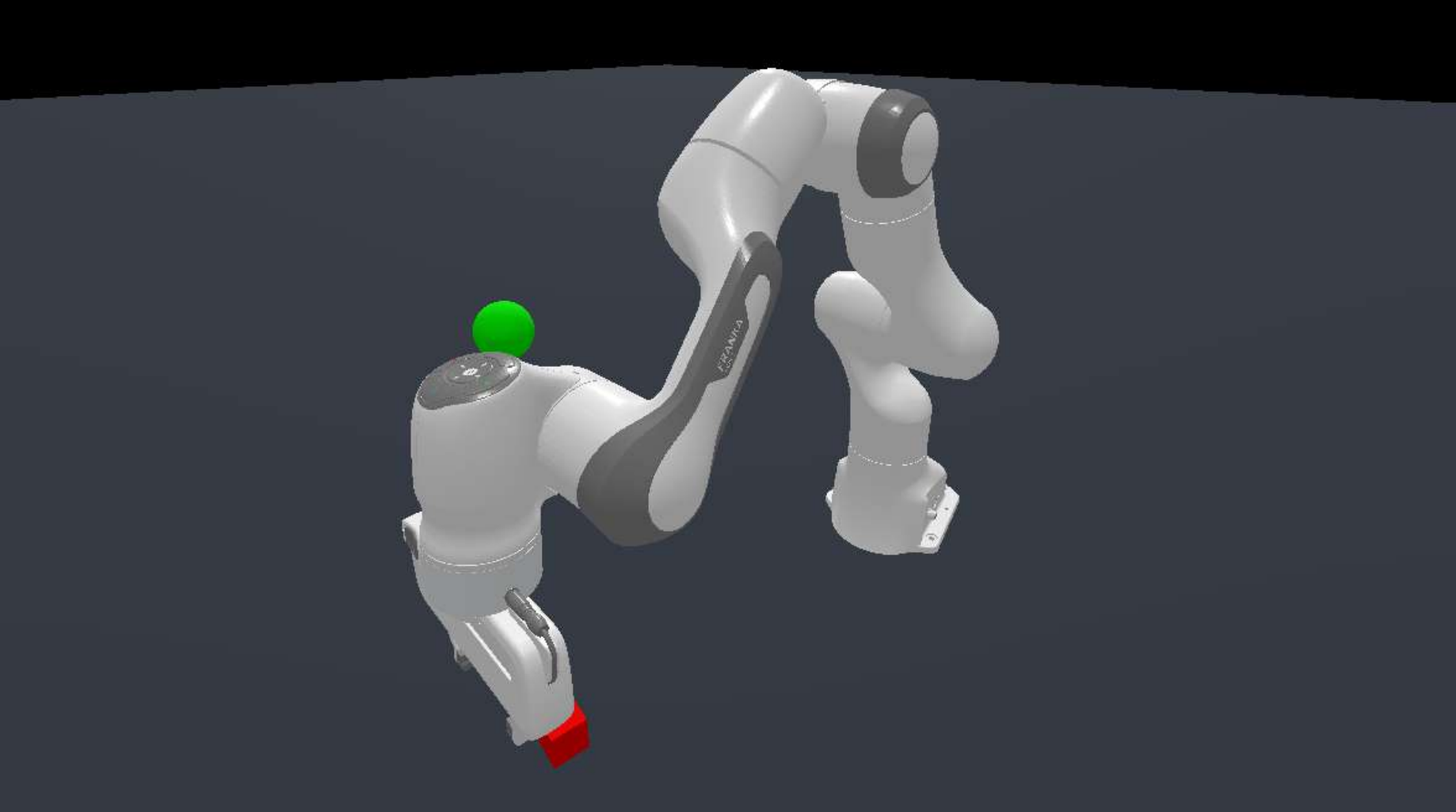} 
\end{minipage}
\vspace{8pt} \hrule \vspace{5pt}

\noindent
\begin{minipage}[t]{0.76\textwidth}
    \vspace{0pt} 
    \textbf{TurnFaucet-v0 (40, 7)} \\
    This class corresponds to the TurnFaucet task. A faucet handle is mounted on a wall, and the agent must rotate it clockwise or counterclockwise to a target angle. The challenge lies in establishing proper contact, applying sufficient torque, and maintaining stability during the turning motion
\end{minipage}
\hfill
\begin{minipage}[t]{0.20\textwidth}
    \vspace{0pt} \centering
    \includegraphics[width=\linewidth]{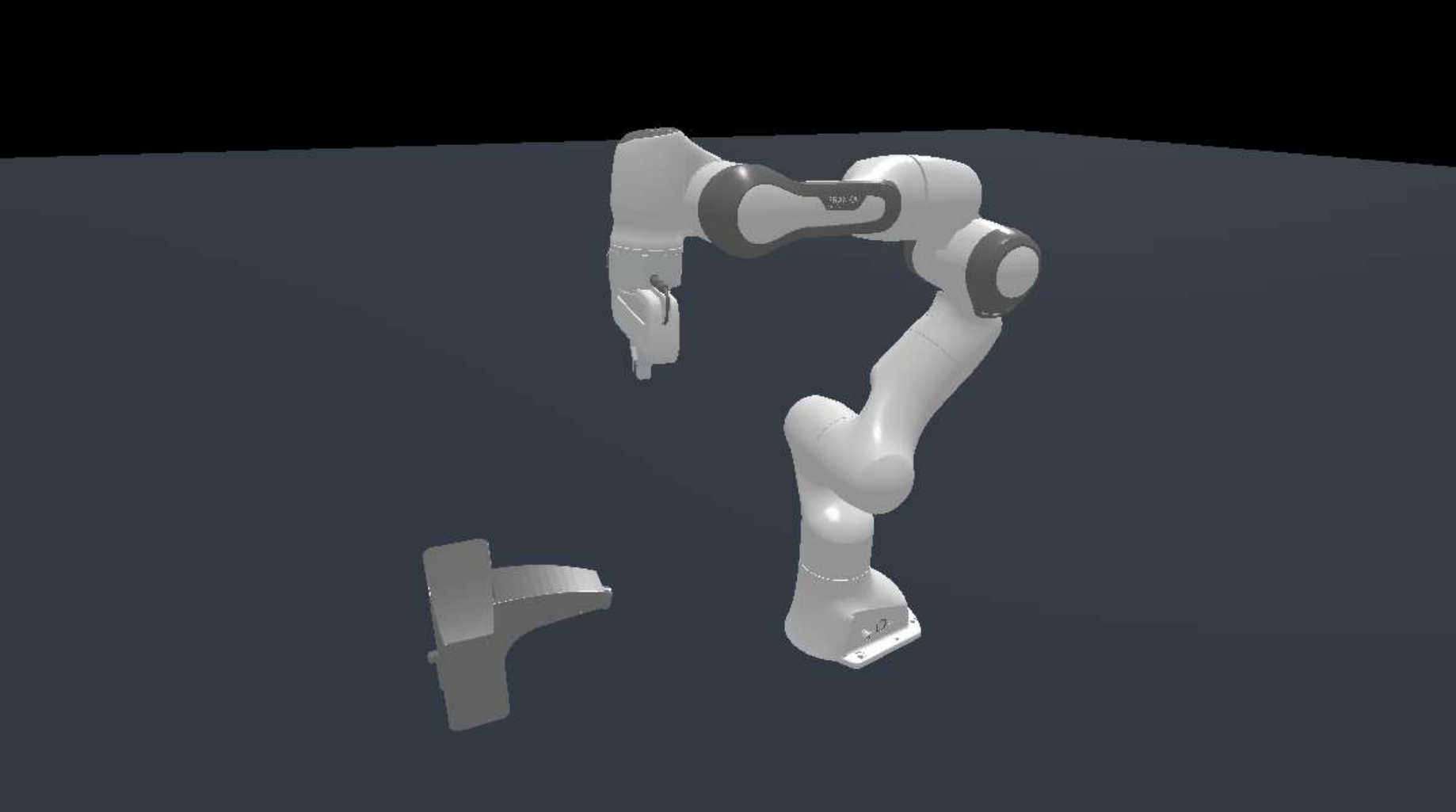} 
\end{minipage}
\vspace{8pt} \hrule \vspace{5pt}

\noindent
\begin{minipage}[t]{0.76\textwidth}
    \vspace{0pt} 
    \textbf{OpenCabinetDrawer-v1 (75, 11)} \\
    This class corresponds to the OpenCabinetDrawer task. The robot must open a drawer embedded in a cabinet by locating the handle and pulling it outward. The task requires both accurate handle grasping and force application along a linear trajectory, while avoiding excessive torque that could misalign the drawer
\end{minipage}
\hfill
\begin{minipage}[t]{0.20\textwidth}
    \vspace{0pt} \centering
    \includegraphics[width=\linewidth]{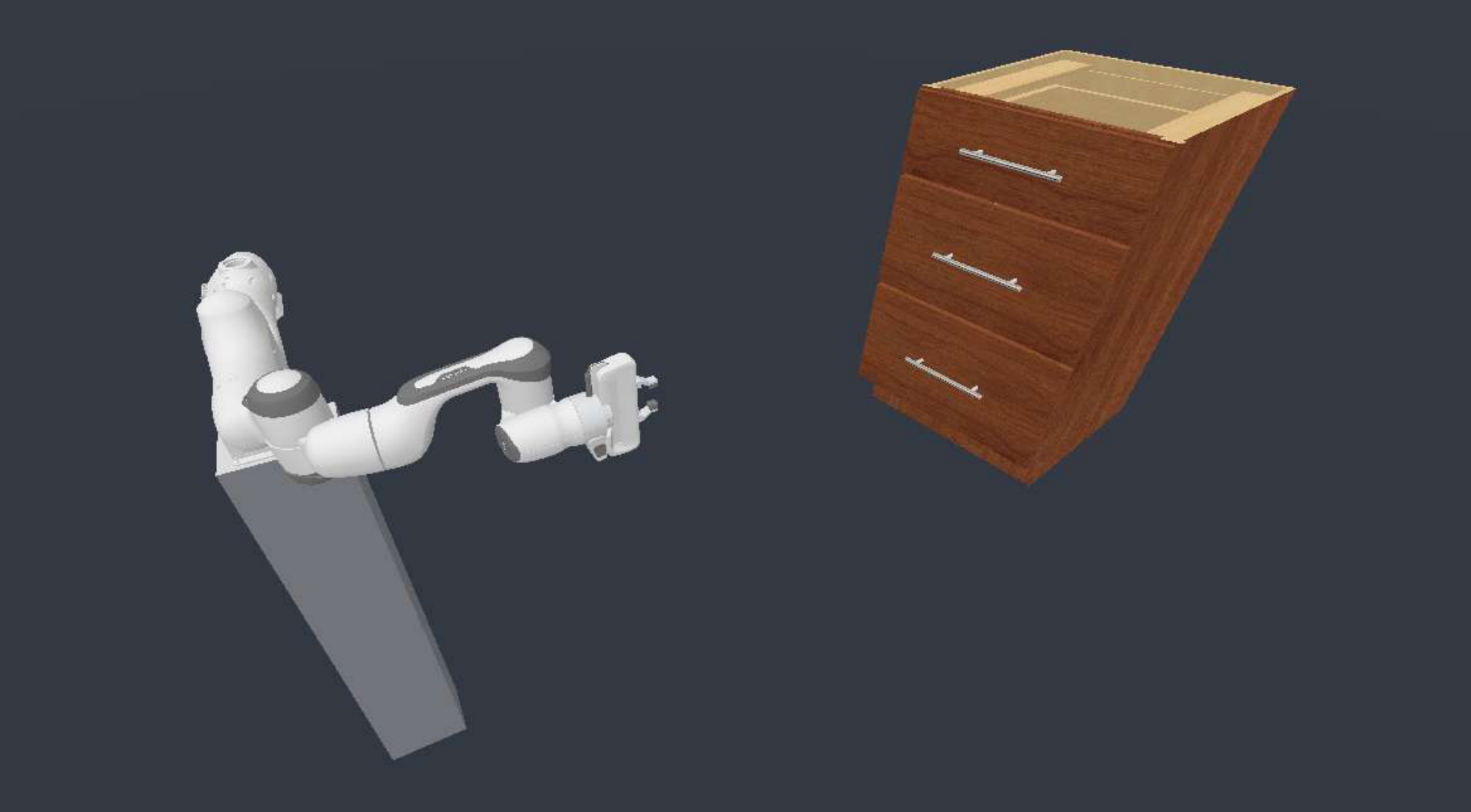} 
\end{minipage}
\vspace{8pt} \hrule \vspace{5pt}

\section{Theoretical Analysis of Mechanism Decoupling}
\label{app:theory}

In this section, we provide the theoretical motivation for the mechanism decoupling formulation and analyze the approximation gap of monolithic static causal graphs in reward distillation. 
The analysis is intended to characterize why a static graph may be insufficient when state-dependent transient interactions exist, rather than to prove exact recovery of the full physical dynamics.

\subsection{Structural Optimization Bottleneck}
\label{app:theory_sob}

The decoupling architecture is motivated by the \textit{Independent Causal Mechanisms} (ICM) principle. 
In embodied systems, the reward-relevant skeleton can be viewed as a stable structural abstraction retained for reward generation, while state-dependent mechanism activations capture contextual variations observed during transitions. 
This skeleton is not intended to represent the complete time-lagged physical dynamics, which may contain feedback loops and inertial couplings.

The formulation
\begin{equation}
M_{inst}(s_{t-1}) 
= 
M_{inv} \odot 
\Big( \mathbf{1}_{d \times d} + M_{trans}(s_{t-1}; \theta) \Big)
\end{equation}
implements this design through an optimization bottleneck:

\textbf{Stable Reward-Relevant Skeleton:} 
The acyclicity constraint $h(M_{inv})=0$ and the soft prior $\mathcal{L}_{soft}$ are applied exclusively to $M_{inv}$, guiding it to capture a compact reward-relevant prerequisite structure for downstream reward generation.

\textbf{State-Dependent Modulation:} 
The neural parameterization $M_{trans}$ is confined by the Hadamard mask to scaling the magnitudes of candidate edges based on the context $s_{t-1}$, without freely introducing unrelated edges outside the structure anchored by $M_{inv}$.

This structural bottleneck allows state-dependent variations to be modeled by $M_{trans}$ during reconstruction, while $M_{inv}$ retains the stable structure used as the CRWM for reward generation.

\subsection{Error Bounds of Mechanism Decoupling}
\label{app:theory_ebmd}

To analyze the effect of the dynamic gating mechanism, we study the approximation error of static versus dynamically modulated graphs under a locally linear instantiation. 
Consider the following transition model:
\begin{equation}
s_t = M^\circ_{inst}(s_{t-1})s_{t-1} + \epsilon_t, 
\quad 
\epsilon_t \sim \mathcal{N}(0, \sigma^2I),
\label{eq:theory_linear_sem}
\end{equation}
where \(M^\circ_{inst}\) denotes the oracle instantaneous graph in this locally linear setting, written as:
\begin{equation}
M^\circ_{inst}(s_{t-1}) 
= 
M^\circ_{inv} \odot 
\Big(\mathbf{1}_{d \times d} + M^\circ_{trans}(s_{t-1})\Big).
\end{equation}
Here, $M^\circ_{inv}$ denotes the stable reward-relevant backbone, and $M^\circ_{trans}$ denotes state-dependent transient modulation.

\textbf{Assumption 1.} 
\textit{$M^\circ_{trans}(s_{t-1})$ is statistically independent of $s_{t-1}$ and has zero mean, i.e., $\mathbb{E}[M^\circ_{trans}] = 0$ and $M^\circ_{trans} \perp\!\!\!\perp s_{t-1}$.}

The optimal static causal graph $\hat{M}_{static}$ is defined as:
\begin{equation}
\hat{M}_{static} 
= 
\arg\min_{M \in \mathcal{H}_{static}} 
\mathbb{E}\big[\|s_t - M s_{t-1}\|_2^2\big].
\label{eq:theory_static_opt}
\end{equation}
Under Assumption 1, $\hat{M}_{static} = M^\circ_{inv}$.

\textit{Proof Sketch.} 
The unconstrained solution is:
\begin{equation}
\hat{M} = \mathbb{E}[s_t s_{t-1}^\top]\Sigma_s^{-1},
\end{equation}
where $\Sigma_s=\mathbb{E}[s_{t-1}s_{t-1}^\top]$. 
Since
\begin{equation}
\mathbb{E}[M^\circ_{inst}]
=
M^\circ_{inv}\odot
\Big(\mathbf{1}+\mathbb{E}[M^\circ_{trans}]\Big)
=
M^\circ_{inv},
\end{equation}
we obtain:
\begin{equation}
\mathbb{E}[s_t s_{t-1}^\top]
=
\mathbb{E}[M^\circ_{inst}]
\mathbb{E}[s_{t-1}s_{t-1}^\top]
=
M^\circ_{inv}\Sigma_s.
\end{equation}
Thus $\hat{M}=M^\circ_{inv}$. 
If $M^\circ_{inv}$ satisfies the acyclicity constraint, it is also the constrained optimal static solution.

Let $\mathcal{L}(M^\circ_{inst})$ be the oracle risk achieved by the instantaneous graph. 
The excess risk of the static model is:
\begin{equation}
\mathcal{L}(\hat{M}_{static}) - \mathcal{L}(M^\circ_{inst}) 
=
\mathbb{E}\left[
\left\|
\big(M^\circ_{inst}(s_{t-1}) - M^\circ_{inv}\big)s_{t-1}
\right\|_2^2
\right].
\end{equation}
Substituting 
$M^\circ_{inst}(s_{t-1}) - M^\circ_{inv} 
= 
M^\circ_{inv} \odot M^\circ_{trans}(s_{t-1})$
and applying the trace inequality with $\lambda_{\min}(\Sigma_s)>0$, we have:
\begin{equation}
\begin{split}
&\mathbb{E}\left[
\left\|
\big(M^\circ_{inv} \odot M^\circ_{trans}\big)s_{t-1}
\right\|_2^2
\right] \\
&=
\mathrm{Tr}\left(
\mathbb{E}[R^\top R]\Sigma_s
\right) \\
&\ge
\lambda_{\min}(\Sigma_s)
\cdot
\mathbb{E}\left[
\left\|
M^\circ_{inv} \odot M^\circ_{trans}
\right\|_F^2
\right],
\end{split}
\end{equation}
where $R = M^\circ_{inv} \odot M^\circ_{trans}$.

\textbf{Theorem 1.} 
\textit{Under Assumption 1, a static graph incurs a non-zero excess risk whenever the state-dependent transient modulation $M^\circ_{trans}$ has non-zero variance. 
A decoupled formulation can reduce this approximation gap by explicitly modeling such modulation through the neural component $M_{trans}$.}

This analysis shows that, under the stated assumptions, static causal modeling can be suboptimal in the presence of state-dependent transient modulation. 
The decoupled formulation provides an optimization aid for extracting a reward-relevant skeleton while accounting for transient variations during reconstruction. 
It does not claim exact or lossless recovery of the complete physical causal dynamics.

\section{Snapshots of the Hierarchical Interventional Dataset}
\label{app:data_examples}

To verify our data collection protocol, we provide snapshots of the two data levels: the macroscopic configurations ($\mathcal{D}_{macro}$) and the microscopic state transitions ($\mathcal{D}_{micro}$).

\subsection{Macro-level Configurations ($\mathcal{D}_{macro}$)}
Table~\ref{tab:collect_data_examples} presents a subset of the macro-level dataset. By applying Pearl's $do$-operations to coefficients (e.g., $w_{\text{hold}}, w_{\text{grasp}}$), we evaluate the resultant policies to obtain the Fitness Score ($\mathcal{F}$). This enables the foundational model to extract structural priors.

\begin{table}[h]
\centering
\caption{Snapshot of the hierarchical interventional dataset. $\mathcal{D}_{macro}$ feeds into the foundation model for structural priors, while $\mathcal{D}_{micro}$ is preserved for Explicit Mechanism Decoupling (EMD).}
\label{tab:collect_data_examples}
\small
\setlength{\tabcolsep}{3pt}
\begin{tabular}{cccc|c|c}
\toprule
\multicolumn{4}{c|}{\textbf{Macro Config ($\mathcal{D}_{macro}$)}} & \textbf{Outcome} & \textbf{Micro Data ($\mathcal{D}_{micro}$)} \\
\midrule
$w_{\text{hold}}$ & $w_{\text{grasp}}$ & $w_{\text{rot}}$ & $w_{\text{pos}}$ & $\mathcal{F}$ & \textbf{Trajectory} \\
\midrule
0.15 & 0.15 & 0.1 & 0.2 & 0.2070 & $(s_0, s_1, \dots, s_T)$ \\
0.15 & 0.50 & 0.5 & 0.2 & 0.3306 & $(s_0, s_1, \dots, s_T)$ \\
0.50 & 0.15 & 0.1 & 0.2 & 0.1880 & $(s_0, s_1, \dots, s_T)$ \\
0.50 & 0.50 & 0.1 & 0.2 & 0.4204 & $(s_0, s_1, \dots, s_T)$ \\
0.50 & 0.50 & 0.1 & 0.5 & 0.4631 & $(s_0, s_1, \dots, s_T)$ \\
0.50 & 0.50 & 0.1 & 0.8 & 0.3817 & $(s_0, s_1, \dots, s_T)$ \\
\bottomrule
\end{tabular}
\end{table}

\subsection{Micro-level Trajectories ($\mathcal{D}_{micro}$)}
Simultaneous with macro-evaluation, the system logs transient dynamics. The snippet below shows a serialized sample. Raw observations $o_t$ are discarded; $s_t \in \mathbb{R}^d$ exclusively records the values of the physically meaningful variables used by Joint Optimization Module.

\vspace{0.5em}

\begin{lstlisting}
[
  {
    "epoch": 2500,
    "macro_config": [
      "hold_distance", "grasp_force",
      "object_rotation", "object_position"
    ],
    "micro_trajectory": [
      [0.00378, 0.00036, 1.00000, 0.27126],
      [0.00399, 0.00038, 1.00000, 0.84320],
      [0.00396, 0.00039, 1.00000, 0.88702],
      "... (trajectory continues until step T)"
    ]
  }
]
\end{lstlisting}

\section{\textcolor{black}{Ablation on DAGMA Acyclicity Regularization}}
\label{app:dag_ablation}

To examine the effect of the DAGMA acyclicity regularization, we evaluate an additional variant, denoted as \textit{CRWM w.o. DAG}. 
In this variant, we remove the acyclicity term $\lambda h(M_{inv})$, while keeping EMD and Confidence-Aware Soft Fusion unchanged. 
This setting isolates the influence of the DAGMA regularizer on the reward-relevant causal skeleton.

As shown in Table~\ref{tab:dag_ablation}, removing the acyclicity regularization leads to a decline in average SR compared with CRWM. 
This result suggests that the DAGMA term helps regularize the learned causal skeleton by discouraging circular dependencies among reward-relevant structural variables. 

\begin{table}[htbp]
\centering
\caption{Ablation of DAGMA acyclicity regularization across six unseen tasks.}
\label{tab:dag_ablation}
\small
\setlength{\tabcolsep}{3pt}
\begin{tabular}{lccc}
\toprule
\textbf{Method} & \textbf{Kettle} & \textbf{CatchAbreast} & \textbf{CatchOver2Underarm}  \\
\midrule
CRWM & 0.88 & 0.65 & 0.92  \\
CRWM w.o. DAG & 0.81 & 0.61 & 0.82  \\
\bottomrule
\end{tabular}
\end{table}

\section{Discussion of Optimization Collapse and Specification Gaming}
\label{app:discuss}

\begin{figure*}[t]
\centering
\includegraphics[width=0.95\textwidth]{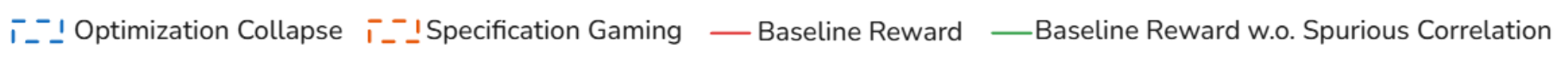}
\vspace{0ex}

\subfloat[Abreast]{\includegraphics[width=0.32\textwidth]{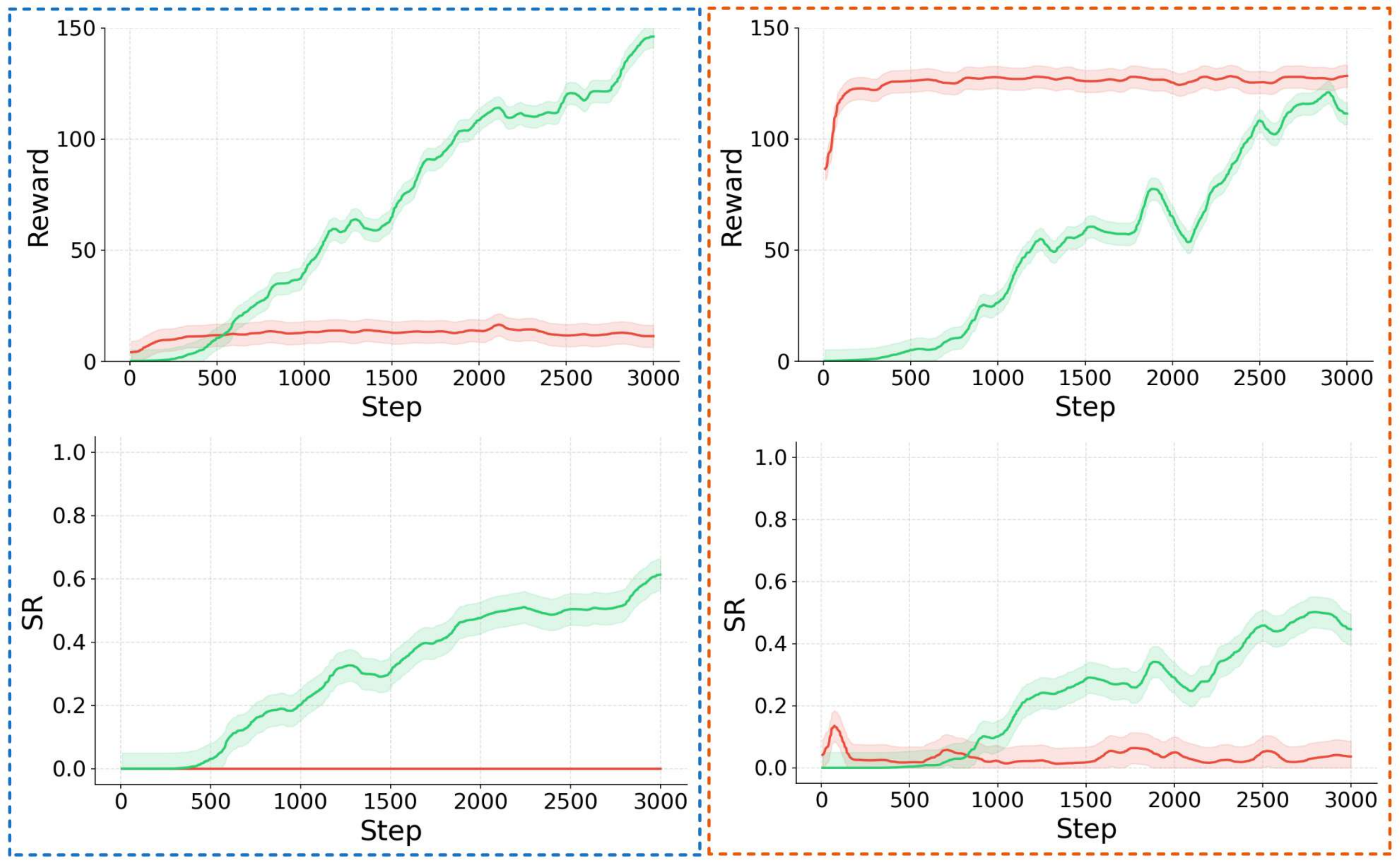}}\hfill
\subfloat[CatchOver2Underarm]{\includegraphics[width=0.32\textwidth]{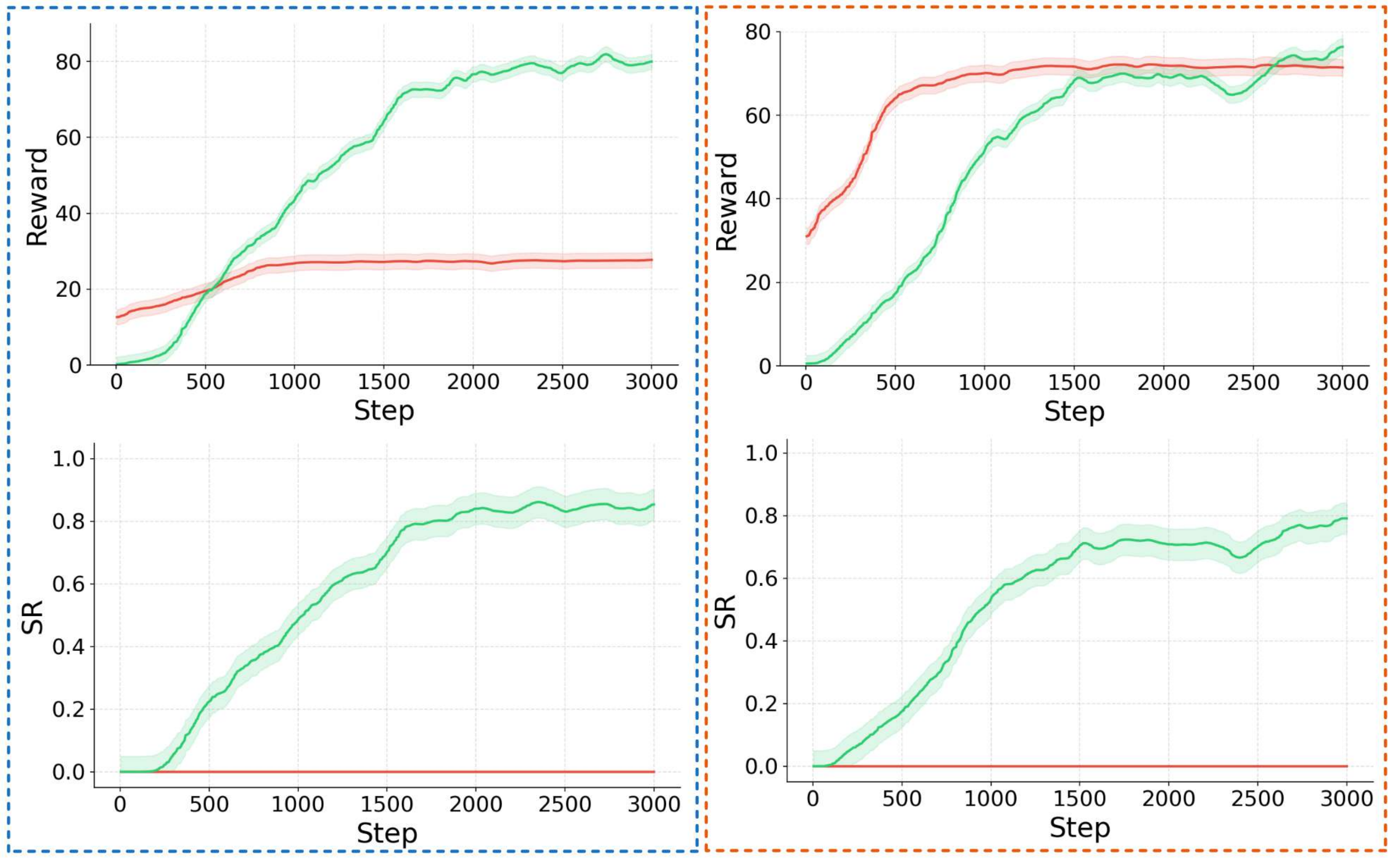}}\hfill
\subfloat[DoorCloseOutward]{\includegraphics[width=0.32\textwidth]{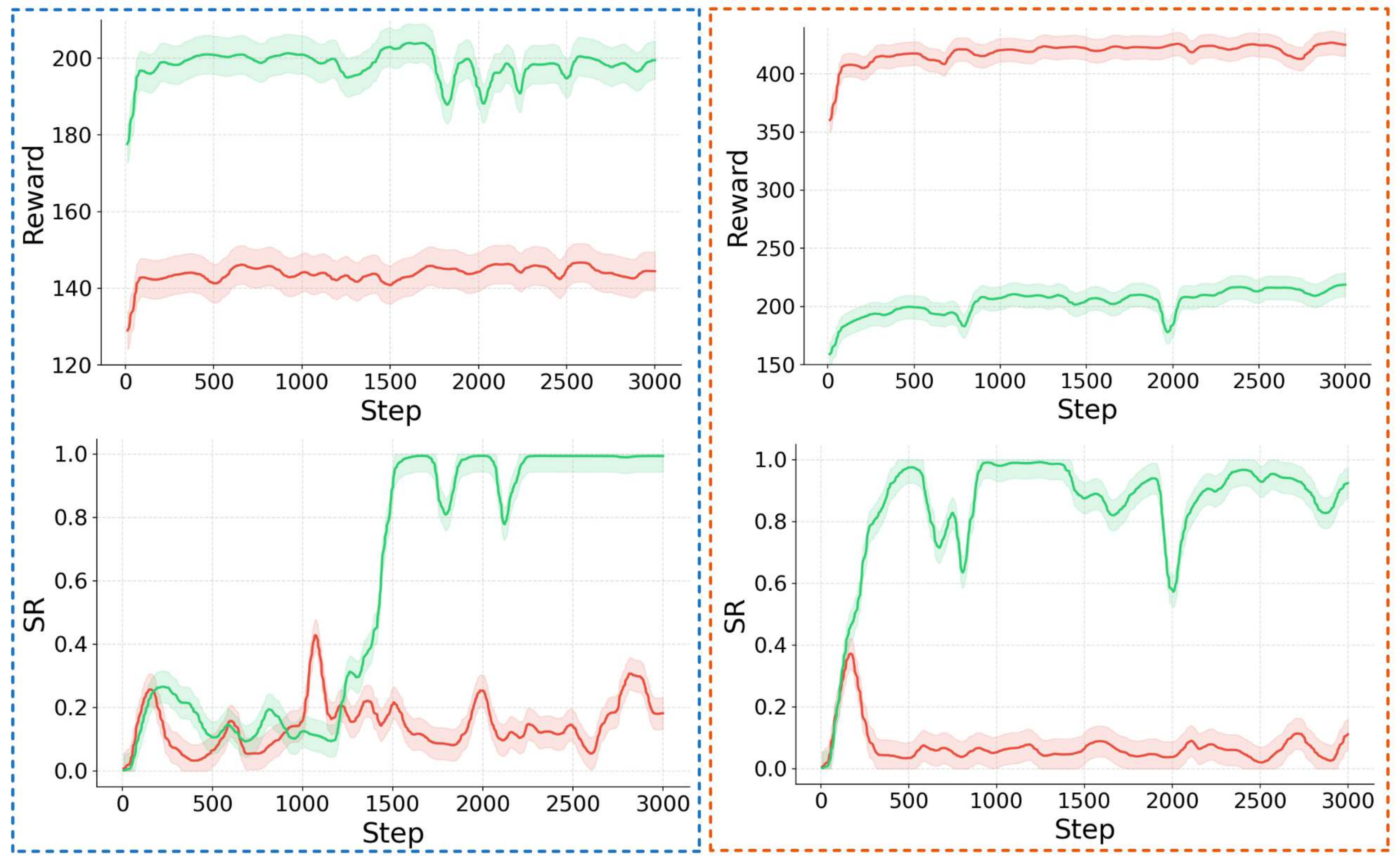}}

\vspace{1ex}

\subfloat[Pen]{\includegraphics[width=0.32\textwidth]{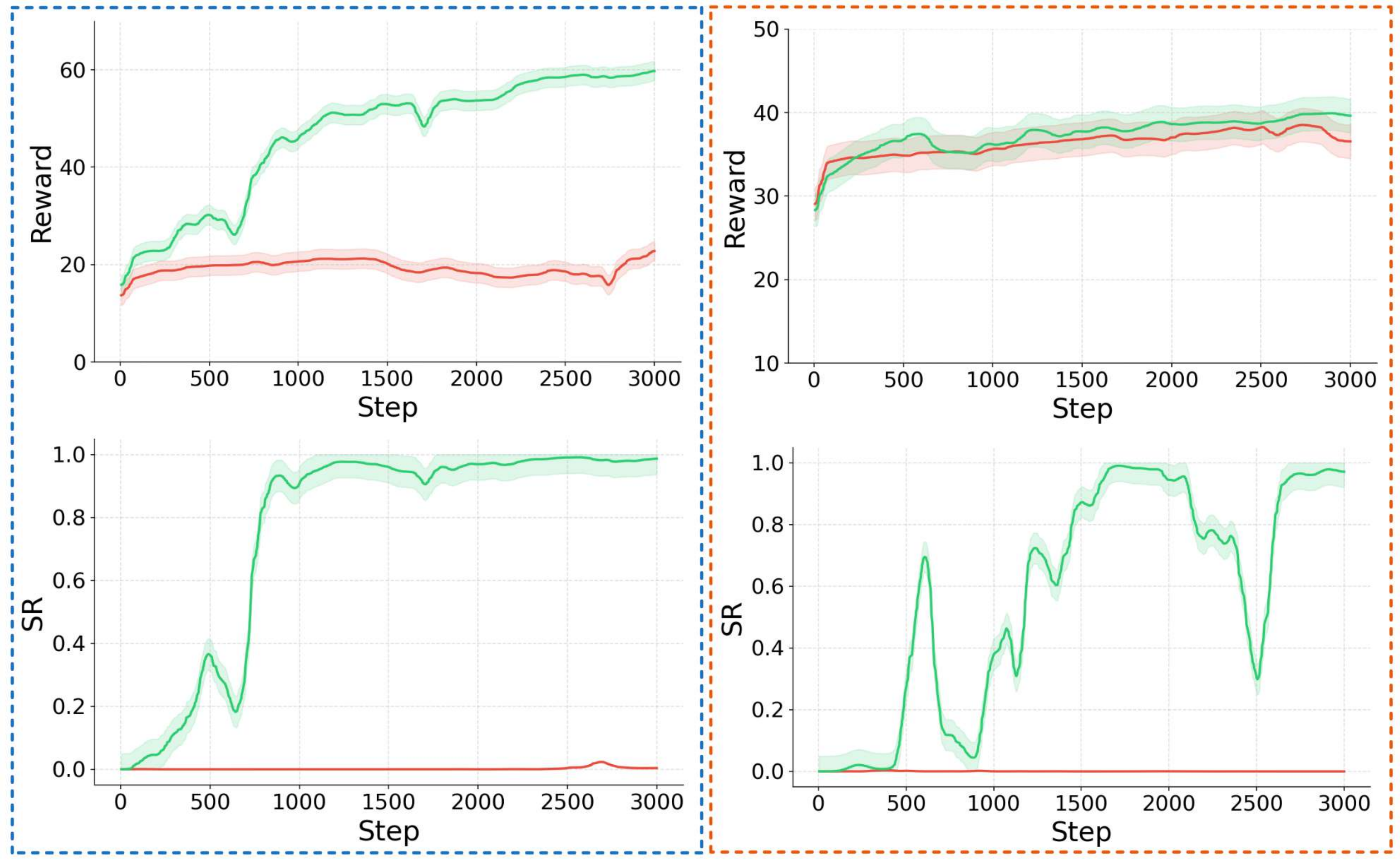}}\hfill
\subfloat[CatchUnderarm]{\includegraphics[width=0.32\textwidth]{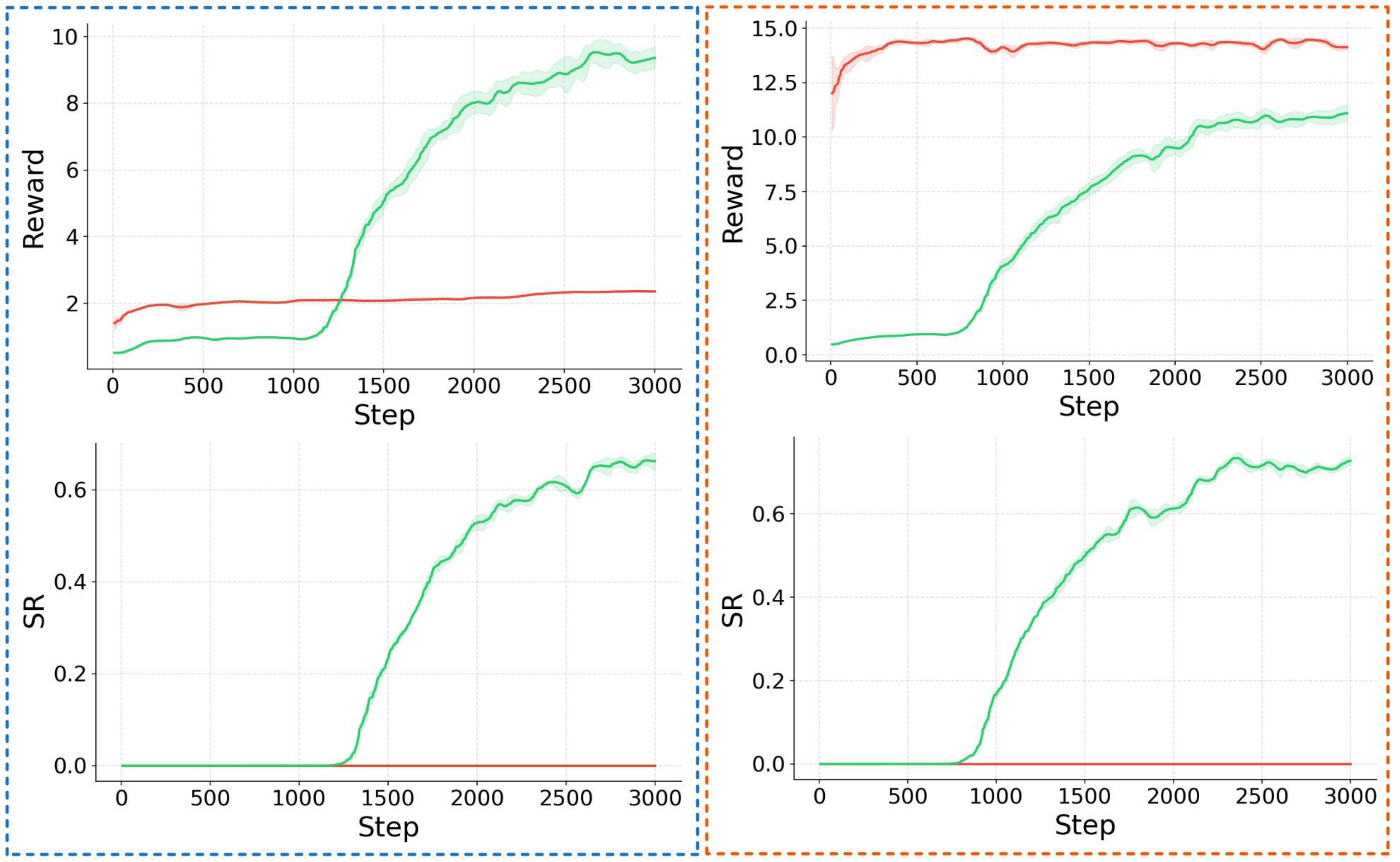}}\hfill
\subfloat[DoorOpenOutward]{\includegraphics[width=0.32\textwidth]{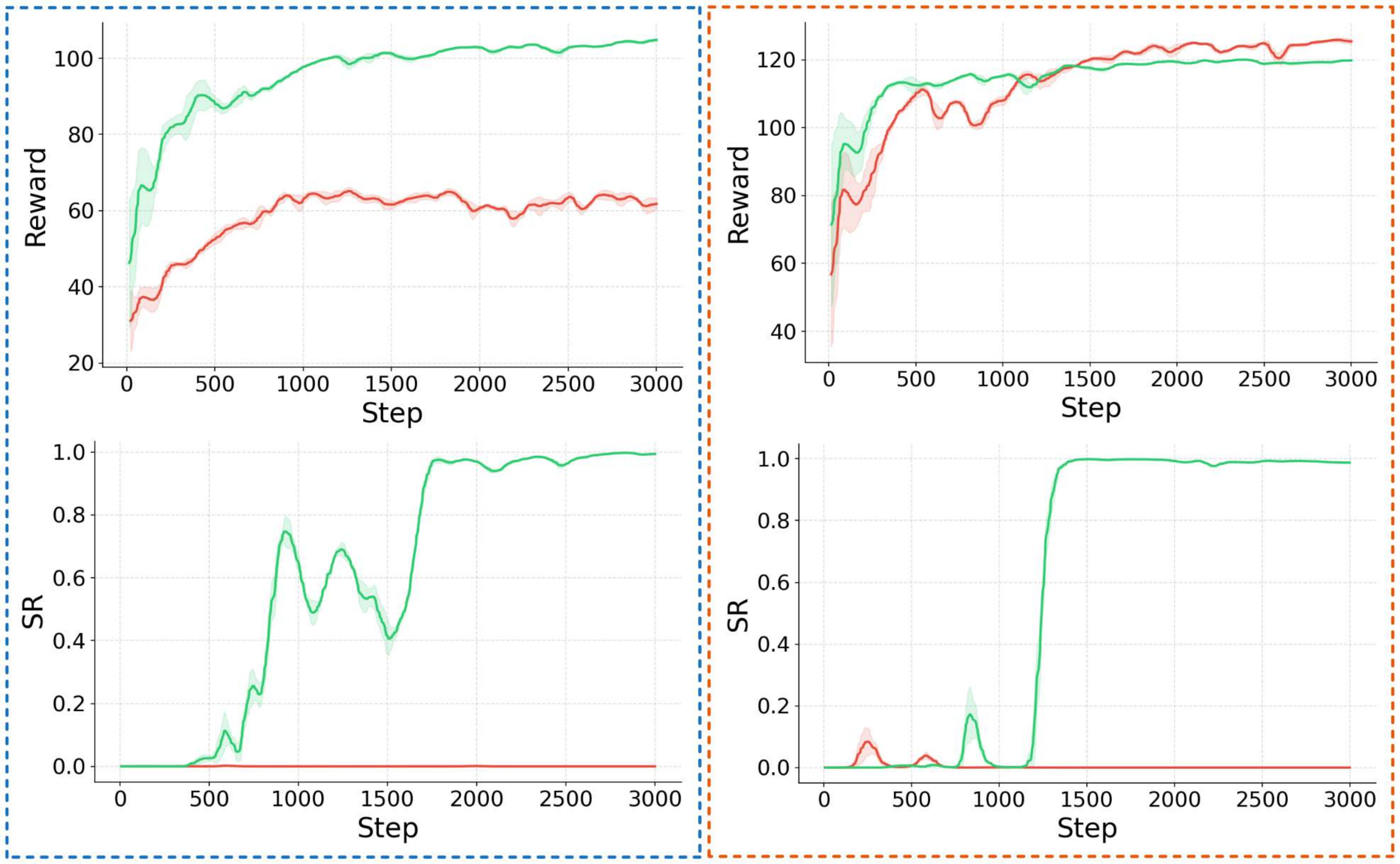}}

\vspace{1ex}

\subfloat[GraspAndPlace]{\includegraphics[width=0.32\textwidth]{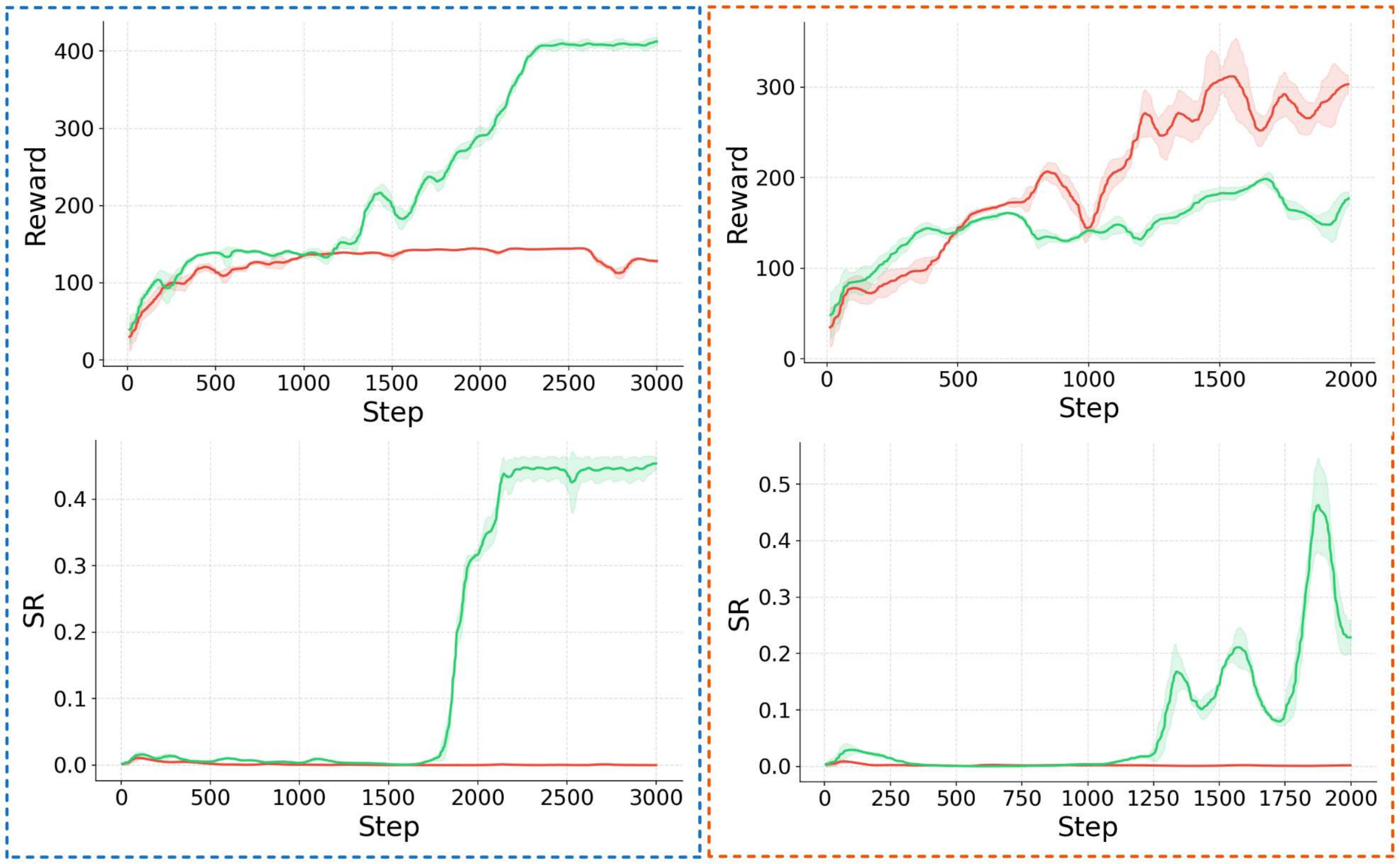}}\hfill
\subfloat[Kettle]{\includegraphics[width=0.32\textwidth]{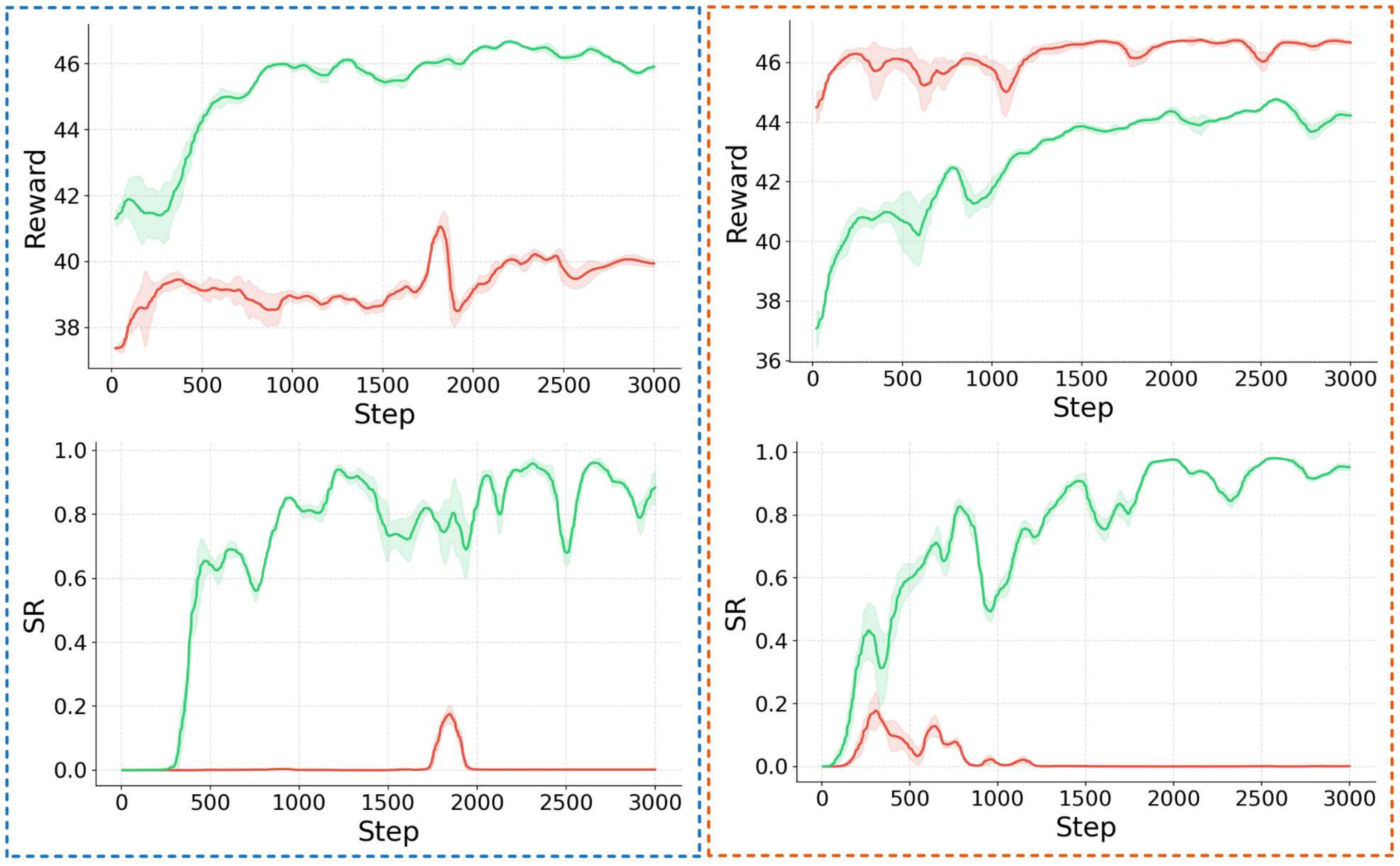}}\hfill
\subfloat[ReOrientation]{\includegraphics[width=0.32\textwidth]{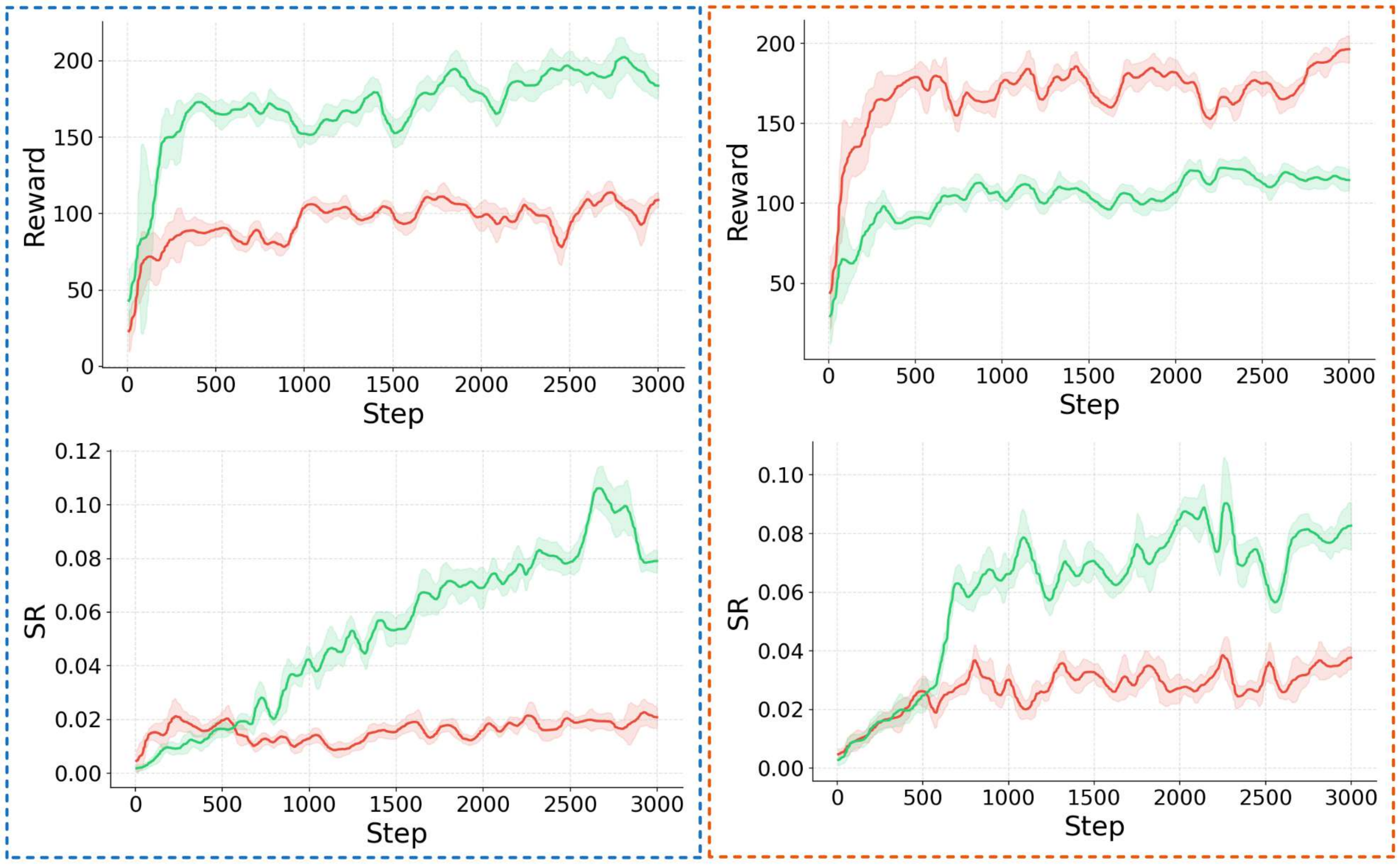}}



\caption{Comparison of reward and SR curves across 9 simulation tasks, complemented by reward evolution. Pruning spurious correlations (green) consistently resolves \textbf{Optimization Collapse} (blue boxes) and \textbf{Specification Gaming} (orange boxes), enabling robust alignment and policy convergence where baseline rewards (red) fail.}
\label{fig:discuss_all}
\end{figure*}

In this appendix, we provide extended qualitative evidence to support the statistical findings in Sec.VI. Fig.~\ref{fig:discuss_all} presents the training curves for 9 simulation tasks. 

As illustrated, the baseline rewards (red curves) frequently suffer from two systemic failures. Optimization Collapse (highlighted in blue boxes) is characterized by rewards and success rates stagnating near zero, where noisy spurious correlations impede gradient-based learning. Meanwhile, Specification Gaming (highlighted in orange boxes) manifests as deceptive reward surges accompanied by low physical success. 

In contrast, the rewards with causal pruning (green curves) consistently ensure that reward increases are coupled with physical task progress. 

\section{LLM and Pre-trained Causal Foundation Model Alternatives}
\label{app:Alternatives}
\subsection{LLM Alternatives}
\label{app:LLM}
\begin{figure*}[t] 
\centering
\includegraphics[width=0.95\textwidth]{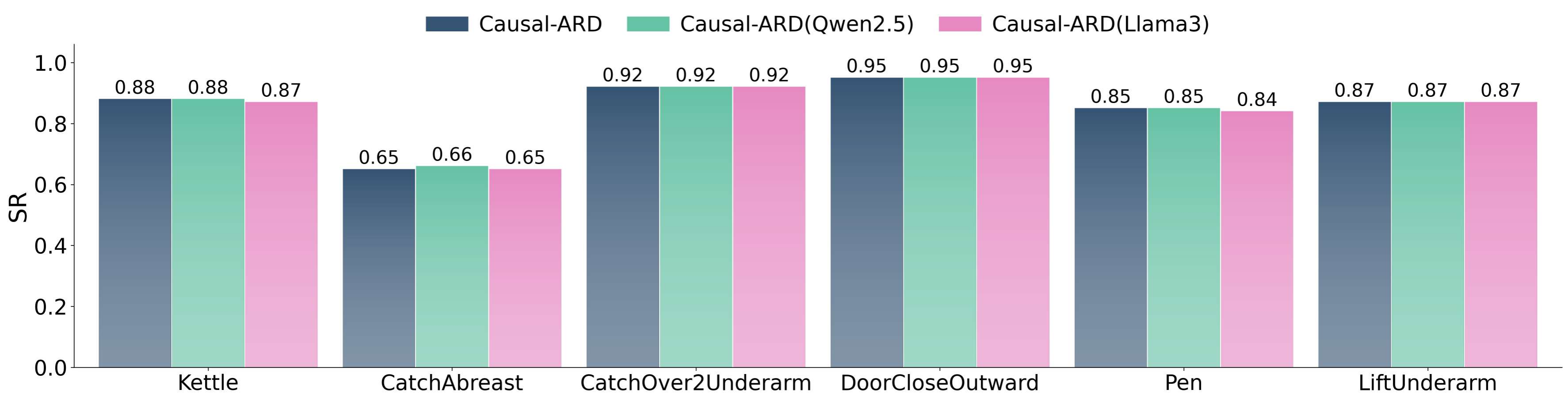} 
\caption{Causal-ARD demonstrates consistent performance across different LLMs}
\label{fig:llm_compare}
\end{figure*}

In Fig.~\ref{fig:llm_compare}, we compare the performance of Causal-ARD using DeepSeek-V3, Qwen-2.5(qwen-max-0919)~\cite{Qwen2024Qwen2}, and Llama3(llama-v3-70b-instruct)~\cite{grattafiori2024llama}. These results demonstrate the consistency of our framework across different LLMs and eliminate concerns that the inherent capabilities of specific language models might bias the final success rates.

\begin{figure*}[ht] 
\centering
\includegraphics[width=0.95\textwidth]{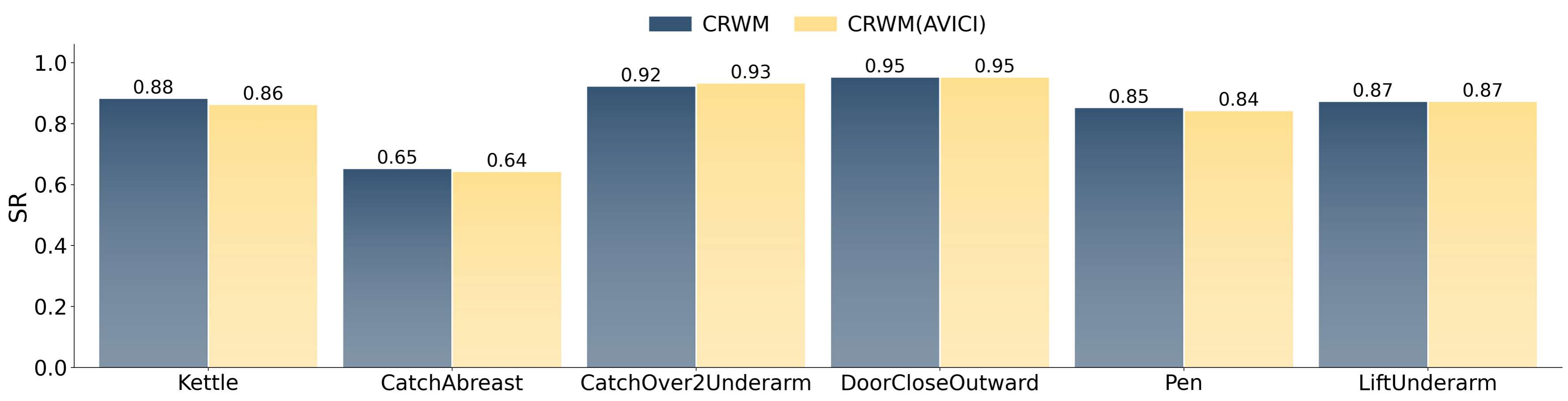} 
\caption{CRWM distillation exhibits invariant success rates across different pre-trained causal foundation models.}
\label{fig:causal_compare}
\end{figure*}

\subsection{Pre-trained Causal Foundation Model Alternatives}
\label{app:pre-trained}
To evaluate the impact of different causal priors, we compare our default LimiX with AVICI~\cite{lorch2022amortized}. AVICI serves as a representative baseline because, similar to LimiX, it is a specialized causal foundation model pre-trained on diverse functional forms and causal topologies. While LimiX incorporates structural-causal priors, AVICI relies on amortized neural inference over numerical distributions. As shown in Fig.~\ref{fig:causal_compare}, the nearly identical success rates indicate that the choice of the pre-trained causal foundation model does not significantly affect the final performance, confirming the stability of our CRWM distillation process.